\documentclass[preprint,12pt]{elsarticle}

\usepackage{amssymb}

\usepackage[caption=false,font=footnotesize]{subfig}
\usepackage{amsmath}
\usepackage{dsfont}
\usepackage{bm}
\usepackage{url}
\usepackage[usenames, dvipsnames]{color}
\usepackage{textcomp}

\newcommand*{\eg}{e.g.\ }
\newcommand*{\ie}{i.e.\ }
\newcommand*{\etal}{et al.\ }
\newcommand{\PP}{\mathds{P}}

\DeclareMathOperator*{\argmax}{arg\,max}

\graphicspath{{./figures/}}

\journal{Computer Vision and Image Understanding}

\begin{document}

\begin{frontmatter}



\title{Exploring structure for long-term tracking of multiple objects in sports videos}


\author[usp]{Henrique Morimitsu\corref{cor1}}
\ead{henriquem87@vision.ime.usp.br}
\cortext[cor1]{Corresponding author}

\author[telecom]{Isabelle Bloch}
\ead{isabelle.bloch@telecom-paristech.fr }

\author[usp]{Roberto M. Cesar-Jr.}
\ead{rmcesar@usp.br}

\address[usp]{Institute of Mathematics and Statistics, University of S\~ ao Paulo, S\~ ao Paulo, Brazil}
\address[telecom]{LTCI, CNRS, T\'el\'ecom ParisTech, Universit\' e Paris - Saclay, Paris, France}

\begin{abstract}
   In this paper, we propose a novel approach for exploiting structural relations to track multiple objects that may undergo
   long-term occlusion and abrupt motion. We use a model-free approach that relies only on annotations given in the first frame
   of the video to track all the objects online, \ie without knowledge from future frames. We initialize a probabilistic Attributed
   Relational Graph (ARG) from the first frame, which is incrementally updated along the video. Instead of using the
   structural information only to evaluate the scene, the proposed approach considers it to generate new tracking hypotheses.
   In this way, our method is capable of generating relevant object candidates that are used to improve or recover the track
   of lost objects. The proposed method is evaluated on several videos of table tennis, volleyball, and on the ACASVA dataset.
   The results show that our approach is very robust, flexible and able to outperform other state-of-the-art
   methods in sports videos that present structural patterns.
   
   This version corresponds to the preprint of the CVIU paper. The final version is available on: \url{http://dx.doi.org/10.1016/j.cviu.2016.12.003}.
   \textcolor{White}{\footnote{\textcopyright 2016. This manuscript version is made available under the CC-BY-NC-ND 4.0 license \url{http://creativecommons.org/licenses/by-nc-nd/4.0/}}}
\end{abstract}


\end{frontmatter}


\section{Introduction}
\label{sec:introduction}

Object tracking is a relevant field with several important applications, including surveillance,
autonomous navigation, and activity analysis. However, tracking several objects simultaneously is often a
very challenging task. Most multi-object trackers first track each object separately
by learning an individual appearance model and then consider all the results globally to improve or correct
individual mistakes. However, especially in sports videos, the use of appearance models proves to be insufficient
because usually many objects (or players) have very similar appearance due to the uniform they wear.
This often causes tracking loss after situations of occlusion between players of the same team. Another difficulty
is that most trackers rely on the constraint that temporal changes are smooth, \ie the position of an object does not change
significantly in a short period of time. Yet, this is not a reasonable assumption for most sports videos, because they are
usually obtained from broadcast television, and thus they are edited and present several camera cuts, \ie
when the scene changes suddenly due to a camera cut off, or change of point of view. Camera cuts often cause
problems of abrupt motion, which is regarded as a sudden change in position, speed or direction of the target. 

\begin{figure}[ht]
\begin{center}
   \subfloat[]{\includegraphics[width=0.22\linewidth]{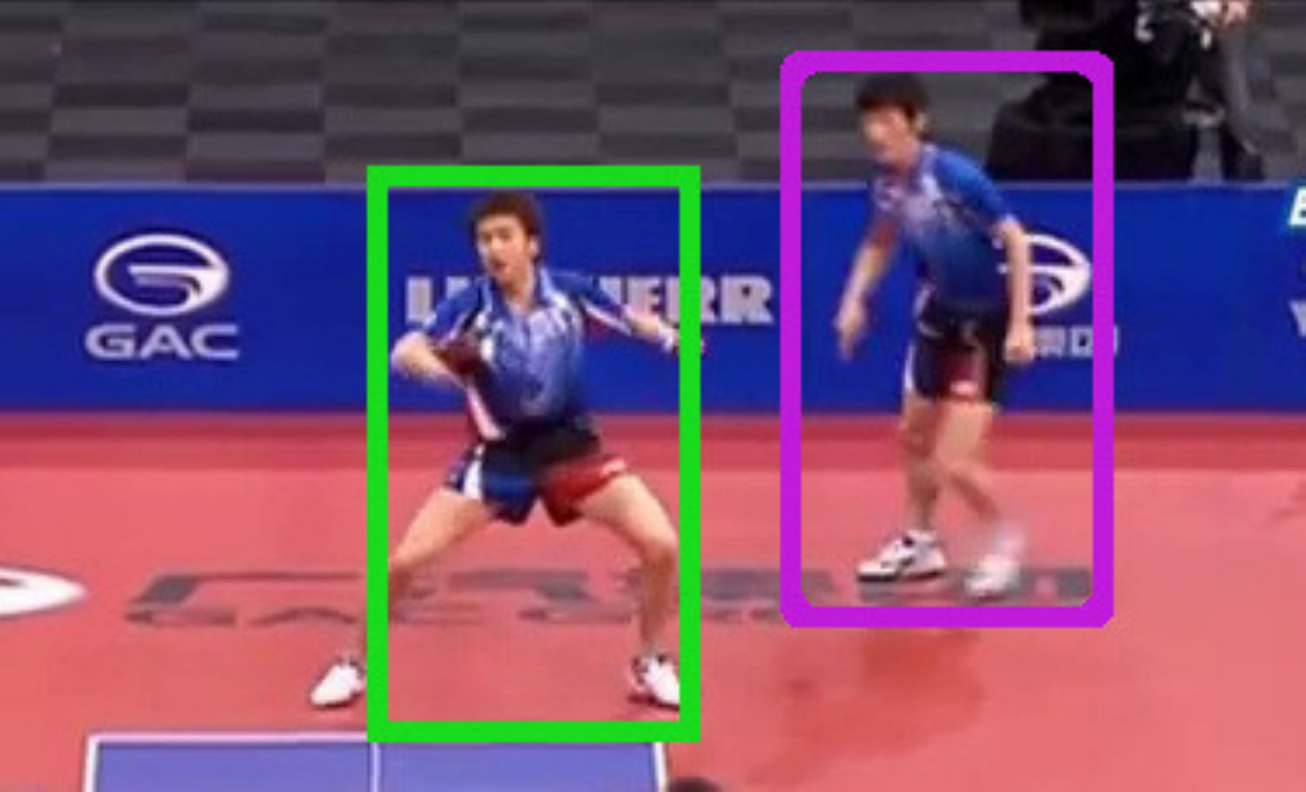}}
   \hfil
   \subfloat[]{\includegraphics[width=0.22\linewidth]{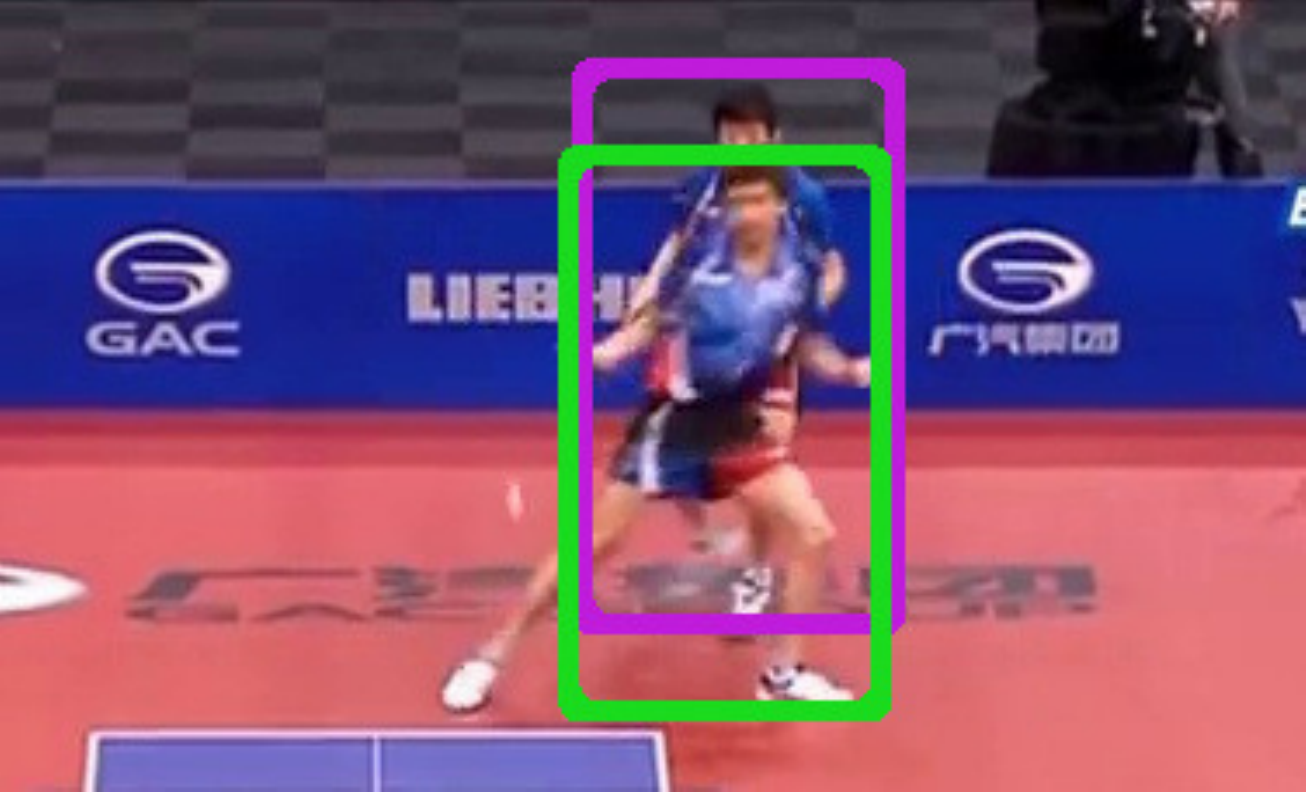}}
   \hfil
   \subfloat[]{\includegraphics[width=0.22\linewidth]{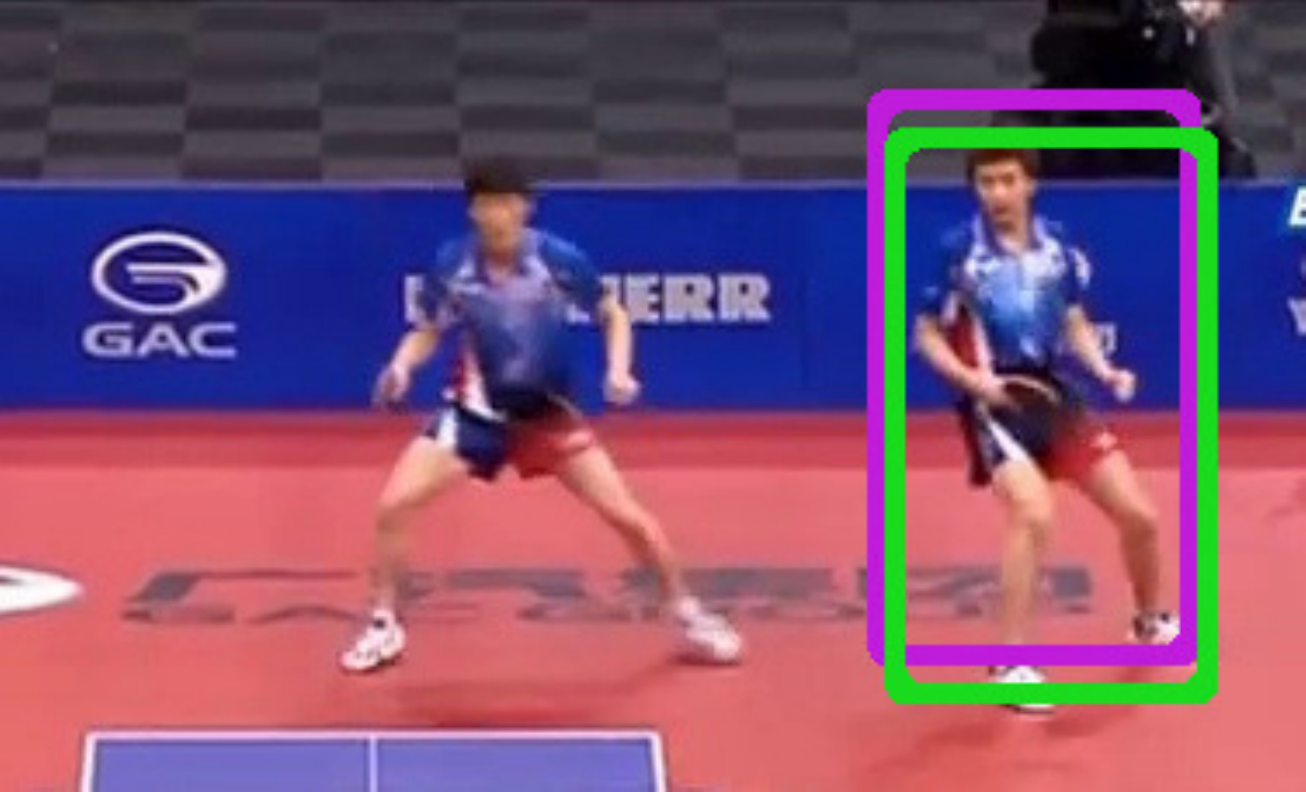}}
   \hfil
   \subfloat[]{\includegraphics[width=0.22\linewidth]{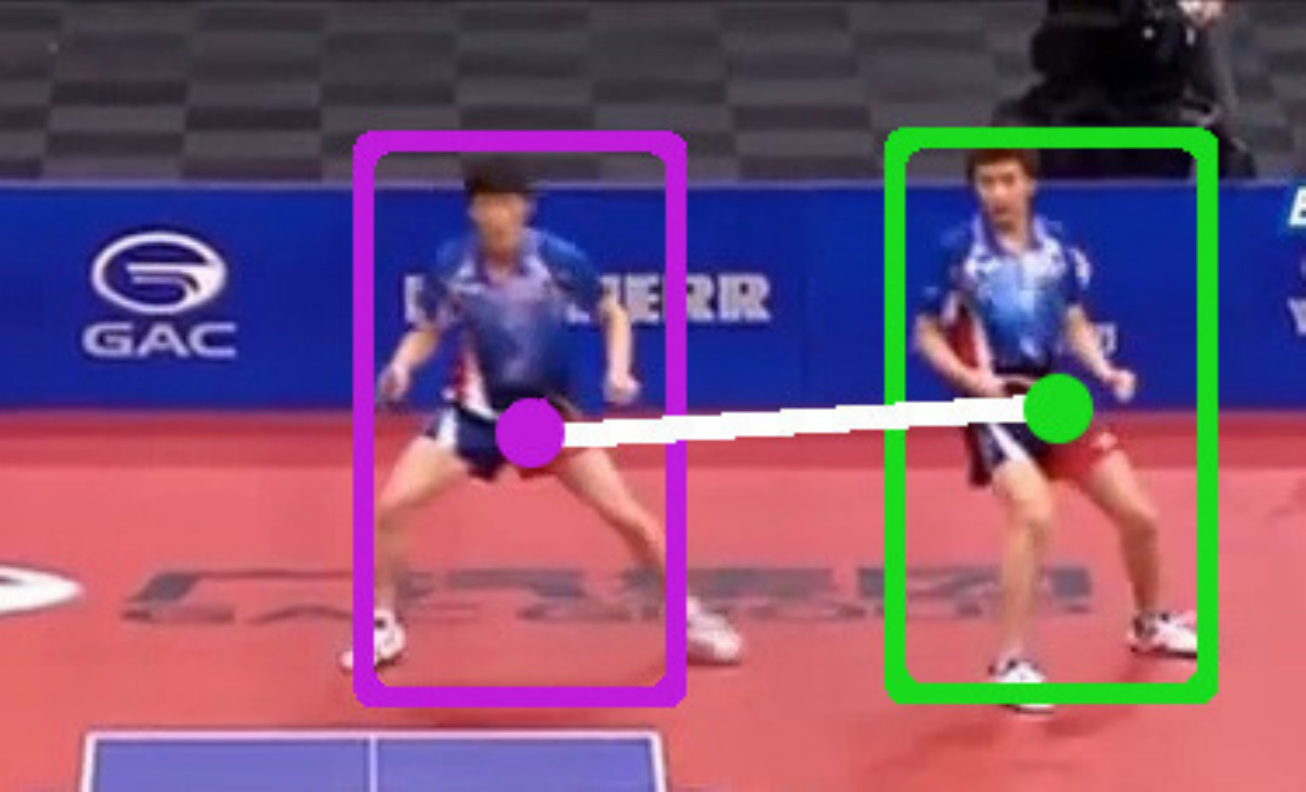}}
\end{center}
   \caption{An example of a multi-object tracking situation. Most single object trackers are able to successfully track the targets when
   their appearance is clear (a). However, when overlap occurs (b), they are not
   able to solve the ambiguity problem in appearance and the tracking is lost (c). We propose recovering tracking
   after such events by using structural spatial information encoded in graphs (d).}
\label{fig:example_tracking}
\end{figure}

In long-term tracking, the objects are subject to situations of full occlusion and abrupt motion,
which may lead to tracking failures. Therefore, the tracker must be able to recover the target
after such events. In this paper, we explore the use of spatial relations between objects
to recover or correct online tracking. Online tracking, as opposed to batch methods,
only uses past information to predict the next state.
We argue that, in some kinds of videos where the scene represents
a situation that usually follows a common spatial pattern, it is possible to
recover tracking by learning some
structural properties. Figure~\ref{fig:example_tracking} shows an example of a table tennis video illustrating a situation where tracking is lost
after two players intersect each other. Although the interaction is brief, this already causes one of the trackers
to misjudge its correct target and start to track the other player instead. We solve this kind
of problem by exploiting the spatial properties of the scene, such as the distance and angle between two objects.

We shall refer to videos that present discernible spatial patterns as structured videos.
It is assumed that scenes (or frames) of these videos contain elements that provide a kind of
stable spatial structure. A good example is sports videos.
Sports rely on a set of rules that usually constrain
the involved objects to follow a set of patterns. These patterns often present some
spatial relationships that may be exploited. For example, the rules may enforce that
the objects must be restricted to a certain area, or that they always keep a certain distance
among them.

Structural relations are utilized by using graphs to encode the spatial configuration of the scene.
In this paper, color-based particle filter was chosen as the single object tracker due to its simplicity and good
results demonstrated in previous studies. However, other trackers could also be
employed instead to benefit from the added structural information. 
This makes the proposed framework very flexible and able to be used to
potentially improve the results of any single object tracker applied in multi-objects
problems. As shall be explained throughout this text, the graphs are utilized to extract
structural information from the scene, to generate candidate object positions,
and then to evaluate the tracking state at each frame. With this approach, it is possible to improve the overall result by recovering
tracking after situations of overlapping between objects with similar appearance or when
abrupt motion happens.

The use of structural information for recovering tracking is a topic that has not been much explored in the literature before.
Indeed, several of the current state-of-the-art methods based on tracking by detection do use structural
information at a different level, for evaluating the tracking state and solving the data association problem between the frames.
However, the detections are usually carried out by off-the-shelf detectors that do not consider
scene information. In this sense, one main contribution of this paper is to introduce a new approach that exploits the learned
structural model to guide the detection. This approach allows tracking the objects even after challenging events
such as long-term occlusions of abrupt motion. The proposed framework is tested on sports
videos obtained from Youtube and also from the ACASVA~\cite{decampos2011evaluation} dataset,
which present real challenging conditions such as changing of lighting, mutual occlusion, and camera cuts.

The main contributions of this paper are the following: (1) to introduce a structural approach for
improving tracking by generating new object candidates, and (2) to formalize tracking as
a flexible graph optimization problem that can be adapted to other situations where structural patterns
exist.

This paper extends our previous work in~\cite{morimitsu2015attributed}. The main novelties of this study are:
firstly, the graph model is trained and updated online, without the need of a training dataset, making the method
much easier to use in different applications; secondly, the graph matching function does not rely on heuristics
for occlusion and change of tracker anymore. This information
is incorporated directly into the scoring function, thus allowing us to formalize the data association as a graph optimization problem;
thirdly, additional results on more cluttered and challenging videos of volleyball matches are included. They
are also analyzed more thoroughly using the widely adopted CLEAR-MOT metrics~\cite{bernardin2008evaluating}.

This paper is organized as follows. In Section~\ref{sec:related_works} we present a review of some relevant previous works and
how they contributed to the development of our method. In Section~\ref{sec:tracking_pf} we detail the particle
filter tracking approach and how it is applied in our problem.
In Section~\ref{sec:tracking} the complete framework for tracking multiple objects using graphs is explained.
In Section~\ref{sec:results} the experimental results obtained are exposed. 
We compare the obtained results for our approach with other state-of-the-art methods from the literature. 
In Section~\ref{sec:conclusion} we discuss the main conclusions of this work as well as suggestions for future
research.

\section{Related work}
\label{sec:related_works}

Visual tracking in sports has been receiving great attention over the years and it
has been tackled in many different ways. Due to its simplicity and robustness to deal with
more complex models, methods based on particle filters became popular~\cite{kristan2009closed,okuma2004boosted,xing2011multiple}.
Another significant approach relies on the fact that often the background is static and, therefore,
tracking may be performed by using background subtraction methods~\cite{figueroa2006background}.
Other authors explore the use of multiple cameras to obtain more reliable results~\cite{morais2014multiple}.

Recently, the tracking-by-detection framework has become the standard method for tracking
multiple objects~\cite{choi2015near,milan2014continuous,solera2015learning,zhang2015multi}.
This approach consists in obtaining noisy detections at each frame and then connecting them into
longer temporal tracks. For this, a data association problem must be solved between all the candidates
in order to create stable tracks for each object. The most important part is formulating the association
function so that the problem can be solved efficiently, while creating good tracks.
Liu \etal~\cite{liu2013tracking} have designed an association function specific for
tracking players in team matches. The tracks are associated by assuming a motion model that depends
on the game context at the moment. By exploiting local and global
properties of the scene, such as relative occupancy or whether one player is chasing another,
the authors show that the method is able to successfully track the players during a basketball match.
One important challenge in multi-object tracking consists in keeping
the correct identities of each object. Shitrit \etal~\cite{shitrit2014multi} demonstrate on several sports videos that
it is possible to keep the correct identity of the players by relying on appearance cues that
are collected sparsely along time. This is done by modeling the problem as flows expressed by
Directed Acyclic Graphs, which is solved using linear programming. 
Lu \etal~\cite{lu2013learning} show that it is possible to keep correct identities
even when using a simple data association strategy. On the other hand, this approach makes use
of a Conditional Random Field framework, which assumes both that the tracks for the whole video
are available, and that external data is available to train the model parameters.
Our proposed method may be interpreted as a tracking-by-detection framework, but instead of
using an object detector to generate candidates, we rely on the structural properties encoded
in the graph. The identification of each player is handled implicitly, by the graphs.
Although this choice is not always optimal, it is efficient, and it only relies on data
obtained from the past frames of the video itself.

One challenging condition frequently found in sports scenes is occlusion. Many previous
works focused on modeling it explicitly to handle these difficult situations~\cite{tang2014detection,xiang2015learning}.
Zhang \etal~\cite{zhang2013object} tackle this issue in sports
by using a structured sparse model for each person. This approach builds on the robustness
of sparse models by assuming that the occlusion model is usually not sparse, but rather
a structured connected area. This allows using better models which are able to
ignore features from large occlusion areas, \eg one player occluding another one.
In a related topic, Soomro \etal~\cite{soomro2015tracking} propose using structural
properties encoded in graphs to associate tracks when the videos of football games are cut, which
causes occlusion of players as well as abrupt motion. This problem is formulated as
a graph matching between different frames. In order to solve the ambiguity problem for the association,
the authors use knowledge about the previously learned team formation.
Therefore, the model requires some additional external information in order to successfully recover tracking.

The use of structural information for multi-object tracking has also been incorporated into
the SPOT tracker~\cite{zhang2014preserving}. The authors use a model-free approach that
learns the appearance and structural relations between the tracked objects online. In the first frame,
manual annotations are provided and used to initially train a Histogram of Oriented Gradients (HOG) detector~\cite{dalal2005histograms}
for finding the object in the next frames, \ie their approach is also based on tracking by detection.
The structural relations are also learned from the first frame by training a structured Support Vector Machine (SVM). The models
are then updated while the tracking is being performed, using a gradient descent approach.
The candidate graphs are evaluated using the information obtained from the HOG detectors as well as
the distances between any two objects.

Although similar, their proposal differs from the one presented in this paper in the following aspects: (1) their structural
model only computes the difference between the observed distance and an ideal value that comes from
the online training. Our model considers both distance and angle information obtained from
a more general probability density function. (2) Although
they use the structure to improve tracking and to deal with occlusion, it is not used to guide
the detection process, which could lead to improved performance by restricting the search space.
Our approach, inspired by~\cite{grabner2010tracking}, uses some objects as supports for 
the structural model to generate candidates of where
the target is likely to be found after tracking loss. (3) Their method of tracking by detection
does not consider motion models, while we rely on particle filters.

Another important issue that must be dealt with during tracking is abrupt motion.
Perhaps, the simplest way to deal with it is by generating some additional
target location hypotheses around the previous location to try to cover
a broader region, as explored in~\cite{zhang2012real}. 
Another proposal is to solve the same problem using spatial position information for finding
better candidates~\cite{kwon2008tracking,zhou2010abrupt}. This is done by dividing the image
into several smaller regions and using the information obtained from the density of states
of each one to find new locations. More recently, Su \etal~\cite{su2014abrupt}
propose relying on visual saliency information to guide the tracking and
restore the target. It is important to note that, although
tracking-by-detection methods should be able to deal with abrupt motion, most of them do
not behave well in this situation because their association function usually
enforces the temporal stability constraint. As previously mentioned,
we use a different approach that relies on the structural relations between the objects to find the most likely new locations
of a lost target.

\section{Tracking objects with particle filters}
\label{sec:tracking_pf}

In this section, the standard method of tracking with particle filters is briefly summarized.
A classical filtering problem operates over a Hidden Markov Model~\cite{fink2008markov}.
Let $\mathcal{X}$ and $\mathcal{O}$ be the sets of states and observations, respectively.
Let $\bm{x}_t \in \mathcal{X}$ be a hidden state at instant $t$ and $\bm{o}_{t} \in \mathcal{O}$
an observation emitted by $\bm{x}_t$. It is assumed that the model is a Markov process of
first order and $\bm{x}_t$ is conditionally independent of the joint of previous states and observations.
The filtering problem consists in estimating recursively the posterior distribution $\PP(\bm{x}_t | \bm{o}_{1:t})$,
where $\bm{o}_{1:t}$ denotes the set of observations from instant 1 to instant $t$.

Except if the system presents some properties such as Gaussian distributions and linear models, the distribution
cannot be computed analytically. When the system is more complex, then the result can only be approximated using, for example,
a particle filter.

Let $\bm{x}_t^i$ be a hypothetical realization of the state $\bm{x}_t$ and $\delta_{\bm{x}_t^i}(\bm{x}_t)$ be the Dirac
delta function centered at $\bm{x}_t^i$. A particle filter solves the filtering problem by approximating the posterior 
probability $\PP(\bm{x}_t | \bm{o}_{1:t})$ by a weighted sum of $N_P$ Dirac masses:
\begin{equation}
  \PP(\bm{x}_t | \bm{o}_{1:t}) = \sum_{i = 1}^{N_P}{\pi_{t}^i \delta_{\bm{x}_t^i}(\bm{x}_t)},
\end{equation}
where each $\bm{x}_t^i$ is called a particle with associated weight $\pi_{t}^i$.

This work employs particle filter using the ConDensation algorithm, which uses factored sampling~\cite{isard1998condensation}
to update the particles. 
The particles are then propagated according to a proposal function
$\bm{x}_t^i \sim q(\bm{x}_t | \bm{x}_{0:t - 1}^i, \bm{o}_{1:t})$,
which is usually assumed to be the dynamics model $\PP(\bm{x}_t | \bm{x}_{t - 1})$,
yielding $\bm{x}_t^i \sim \PP(\bm{x}_t | \bm{x}_{t - 1})$. 

The propagation phase involves two steps: drift and diffusion.
Drift is a deterministic step, which consists in applying the motion dynamics for each particle.
Diffusion, on the other hand, is random and it is used to include noise in the model.
The new state of a particle $i$ can then be expressed as:
\begin{equation}
  \bm{x}_{t}^i = \bm{D} \bm{x}_{t - 1}^i + \bm{u},
\end{equation}
where $\bm{D}$ is the motion dynamics and $\bm{u}$ is the noise vector.

Finally, the weights of the particles are updated according to the new observations $\bm{o}_t$ and,
if the proposal function is the dynamics model, the weight update is simply:
\begin{equation}
  \pi_{t}^i \propto \pi_{t - 1}^i \PP(\bm{o}_{t} | \bm{x}_{t}^i).
\end{equation}

The final estimated state of a cloud of particles $P$ may be computed using several heuristics,
but the most widely used is by computing the weighted average:
\begin{equation}
  r(P) = \bm{x}_t = \sum_{i = 1}^{N_P}{\pi_t^i \bm{x}_t^i}
  \label{eq:particle_state}
\end{equation}

In this work, we are also interested in evaluating the overall quality of a cloud $P$. We propose doing this by computing
a confidence score based on the non-normalized weights of the particles:
\begin{equation}
  \zeta(P) = 1 - \exp\left(-\sum_{i=1}^{N_P}{\pi_{t - 1}^i \PP(\bm{o}_{t} | \bm{x}_{t}^i)}\right).
  \label{eq:particle_score}
\end{equation}
This increasing function ensures that the score is close to $1$ when the sum of the weights is high.

\subsection{State and dynamics models}
\label{subsubsec:particle_models}

The objects to be tracked are represented by rectangular bounding boxes. Each box is parameterized
by its centroid and two measures: height and width. It is assumed that the variation in scale is not significant.
Therefore, the state of each particle is given by a column vector $\bm{x}_{t}^i = (x^i_t, y^i_t)^T$, which represents one candidate centroid.
The motion model is a random walk, yielding:
\begin{equation}
  \bm{x}_{t}^i = \bm{x}_{t - 1}^i + \bm{u}
\end{equation}
where $\bm{u} = (u_x, u_y)^T$ is a noise vector whose terms follow a
Gaussian distribution $\mathcal{N}(0, \hat{\sigma_u})$. In this work, we adopt an adaptive variance $\hat{\sigma_u}$
that is computed by weighting
the fixed variance $\sigma_u$ according to the quality of the particles. The weight is obtained by:
\begin{equation}
  \lambda = \alpha \left(1 - \frac{\sum_{i = 1}^{N_P}{\PP(\bm{o}_{t} | \bm{x}_{t}^i)}}{\beta}\right),
\end{equation}
where $\alpha$ is a given spreading factor that controls the impact of the weight on the real variance and
$\beta$ is an upper bound for the maximum weight of the particles. We can choose $\beta = N_P$ or another
suitable value according to the data. However, it is not interesting to remove the variance altogether,
thus we set a lower bound $\tau_\lambda$, yielding the final weight:
\begin{equation}
  \hat{\lambda} = \max\{\lambda, \tau_\lambda\}.
\end{equation}
Finally, the adaptive variance is computed as:
\begin{equation}
  \hat{\sigma_u} = \hat{\lambda} \sigma_u.
\end{equation}

More complex states and motion models could be used. For example, the state could also include additional information
such as the velocity, acceleration, orientation, scale and so on. The greatest problem when considering more information
is that each additional parameter increases the search space in one dimension. Since the amount of particles required to
track multiple objects increases fast, the smallest number of particles that produced good results was chosen.

\subsection{Color histogram-based tracking}
\label{subsubsec:particle_color}

The objects are tracked using color histograms~\cite{perez2002color}. 
The method works by using color information obtained from the HSV color space.
This color model is interesting because it separates the chromatic information (Hue and Saturation) from the shading (Value).
However, the authors point out that the chromatic information is only reliable when both the Saturation and the Value are
not too low. Therefore, first an $H_{HS}$ histogram with $N_{H_H} N_{H_S}$ bins is populated using only information obtained from pixels whose Saturation
and Value are above some given thresholds of 0.1 and 0.2, respectively. The remaining pixels are not discarded because their
information can be useful when dealing with images which are mostly black and white, and they are used
to populate an $H_V$ histogram that is concatenated to the $H_{HS}$ one built before. The resulting histogram is composed
of $N_{H_H} N_{H_S} + N_{H_V}$ bins. Following the aforementioned paper, the variables are set as $N_{H_H} = N_{H_S} = N_{H_V} = 10$.

Each histogram corresponds to one observation $\bm{o}_t^j$ for object $j$ at instant $t$ for the particle filter.
Section~\ref{subsubsec:appearance_score} presents more details about how the histograms are compared in order to
track each object.

\section{Multi-object tracking based on structural information}
\label{sec:tracking}

\begin{figure}[ht]
\begin{center}
   \includegraphics[width=\linewidth]{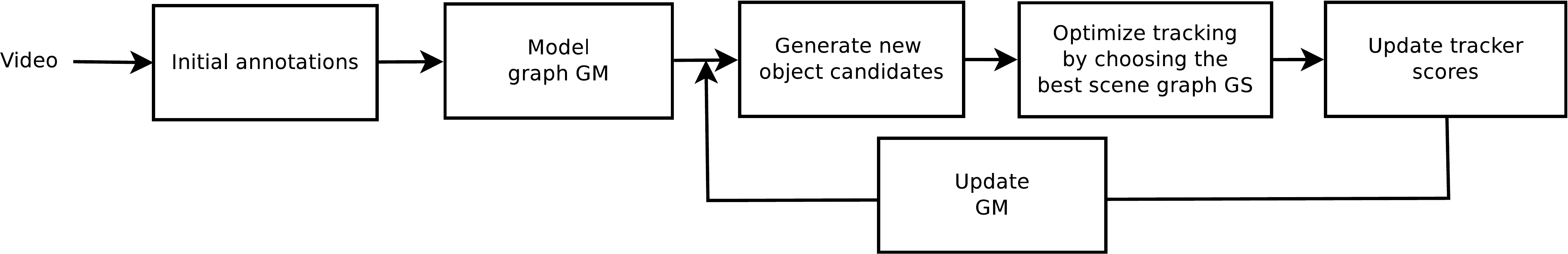}
\end{center}
   \caption{An overview of the proposed framework.}
\label{fig:framework}
\end{figure}

In this section, the proposed tracking framework is explained. Figure~\ref{fig:framework} shows a 
flowchart of the process. First, a video with annotations for the first frame is provided as input.
Then, the annotations are used to learn the initial structural graph model, which is employed to generate
some additional object candidates for recovering tracking in case of loss. The candidates are evaluated
according to the model and the final result is used to update the model before processing the next frame.

Tracking is carried out by evaluating two Attributed Relational Graphs (ARGs): a model graph $G^M$ and a scene graph $G^S$.
The definition of both ARGs is presented in Section~\ref{subsec:arg}.
For the tracking step, the position of each of the $N_O$ objects can be either manually annotated in the first frame of the video or
obtained automatically using detectors.
In this work, we adopt the former option. First, the model graph $G^M$ of the image structure
is learned using the annotations from the first frame and then updated online at each frame of the video (Section~\ref{subsec:updating_model}).
Multiple hypotheses about the state of each object $i$ are kept by using a set of trackers 
\begin{equation}
  \mathcal{P}_i^t = \{(P_j^i, w_j^i) | j = 1, ..., n_i^t\},
\end{equation}
where $P_j^i$ is the $j$-th tracker of object $i$, $w_j^i$ is a temporal confidence score and $n_i^t$
represents the number of trackers for object $i$ at instant $t$. 
For this, $G^M$ is used to generate candidates on the most likely locations (Section~\ref{subsec:generating_candidates}).
Each candidate yields a new pair $(P_k^i, w^i_k = 0)$ which is then added to $\mathcal{P}_i^t$.
All trackers in $\mathcal{P}_i^t$ are then updated by applying their respective state dynamics.
After including the candidates in the set, the global tracking result is obtained by optimizing
a matching function between the model and the scene graphs (Section~\ref{subsec:optimizing_tracking}). Having found the
best trackers, the temporal scores of all of the candidates are updated (Section~\ref{subsec:updating_scores}).
The next sections detail each step.

\subsection{Attributed Relational Graph (ARG)}
\label{subsec:arg}

An ARG is a tuple
\begin{equation}
 G = (\mathcal{V}, \mathcal{E}, \mathcal{A}_{\mathcal{V}}, \mathcal{A}_{\mathcal{E}}),
\end{equation}
where $\mathcal{V} = \{v_i | i = 1, ...,  N_O\}$
represents a set of vertices (or nodes), $\mathcal{E} = \{e_{ij} = (v_i, v_j) | i,j \in \{1, ..., N_O\}\}$ is a set of directed edges (or arcs),
\ie $e_{ij} \neq e_{ji}$, and $\mathcal{A}_{\mathcal{V}}$ and $\mathcal{A}_{\mathcal{E}}$ are sets of attributes of
vertices and edges, respectively.

\begin{figure}[ht]
\begin{center}
   \includegraphics[width=0.8\linewidth]{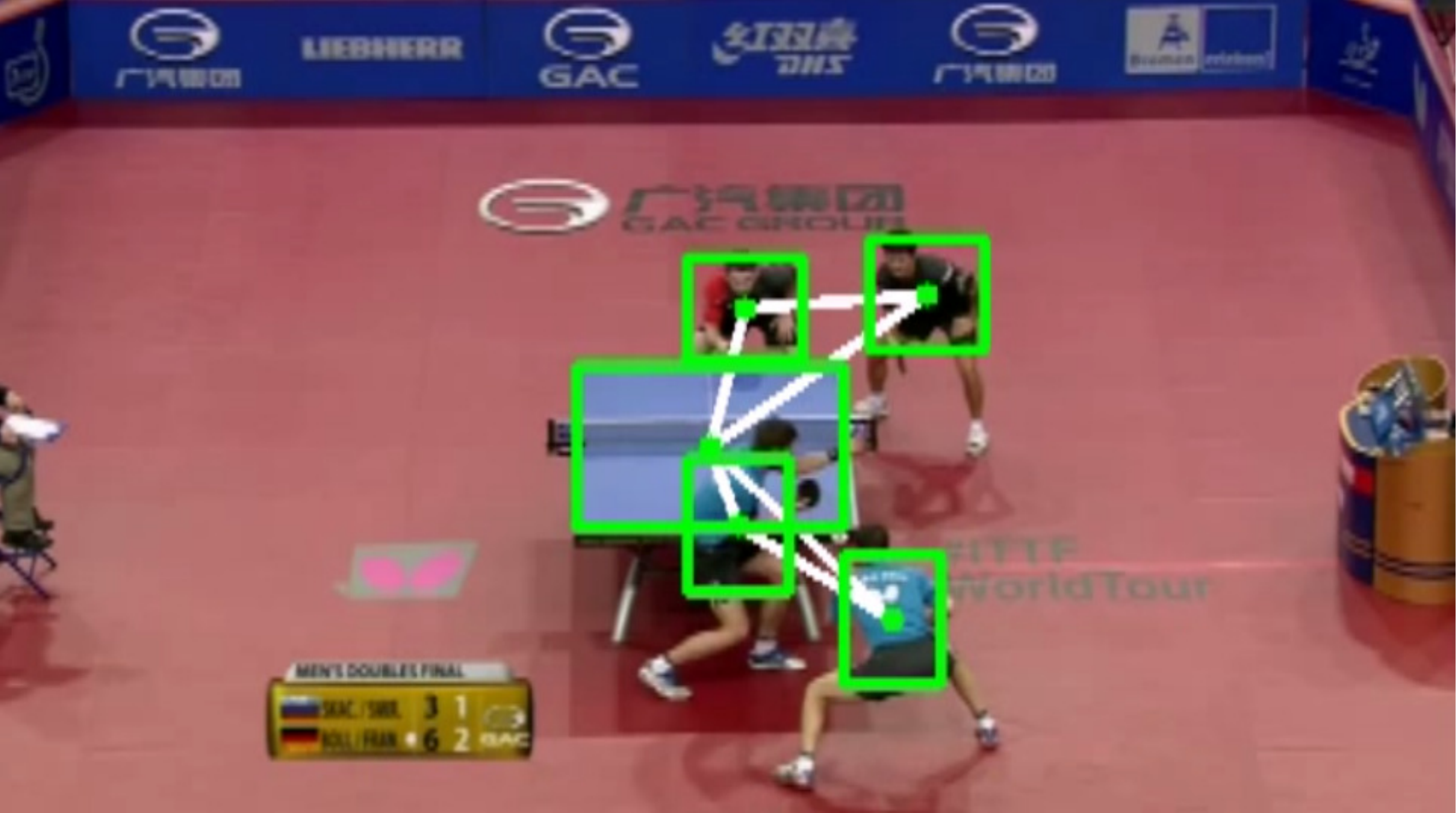}
\end{center}
   \caption{An example of a scene graph.}
\label{fig:graph}
\end{figure}

Each frame of the video (also referred to as scene) is represented by 
one or more ARGs. The vertices of $G$ are the tracked objects,
while the edges connect objects whose relations will be analyzed. The desired relations are expressed using
a binary adjacency matrix $M_A = (m_{ij})$ where $m_{ij} = 1$ if there is an edge from $v_i$ to $v_j$.
Figure~\ref{fig:graph} shows one possible scene graph generated from the following adjacency matrix:
\begin{equation}
 M_A = \left(
\begin{matrix}
  0 & 1 & 0 & 0 & 1 \\
  1 & 0 & 0 & 0 & 1 \\
  0 & 0 & 0 & 1 & 1 \\
  0 & 0 & 1 & 0 & 1 \\
  1 & 1 & 1 & 1 & 0 \\
\end{matrix}\right).
\label{eq:adjacency_matrix}
\end{equation}

Two different kinds of attributes are used: appearance and structural attributes.
Appearance attributes are related to each object and they are stored in $\mathcal{A}_{\mathcal{V}}$.
On the other hand, structural attributes represent relations among objects and thus constitute edge
attributes in $\mathcal{A}_{\mathcal{E}}$.

\subsubsection{Model graph}
\label{subsubsec:model_graph}

The model graph $G^M$ is an ARG, whose topology is obtained by means of an adjacency matrix $M_A$.
The choice of the topology depends on the problem, and it may be defined based on common sense
or expert knowledge. As a guideline, edges should be added: (1) between objects with
similar appearance that tend to cause inter-occlusion (in this way, the graph can be used to
handle the ambiguity problem after occlusion), (2) between pairs of objects that present
a clear positional relationship (thus allowing the generation of good candidates to search
in case of loss), and (3) to exploit the relation between moving objects and some
reference (stable) object.
The appearance attributes in $\mathcal{A}_{\mathcal{V}}^M$ are computed from annotations
provided in the first frame of the video. In our experiments,
the appearance is described by using color histograms as presented in~\cite{perez2002color}.
However, any other appearance descriptor could also be considered, like
HOG~\cite{dalal2005histograms} or SIFT~\cite{lowe2004distinctive}, and the proposed method
can be applied directly.

\begin{figure}[ht]
\begin{center}
   \includegraphics[width=0.8\linewidth]{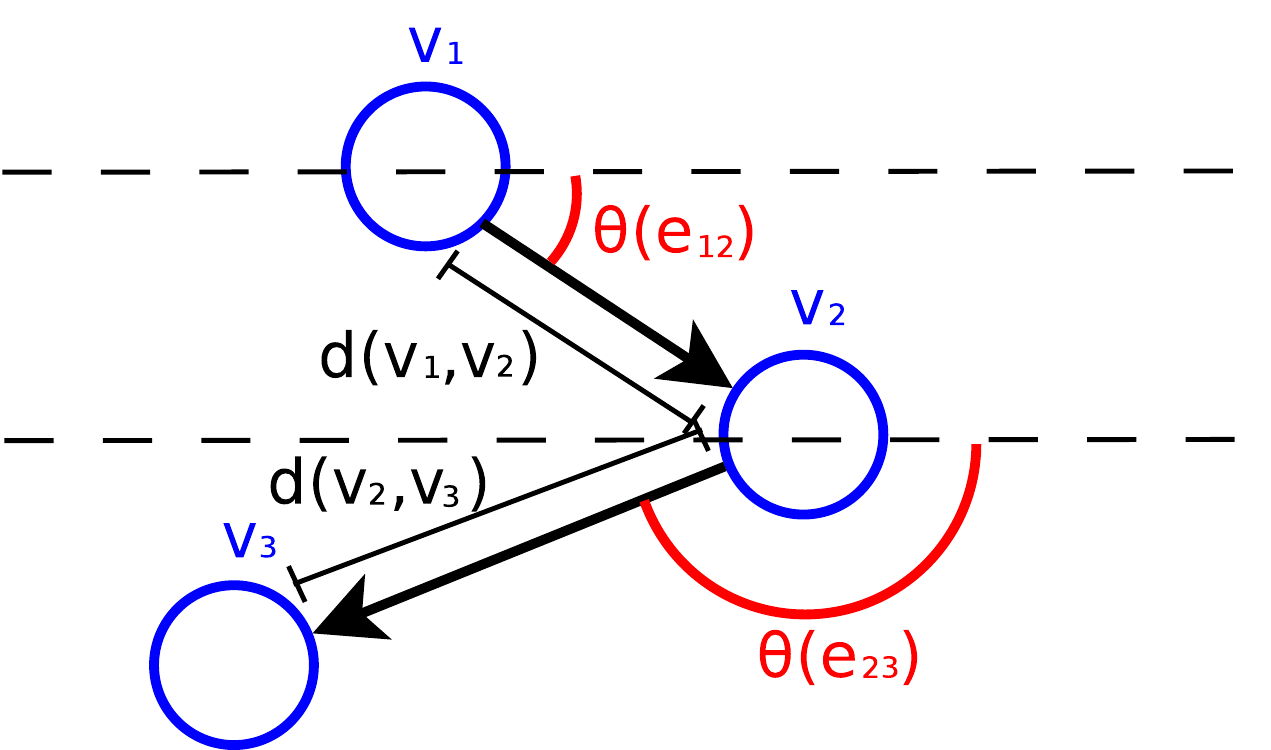}
\end{center}
   \caption{The structural attributes of the edges.}
\label{fig:arc_attribute}
\end{figure}

The initial set of attributes $\mathcal{A}_{\mathcal{E}}^M$ of $G^M$ also comes from the structure
observed from the annotations of the first frame. However, they are also constantly updated after
every frame (Section~\ref{subsec:updating_model}). Let $\delta$ be one of the structural attributes to be measured
(\eg the distance between two objects). The structural attributes of the model graph consist of the
probability density function (PDF) of $\delta$. Inspired by~\cite{cho2013learning}, the chosen set of
attributes is
\begin{equation}
  \Delta(i, j) = \{(\theta(e_{ij}), d(v_i, v_j))\},
\end{equation}
and the PDF is estimated by means of histograms $H_{\delta}$ ($H_{\theta}$ or $H_{d}$ in this case).
The function $\theta(e_{ij})$ represents the clockwise angle between the horizontal axis and the vector $\overrightarrow{v_iv_j}$,
and $d(v_i, v_j)$ is the Euclidean distance between the two vertices (Figure~\ref{fig:arc_attribute}).
In order to obtain more robustness, the distance is normalized by dividing it
by the width of the image. The PDFs are initialized
from the annotations obtained from the first frame. For this, we assume that these annotations
have the highest confidence value (equals to $1$) and we use the convolutional kernels explained in Section~\ref{subsec:updating_model}.

\begin{figure}[ht]
\begin{center}
   \includegraphics[width=0.8\linewidth]{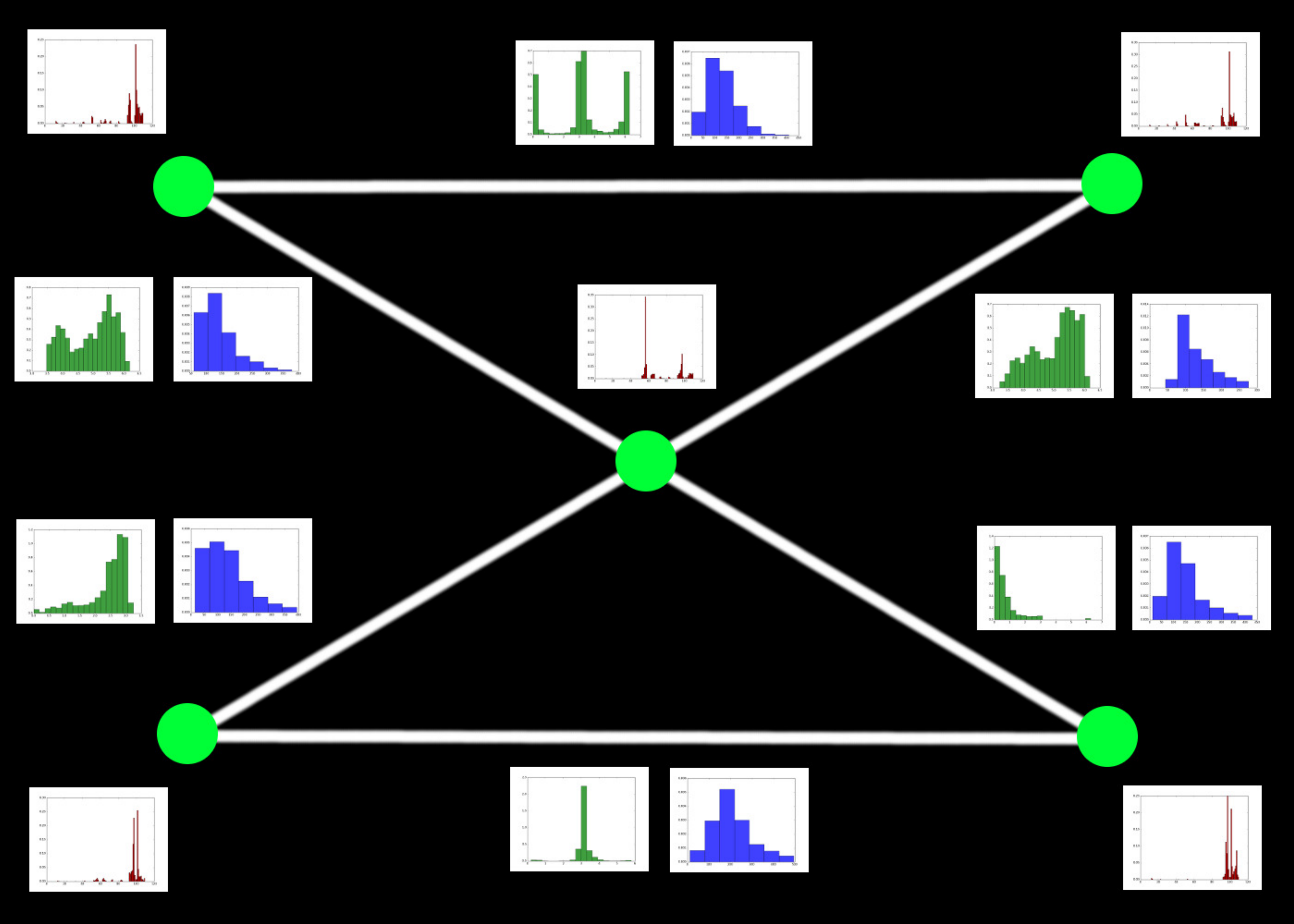}
\end{center}
   \caption{An example of a model graph with the learned attributes. The red histograms represent
   the attributes of the vertices (color model), while the green ones represent the angles, and the blue ones represent
   the distances.}
\label{fig:model_graph}
\end{figure}

Figure~\ref{fig:model_graph} shows one example of a graph and the learned histograms for each attribute.

\subsubsection{Scene graph}
\label{subsubsec:scene_graph}

The scene graph has the same topology as the model graph, and it is built at each frame as follows.
Each scene configuration $k$ is represented by a graph $G_k^S$ (some examples of different graph
configurations for one scene can be viewed in Figure~\ref{fig:multiple_graphs}).
A vertex $v_i \in \mathcal{V}^S$ of the scene graph $G_k^S$ is associated with one cloud of particles
$P_j^i$ for object $i$. Let us assume that the state of each object is composed of $|X| \geq 2$
components (in this paper the state is composed of exactly $2$ components).
Let $r(P_j^i) = (x_1^{i, j}, x_2^{i, j}, ..., x_{|X|}^{i, j})$ represent the final vector state
obtained from Equation~\ref{eq:particle_state} for the particle cloud $P_j^i$. 
In our experiments, $x_1^{i, j}$ and $x_2^{i, j}$ represent the 2D coordinates of the object in the image space,
and the position of $v_i$ is given by
\begin{equation}
  r_{P}(P_j^i) = (x_1^{i, j}, x_2^{i, j}),
\end{equation}
\ie $r_{P}(P_j^i)$ is a truncated version of $r(P_j^i)$ that only includes the spatial coordinates.

The edges are then built using the same matrix $M_A$ as in the training and in the model graph. 
However, recall that each object is tracked by a set of different trackers. Therefore, each
scene may be described by multiple graphs that have the same topology, which represent all the
possible combinations of different trackers (Figure~\ref{fig:multiple_graphs}
displays candidates that may be added to each tracker set). 

The set of structural attributes $\mathcal{A}^S$ of $G^S$ is not composed
of PDFs as in $G^M$, but of single values for each measurement $\delta$ extracted from the current frame
(\ie the observations of $\delta$). The attributes
of the vertices are the associated pairs $(P_j^i, w_j^i)$.

\subsection{Updating the model graph}
\label{subsec:updating_model}

In the beginning, for each attribute histogram $H_{\delta}$, the range $H_{\delta}[min, max]$ and the number of bins $bins(H_{\delta})$ must be specified.
The tracking results are then used to update the model at each frame. The structural measurements
extracted from the scene graph are used to cast a vote and update the model histograms.
However, in order to deal with uncertainty and create a more general model, the measurements
are first convolved as $\delta \ast k_C$ with a kernel $k_C$ that depends on the confidence of the tracking result.
The confidence is measured according to the likelihood of the particle filter estimation
(Equation~\ref{eq:particle_state}) as $\PP(\bm{o}_{t}^i | r(P^i))$.

The convolution kernel is chosen according to the confidence as follows:
\begin{equation}
  k_C = 
  \begin{cases}
    [0.3, 0.4, 0.3] & \text{if } \PP(\bm{o}_{t}^i | r(P^i)) > 0.7,\\
    [0.15, 0.2, 0.3, 0.2, 0.15] & \text{if } 0.3 < \PP(\bm{o}_{t}^i | r(P^i)) \leq 0.7,\\
    [0.1, 0.13, 0.17, 0.2, 0.17, 0.13, 0.1] & \text{otherwise.}\\
  \end{cases}
\end{equation}
Therefore, less confident results are more spread throughout the histogram range to reflect
the uncertainty of the observations.

\subsection{Generating new candidates}
\label{subsec:generating_candidates}

Besides for tracking evaluation, we propose using the structural information of $G^M$ to generate new
candidate positions for each tracked object. This step is important to improve overall tracking,
particularly to recover the target after a tracking failure occurs.

Since the attributes $\mathcal{A}_{\mathcal{E}}^M$ are all relative to the origin of each arc $e_{ij}$,
the position of $v_i$ must be known. Therefore, it is assumed that the trackers for every object will not
all fail at the same time. Good candidates can be generated by selecting the positions
given by the best trackers as origins. The candidate generation is controlled by using a matrix $M_C = (m_{ij})$,
where $m_{ij}$ indicates that, if object $i$ is used as reference, then it generates $m_{ij}$
candidates for object $j$.

\begin{figure}[ht]
\begin{center}
   \includegraphics[width=0.8\linewidth]{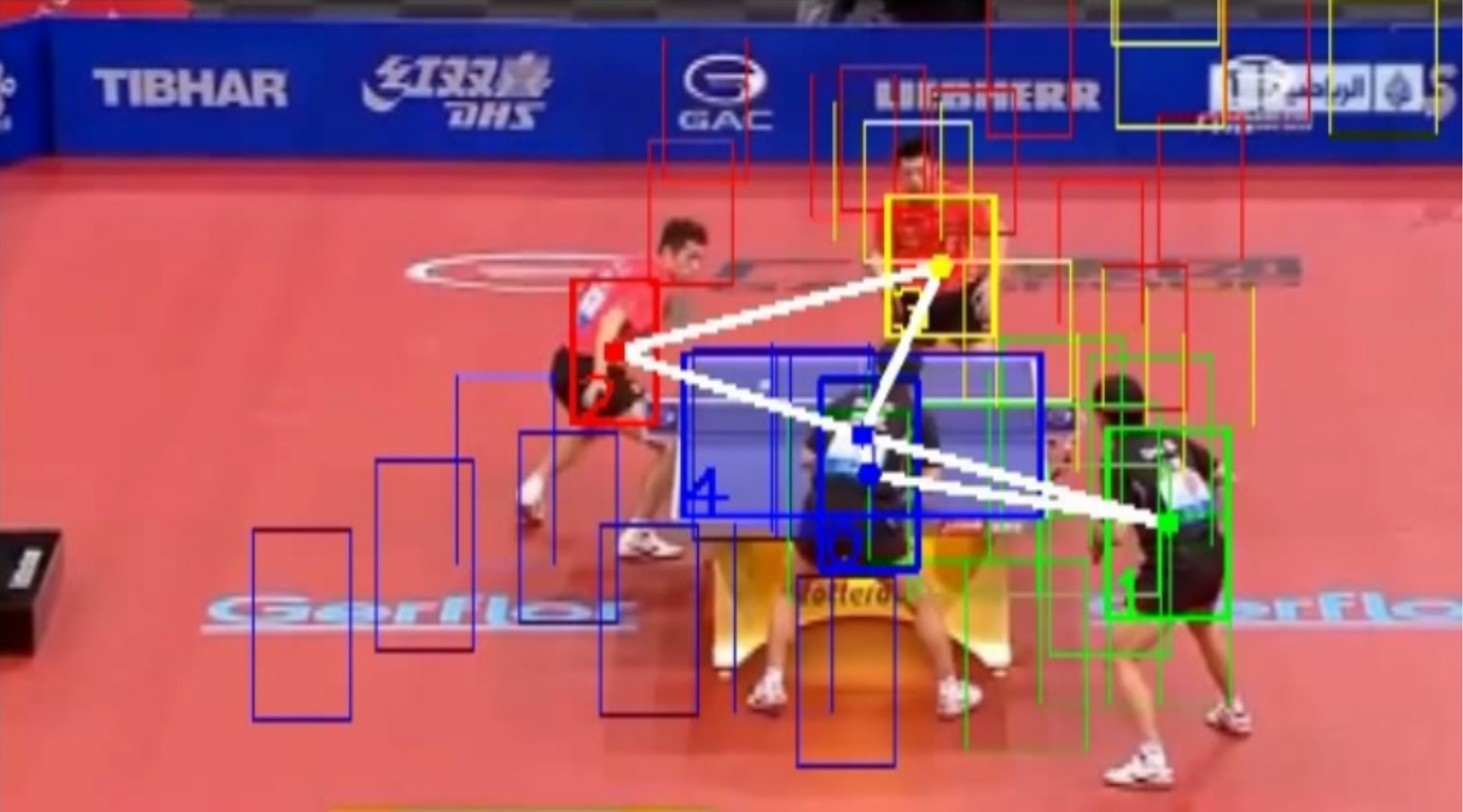}
\end{center}
   \caption{An example of candidates generated for one scene. Rectangles of the same color
   indicate that they belong to the same object.}
\label{fig:graph_candidates}
\end{figure}

Let $a_{e_{ij}}^M = \{H_{\theta}(\theta(e_{ij})), H_d(d(v_i, v_j))\}$ be the attributes
of an edge $e_{ij}$ from $G^M$. Candidates $k$ are generated for object $j$ as 
\begin{equation}
  (\hat{\theta}_k = \theta_k + u_{\theta}, \hat{d}_k = d_k + u_d)
\end{equation}
by simulating according to the distributions given by the histograms $\theta_k \sim H_{\theta}(\theta(e_{ij}))$ 
and $d_k \sim H_d(d(v_i, v_j))$, where $u_{\theta} \sim \mathcal{N}(0, \sigma_{\theta})$
and $u_d \sim \mathcal{N}(0, \sigma_d)$ are Gaussian noises.
Each candidate position is then obtained by 
\begin{equation}
  (v_i(x) + \hat{d}_k \cos(\hat{\theta}_k), v_i(y) + \hat{d}_k \sin(\hat{\theta}_k)).
\end{equation}
Figure~\ref{fig:graph_candidates} shows the candidates generated for each object.
The candidates are then used to generate new particle clouds $P_k^j$ which are inserted
in the set $\mathcal{P}_j^t$. The clouds are initialized by spreading the particles according to a
Gaussian distribution centered on the candidate position.

In order to avoid generating graphs using unreliable sources, the generated candidates are
filtered according to two criteria. Firstly, we check whether a candidate significantly overlaps another
older tracker. Therefore, if the overlap ratio is above a given threshold $\tau_O$, the candidate
is discarded. Secondly, we also compute the appearance score (see Section~\ref{subsubsec:appearance_score})
and remove candidates whose score is below $\tau_S$.

\subsection{Optimizing tracking using graphs}
\label{subsec:optimizing_tracking}

In this section, the method used for choosing the best trackers for each object is explained. Each tracker
receives a score based on a matching function according to its respective scene graph and the model. Later,
the best candidates are chosen by optimizing a global function over all the objects.

\subsubsection{Computing graph score}
\label{subsubsec:pf_score}

The score of a scene graph $G^S$ is obtained by summing over the temporal scores of all of its
vertices:
\begin{equation}
  s(G^S) = \sum_{i = 1}^{N_O}{w^i_{j_G}},
\end{equation}
where $w^i_{j_G}$ corresponds to the temporal weight of the candidate $j$ of object $i$ that is used
to create the graph $G^S$. This score measures the
reliability of the associated tracker  $P^i_{j_G}$ along time. This is done by computing
a weighted accumulation of instantaneous scores:
\begin{equation}
  (w^i_{j_G})^{t} = \rho_T (w^i_{j_G})^{t - 1} + f(i, G^S, G^M),
  \label{eq:instantaneous_score}
\end{equation}
where $t$ indicates the time, $\rho_T$ is a given constant and $f(i, G^S, G^M)$ is the instantaneous score function
for the vertex $v_i^S$, which is associated with $((P^i_{j_G})^t, (w^i_{j_G})^t)$. By doing so, trackers
which consistently perform well during longer periods of time have higher
scores than those that are only eventually good (usually incorrect trackers).

The instantaneous score is divided into four parts:
\begin{multline}
  f(i, G^S, G^M) = \rho_A \phi_A(i, G^S) + \rho_S \phi_S(i, G^S, G^M) - \\ \rho_O \phi_O(i, G^S) - (1 - \rho_F - \rho_S - \rho_O) \phi_C(i, G^S),
  \label{eq:score_function}
\end{multline}
where $\rho_A, \rho_S$ and $\rho_O$ are given weighting factors such that $\rho_F + \rho_S + \rho_O \leq 1$.
The functions $\phi_A, \phi_S, \phi_O$ and $\phi_C$ are the appearance, structural, overlapping and changing
score functions of $v_i$, respectively. Notice that the last two functions act as penalty terms and thus,
they decrease the score. These functions will be further explained in the following sections.

\subsubsection{Appearance score}
\label{subsubsec:appearance_score}

The appearance score is actually the confidence of the particle cloud $P_i$ associated with
object $i$ (vertex $v_i$) as in Equation~\ref{eq:particle_score}. Hence, it is set to
\begin{equation}
  \phi_A(i, G^S) = \zeta(P_i).
\end{equation}

The confidence score depends on the weights of the particles, which
are based on the likelihood $\PP(\bm{o}_{t} | \bm{x}_{t}^i)$.
The distribution is computed in the same way as in~\cite{erdem2012fragments}, 
using the Bhattacharyya distance $d_B$:
\begin{equation}
  \PP(\bm{o}_{t} | \bm{x}_{t}^i) = \exp\left(-\frac{d_B(H^M, H^S)^2}{2 \sigma^2}\right),
\end{equation}
where $H^M$ and $H^S$ are histograms, normalized to sum to one, of the model and the scene, respectively and
\begin{equation}
  d_B(H^M, H^S) = \sqrt{1 - \sum_j {\sqrt{H^M(j) H^S(j)}}},
  \label{eq:bhattacharyya}
\end{equation}
where $H(j)$ is the $j$-th bin of histogram $H$.

\subsubsection{Structural score}
\label{subsubsec:structural_score}

Let $\bm{m_i}$ be a vector representing the line $i$ from the adjacency matrix $M_A$,
\ie corresponding to object $i$. Let also
$\bm{\theta_i^S} = (H_{\theta}^M(\theta^S(e_{ij})))_{j = 1}^{N_O}$ and 
$\bm{d_i^S} = (H_d^M(d^S(v_i, v_j)))_{j = 1}^{N_O}$ be the vectors of the likelihoods of each
structure measurement coming from the scene according to the model histograms. The structure score is computed using the dot product:
\begin{equation}
  \phi_S(i, G^S, G^M) = \frac{1}{2 \|\bm{m_i}\|_1} \bm{m_i} \cdot \bm{\theta_i^S} + \bm{m_i} \cdot \bm{d_i^S},
\end{equation}
where $\|\bm{m_i}\|_1$ is the $L_1$ norm of $\bm{m_i}$. In other words, this score
corresponds to the average of the attributes of the edges originating from $v_i$.

\subsubsection{Overlapping score}
\label{subsubsec:overlapping_score}

The overlapping score is used to penalize configurations with a high intersection
between objects of similar appearance. This is done to facilitate the recovery after
one or more trackers with similar models lose the target in case of temporary occlusion.
Let $\mathcal{B}^i$ be the set of pixels inside the bounding box of particle cloud $P^i$
and $H^i$ the color histogram of $\mathcal{B}^i$. The overlapping score
is obtained by:
\begin{equation}
  \phi_O(i, G^S) = \sum_{j = 1, j \neq i}^{N_O}{\left[(1 - d_B(H^i, H^j)) \frac{|\mathcal{B}^i \cap \mathcal{B}^j|}{|\mathcal{B}^i|}\right]},
\end{equation}
where $d_B(H^i, H^j)$ is the Bhattacharyya distance defined in Equation~\ref{eq:bhattacharyya}.

\subsubsection{Changing score}
\label{subsubsec:changing_score}

This function is defined to penalize the change of trackers between consecutive frames.
The reason is to reinforce the assumption of smooth movement and, therefore, to keep the same
tracker for as long as possible. This score is obtained from a simple indicative function that
checks whether the tracker chosen in the last frame is being used in the current frame:
\begin{equation}
  \phi_C(i, G^S) = 
  \begin{cases}
    0 & \text{if tracker $j$ for object $i$ from instant $t - 1$ is kept,}\\
    1 & \text{if a new candidate $k$ replaces the old tracker $j$.}
  \end{cases}
\end{equation}

\subsubsection{Choosing the best scene graph}
\label{subsubsec:choosing_scene_graphs}

\begin{figure}[ht]
\begin{center}
   \subfloat{\includegraphics[width=0.4\linewidth]{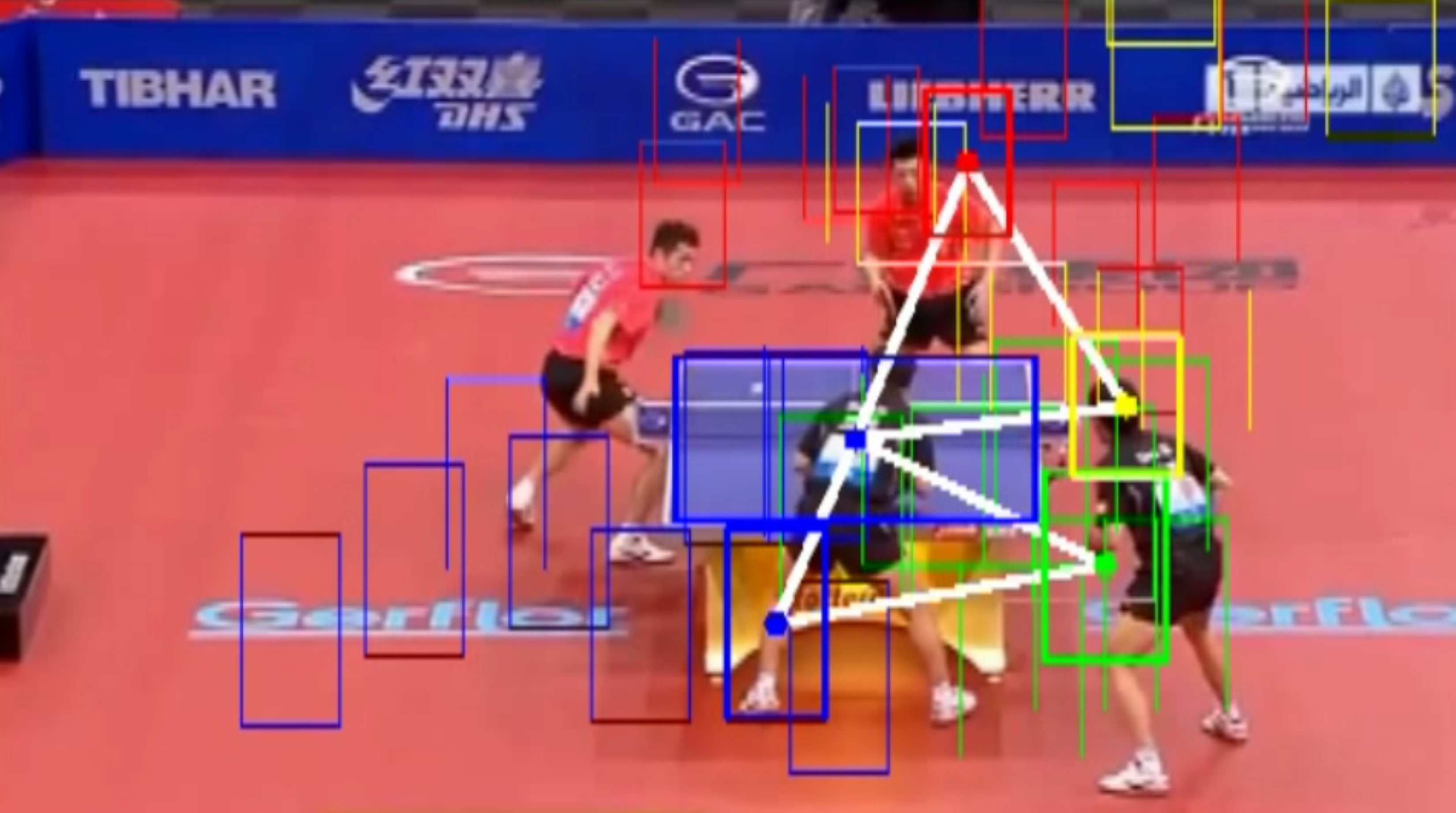}}
   \hfil
   \subfloat{\includegraphics[width=0.4\linewidth]{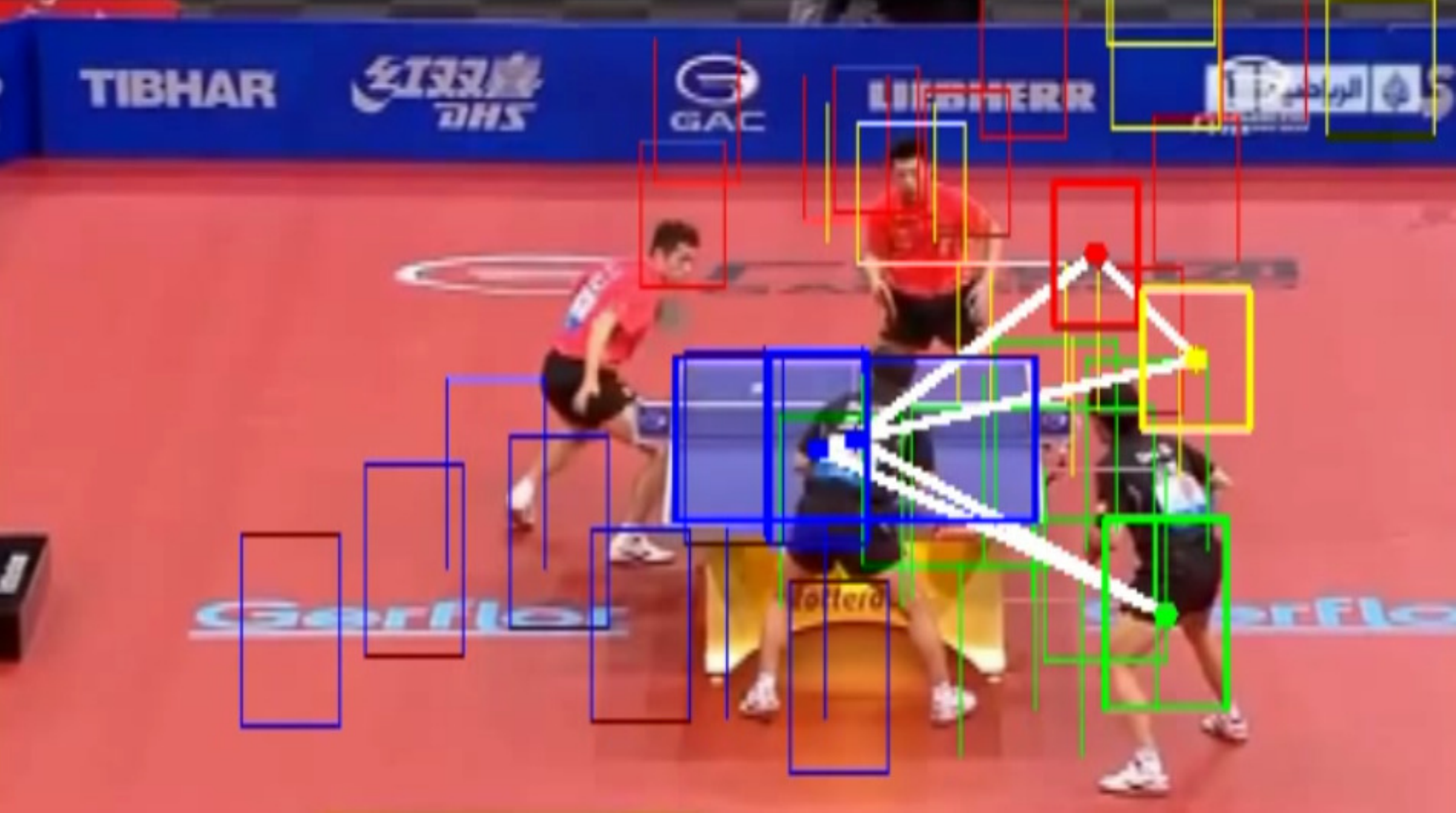}}
   \\
   \subfloat{\includegraphics[width=0.4\linewidth]{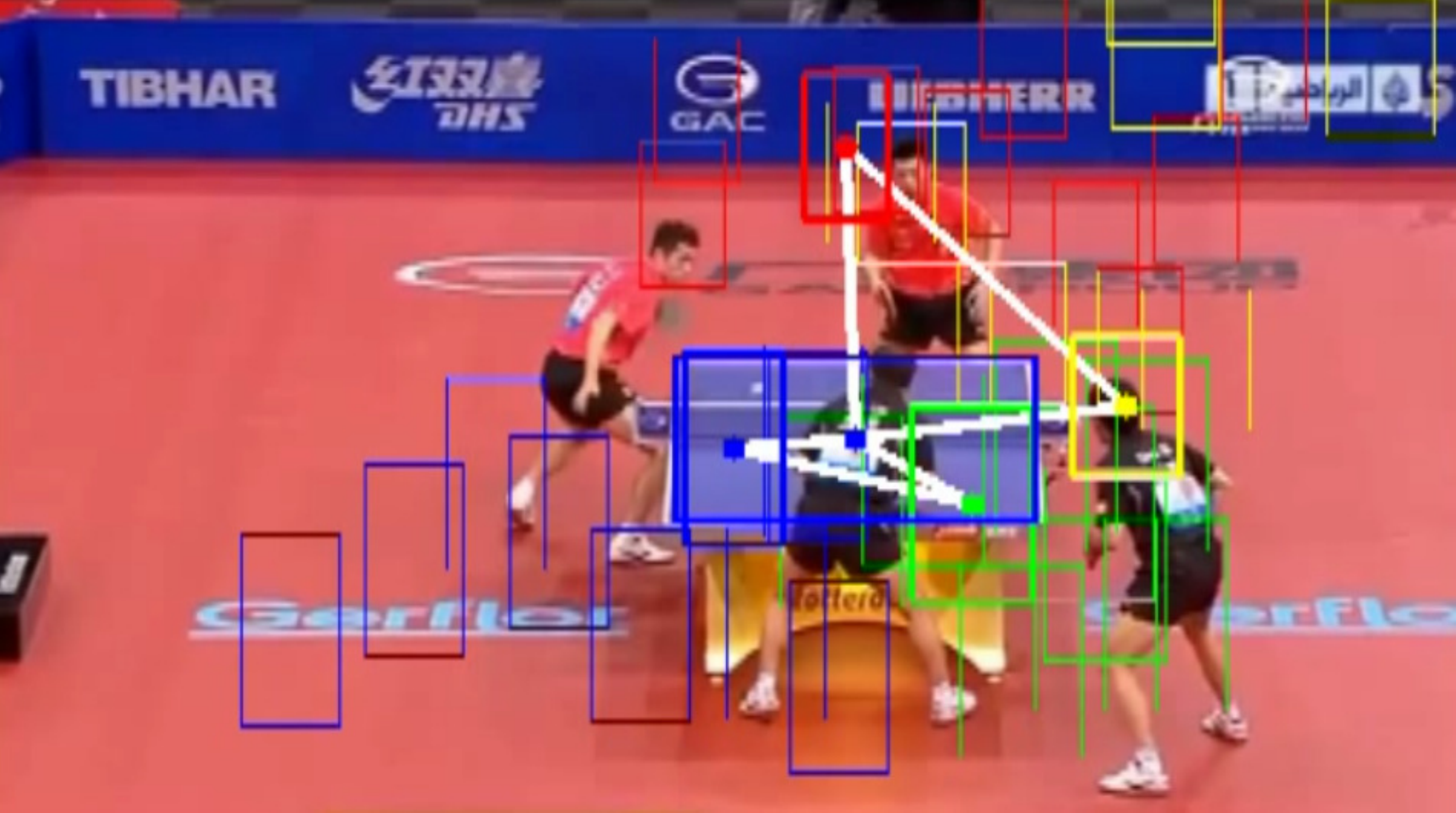}}
   \hfil
   \subfloat{\includegraphics[width=0.4\linewidth]{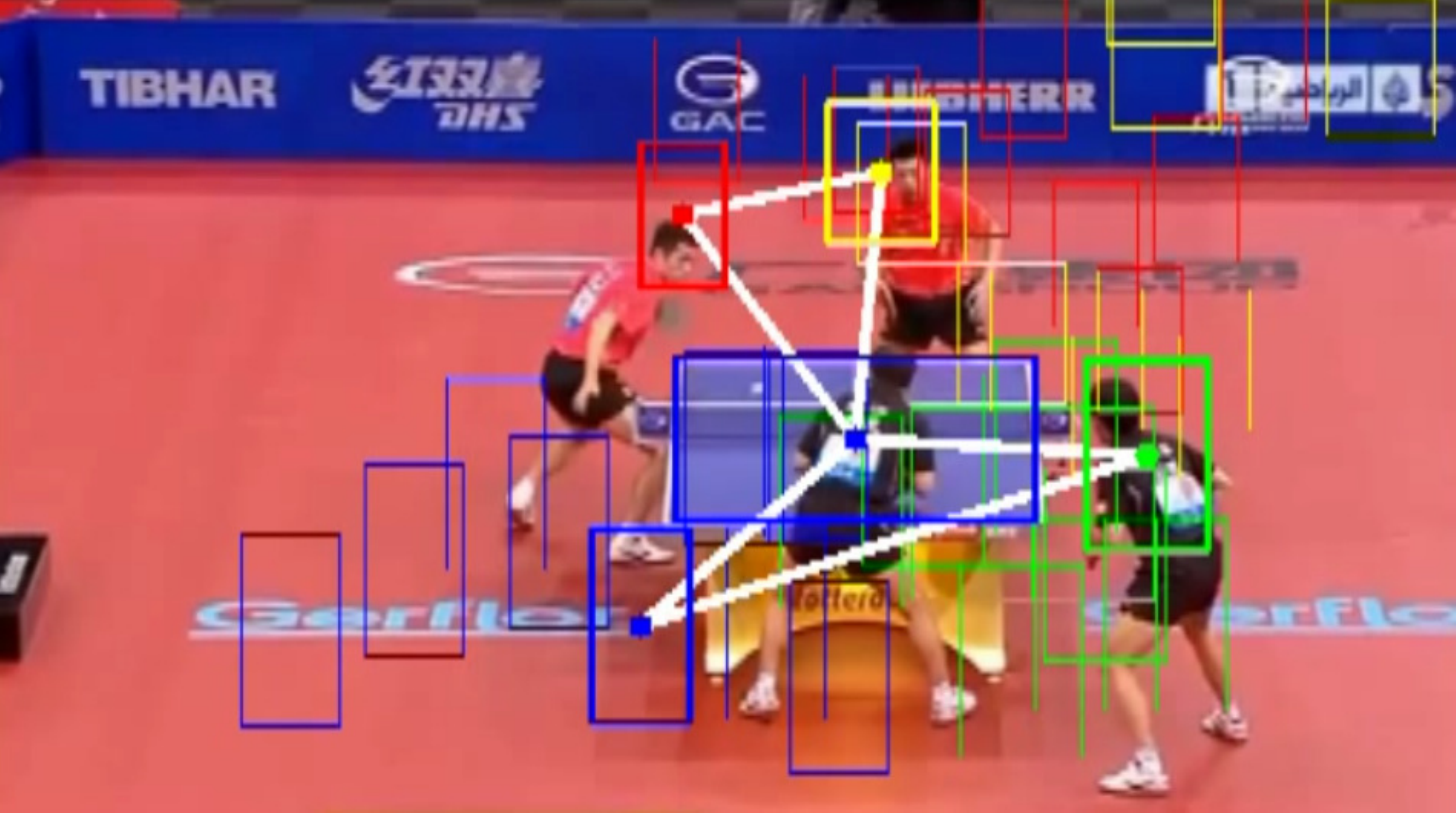}}
\end{center}
   \caption{Some examples of candidate graphs to be analyzed.}
\label{fig:multiple_graphs}
\end{figure}

The best trackers are selected by building the scene graphs $G_k^S$ and computing
the scores explained before. Figure~\ref{fig:multiple_graphs} shows possible
graphs that can be generated from some given candidates. Therefore, one option would be to build all possible graphs and to find
the one which maximizes the overall score for every tracker. However, this approach was not chosen
because the number of graph combinations
is usually very large and unfeasible to be processed in real time.
Instead, it was chosen an iteratively greedy approach
that fixes the vertices for all objects except one and optimizes the score for one object at a time.

Let $\mathcal{V}_*^S = \{v(P_*^i) | i = 1, ..., N_O\}$ be the set of vertices with
the best scores found at a certain iteration step and whose graph score is $s(G_*^S)$.
The initialization procedure of this set will be discussed later. Assume that the optimization
is being performed for object $j$ and we are testing its $l$-th candidate. We fix the
set of vertices $\mathcal{V}_{j, l}^S = \mathcal{V}_*^S \cup \{v(P_l^j)\} \setminus \{v(P_*^j)\}$
and compute the score of the associated graph $G_{j, l}^S$. If $s(G_{j, l}^S) > s(G_*^S)$, then we change
the set of best vertices as $\mathcal{V}_*^S = \mathcal{V}_{j, l}^S$. This step is repeated
for one object at a time and all of its candidates are tested. After all of the objects
are evaluated, we repeat the process again until $\mathcal{V}_*^S$ does not change
during one whole iteration or if the iteration limit $\tau_I$ has been reached.
Since the score function is upper bounded, this procedure is guaranteed to converge
if enough time is available. However, it does not always find the global maximum,
sometimes being stuck at a local peak. In order to try to overcome this, we reinitialize
the optimization procedure $N_{RI}$ times and take the best results of all runs. We also
employ two different heuristics for choosing the initial vertices and the order of testing
the objects.

The first procedure is the score-guided optimization. For this step, the initial set of
vertices $\mathcal{V}_*^S$ is chosen as the ones used in the last frame.
Let $\{(P_*^i, w_*^i) | i = 1, ..., N_O\}$
be the set of the best trackers of each object, \ie 
\begin{equation}
  (P_*^i, w_*^i) = \argmax_{(P_j^i, w_j^i) \in \mathcal{P}_i^t}{w_j^i}.
  \label{eq:optimal_tracker}
\end{equation}
A sequence is created by sorting $w_*^i$ in ascending order and processing the objects $i$ one by one
according to this sequence. The rationale is that, since all the other vertices will be fixed, it is better
to let the worst tracker vary first in order to have good references for the resulting graph.
The second heuristic is a completely random run. In this case, both the initial set of vertices and
the order of testing objects are chosen randomly.
The final set of vertices is chosen by running the score-guided optimization once and repeating the
random procedure $N_{RI}$ times to find the set of trackers that yields the best global score.

\subsection{Updating trackers} 
\label{subsec:updating_scores}

When running the graph optimization, temporal scores are computed but not stored
as the final scores of each tracker. The reason is that, since they depend on the whole
graph, they should not be updated until the best graph $G_*^S$ has been found. Once the graph
is decided, the temporal scores of all candidates can be updated by following the same
greedy procedure as explained in Section~\ref{subsubsec:choosing_scene_graphs} using
$\mathcal{V}_*^S$ as the initial set of vertices.

We also want to remove trackers whose scores are too low. Therefore, whenever the temporal
score of a tracker falls below a threshold $\tau_R$, the tracker is removed. The only exception
is when a given object $i$ has only one tracker in its set. In this case, the tracker is
kept despite its score.

\section{Experimental results}
\label{sec:results}

The software was developed in Python with the OpenCV library\footnote{\url{http://opencv.org/}}.
As explained in Section~\ref{subsubsec:particle_models}, each object is individually tracked using particle filters with
color histograms as proposed in~\cite{perez2002color}. In our experiments, we observed that the
tracking speed was between 3 and 4 frames per second (FPS) on the table tennis and badminton videos,
and around $1.5$ FPS on the volleyball sequences, using a machine equipped with an Intel Core i5 processor
and 8GB RAM. Notice that the current code is neither parallelized
nor utilizes the GPU. Therefore, there is still room for significant improvement in the performance.
Specifically, the particle filter and graph evaluations would greatly benefit from parallelization, since
all the instances could be processed simultaneously. The source code is publicly available for downloading
and testing\footnote{\url{https://github.com/henriquem87/structured-graph-tracker}}.

\subsection{Evaluation metrics}
\label{subsec:tracking_metrics}

We use the widely adopted CLEAR-MOT metrics~\cite{bernardin2008evaluating}, whose most important
results are the MOTP and MOTA. In this section, we will briefly explain each of them.

MOTP is an acronym for Multiple Object Tracking Precision, while MOTA is the Accuracy.
The former metric represents the capacity of the tracker to produce results which are close to the
ground truth, even if the identity of the objects are swapped. The latter, on the other hand,
measures how well the tracker is able to assign tracks to the correct object, without considering
how close it actually is from the correct position.

Let $d^i_t$ be the distance between
the estimated result and the ground truth for object $i$ at time $t$ and $c_t$ the number of
matches found, then we can compute:
\begin{equation}
  MOTP = \frac{\sum_{i, t}{d^i_t}}{\sum_{t}{c_t}}.
\end{equation}
In this paper, the distance $d^i_t$ is actually the overlapping between
the estimated bounding box and the ground truth. Therefore, higher values of MOTP indicate better
results.

For the MOTA, let $g_t$ be the number of objects that exist at instant $t$. Let also $m_t$, $fp_t$ and $mme_t$
be the number of misses, false positives, and mismatches, respectively. Then, the metric can be obtained by:
\begin{equation}
  MOTA = 1 - \frac{\sum_{t}{(m_t + fp_t + mme_t)}}{\sum_{t}{g_t}}.
\end{equation}

We shall call MOT General or $MOTG$, a metric that corresponds to the average of $MOTP$ and $MOTA$:
\begin{equation}
  MOTG = \frac{MOTP + MOTA}{2}.
\end{equation}

We also evaluate the results using the following additional metrics obtained from CLEAR-MOT:
\begin{itemize}
  \item $IDSW$: number of ID switches, \ie number of times the current association contradicts the one from the previous frame;
  \item $TPrate$: rate of true positives, \ie total number of true positives divided by $\sum_{t}{g_t}$;
  \item $FPrate$: rate of false positives.
\end{itemize}

\subsection{Datasets}

The three datasets used for testing the tracking framework are described below. The proposed method is designed
for handling videos where some structural properties can be observed. Among the many existing broadcast sports videos,
we found out that sports like volleyball, table tennis, and badminton exhibit a clear structure that could be exploited.
These videos also present some challenging conditions, such as similar appearance, occlusion, and rapid motion, which may
cause even state-of-the-art methods to struggle.
To the best of our knowledge, there is no public sports dataset that presents the desired characteristics for evaluating
this method~\cite{dubuisson2016survey}. Therefore, we decided to collect appropriate videos to verify
if the proposed approach could contribute for handling tracking problems in this class.

\subsubsection{Youtube table tennis}

\begin{figure}[!ht]
\begin{center}
   \subfloat{\includegraphics[width=0.3\linewidth]{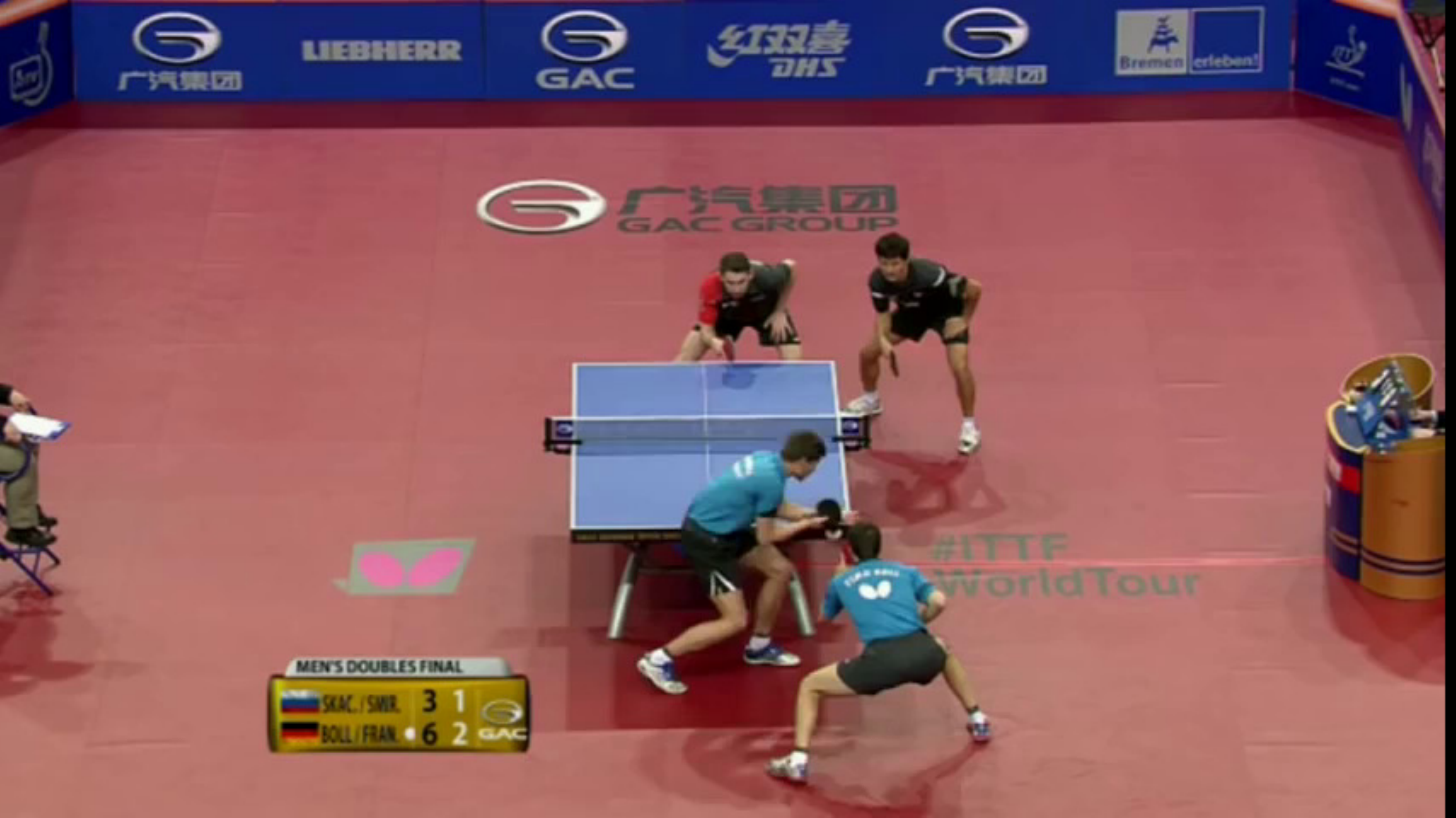}}
   \hfil
   \subfloat{\includegraphics[width=0.3\linewidth]{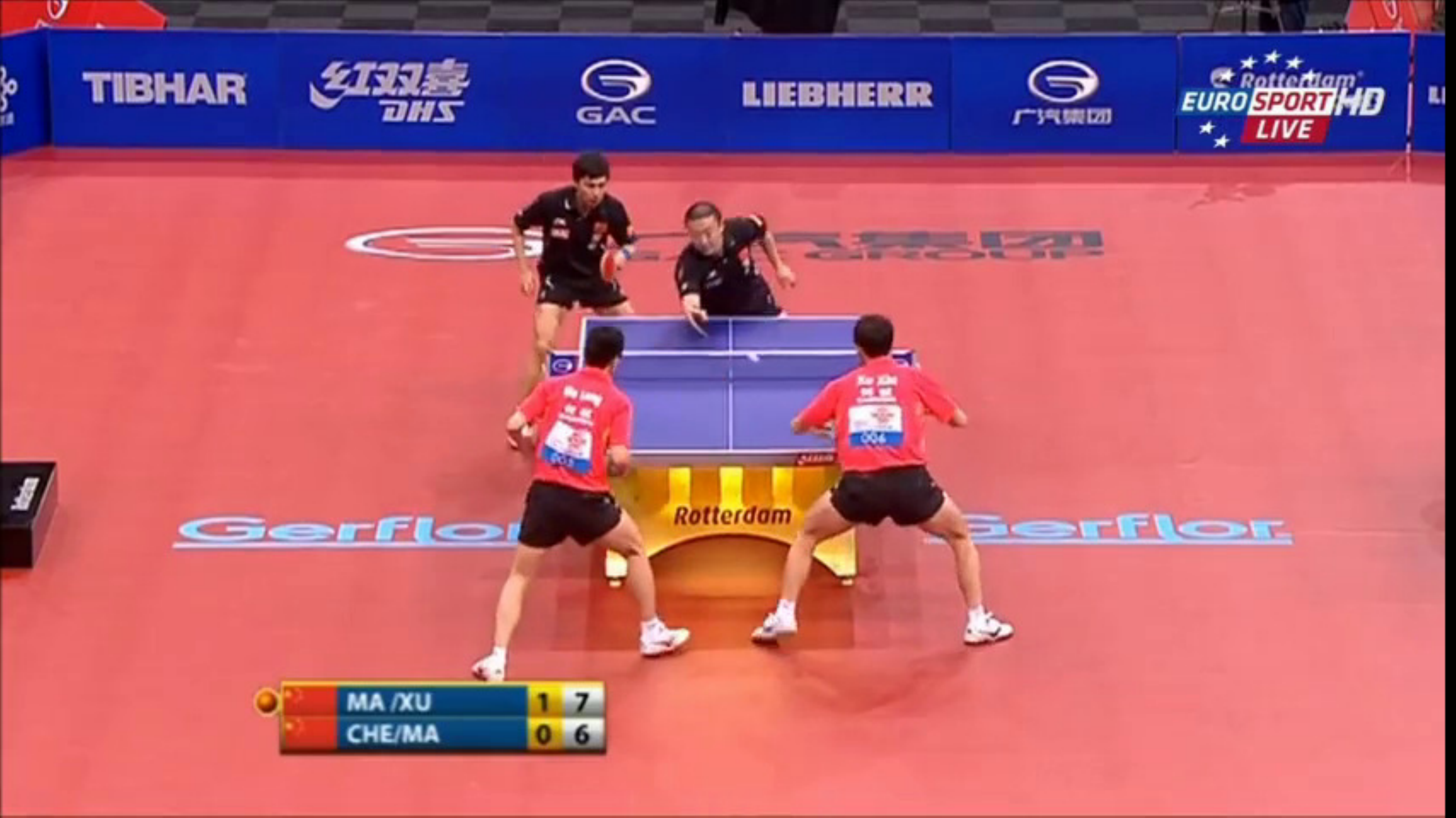}}
   \hfil
   \subfloat{\includegraphics[width=0.3\linewidth]{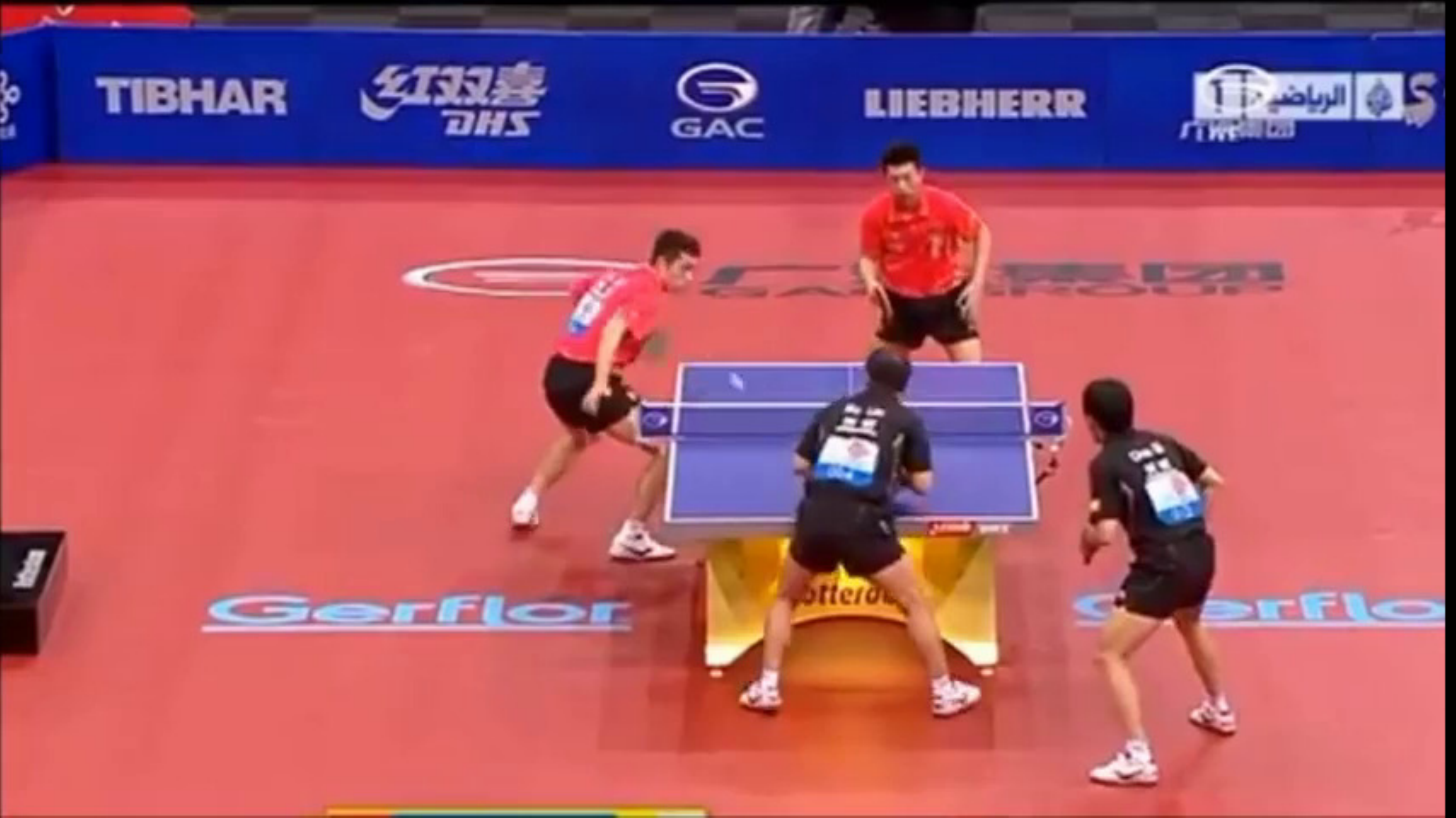}}
   \\
   \subfloat{\includegraphics[width=0.3\linewidth]{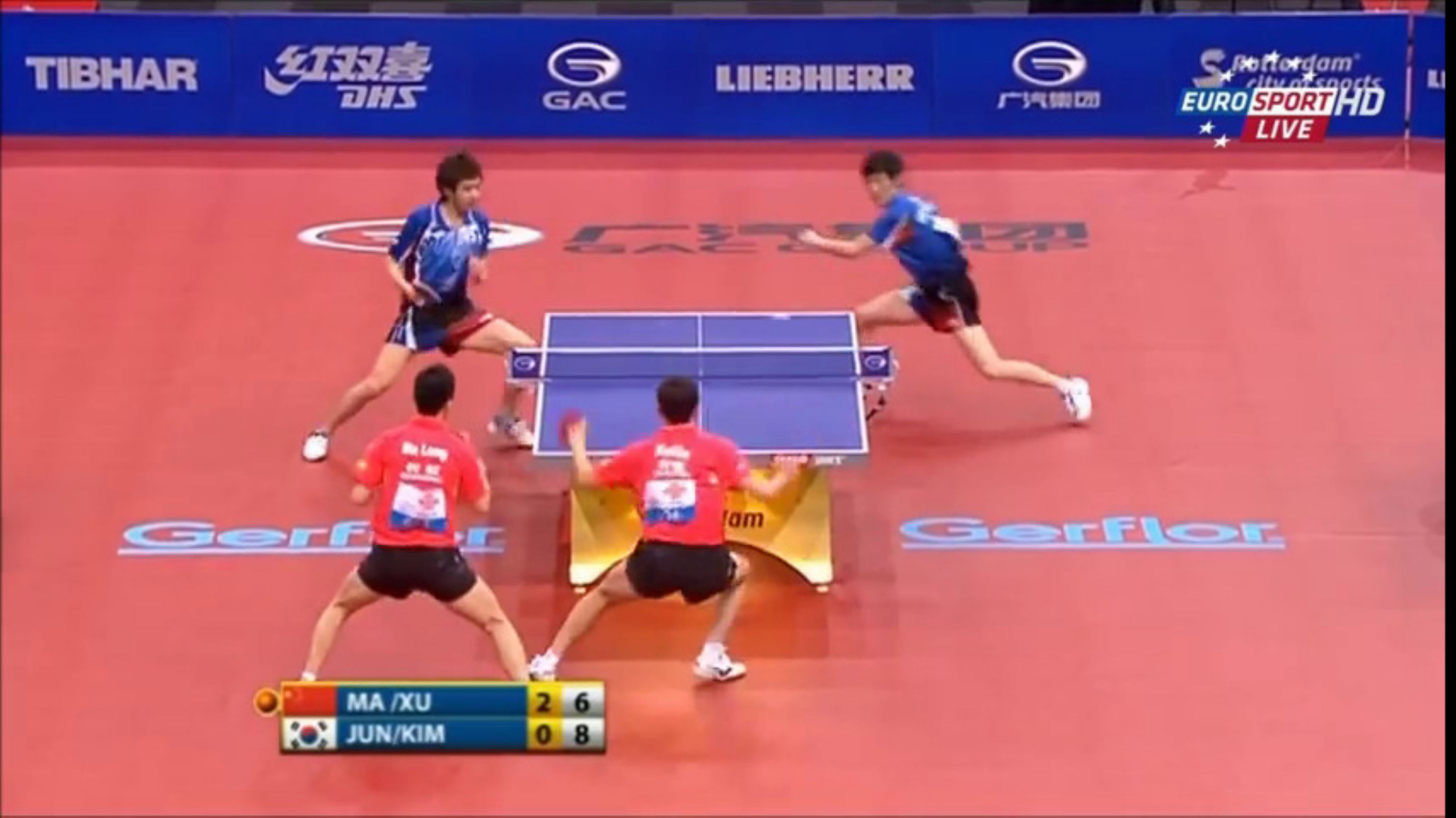}}
   \hfil
   \subfloat{\includegraphics[width=0.3\linewidth]{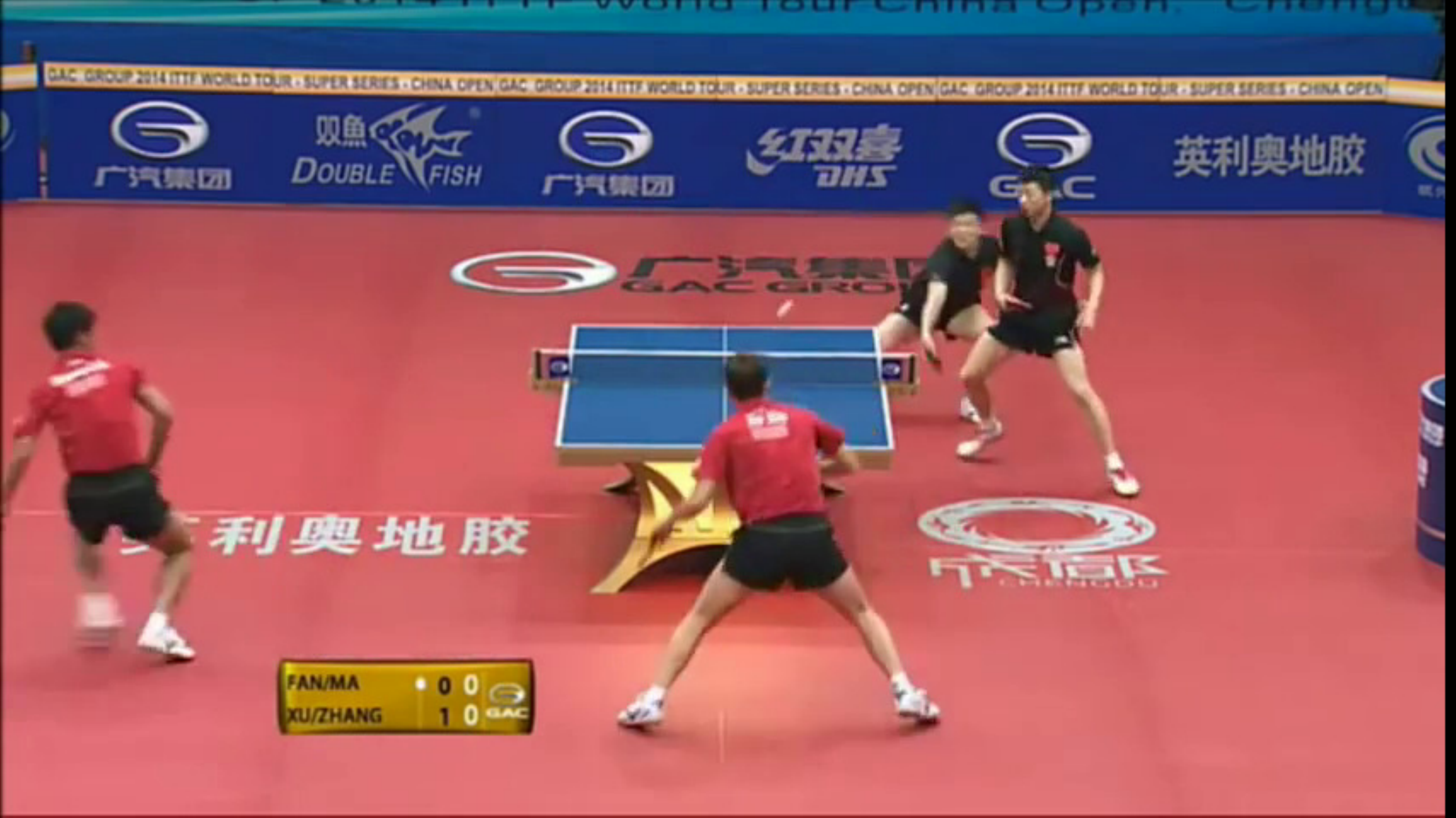}}
   \hfil
   \subfloat{\includegraphics[width=0.3\linewidth]{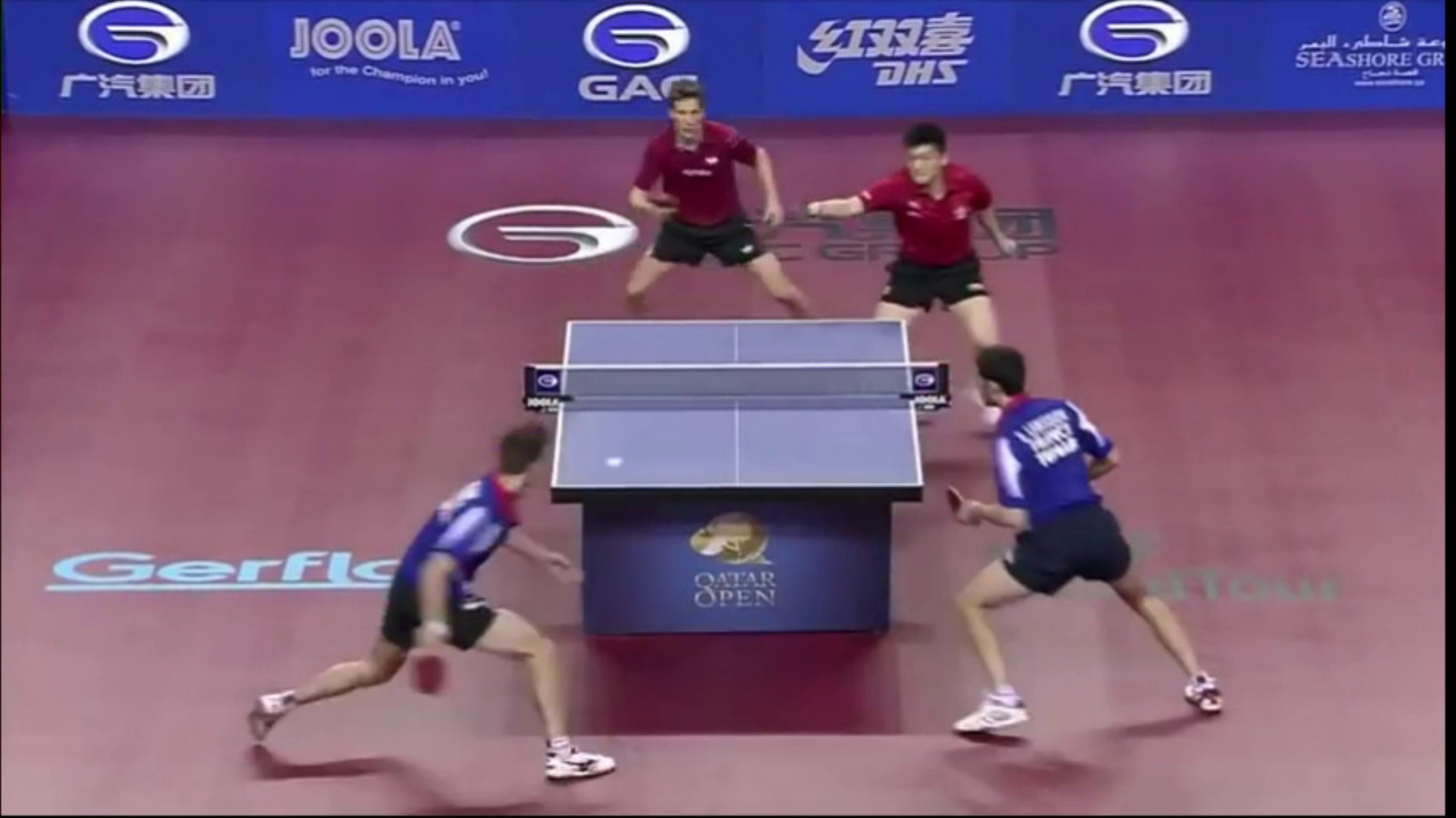}}
\end{center}
   \caption{Sample frames from the Youtube table tennis dataset.}
\label{fig:tt_dataset}
\end{figure}

This dataset is composed of 6 videos containing 6737 frames in total. All the videos are of doubles matches of competitive
table tennis collected from Youtube. Figure~\ref{fig:tt_dataset} shows some sample frames from this dataset. The videos were edited to remove 
unrelated scenes (\eg preparation stage, crowd) and then manually annotated with bounding boxes for ground truth.
The videos are encoded at resolutions varying from $640 \times 360$ to $854 \times 480$ pixels in each frame and at 30FPS.

\subsubsection{ACASVA}

\begin{figure}[!ht]
\begin{center}
   \subfloat{\includegraphics[width=0.3\linewidth]{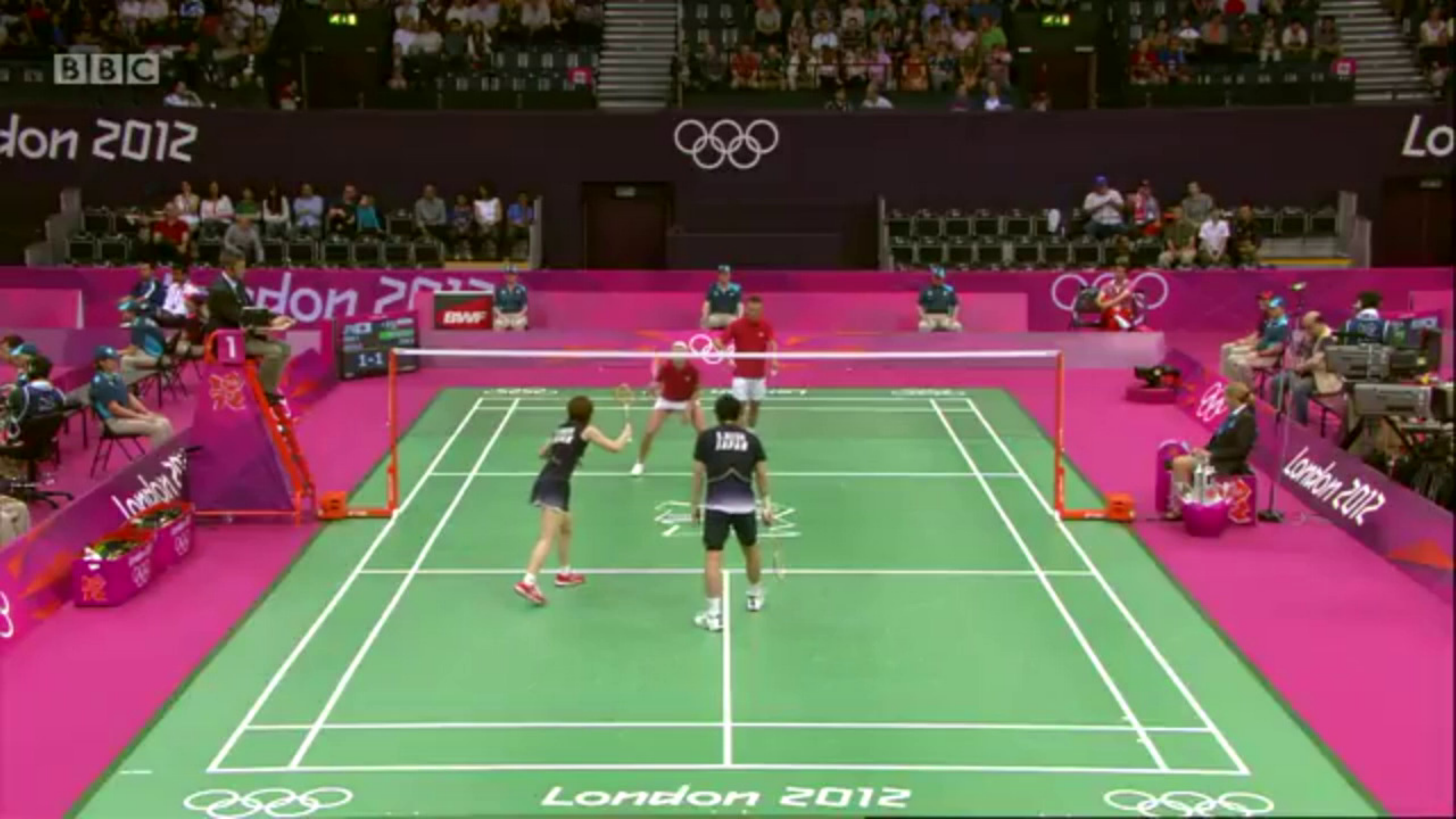}}
   \hfil
   \subfloat{\includegraphics[width=0.3\linewidth]{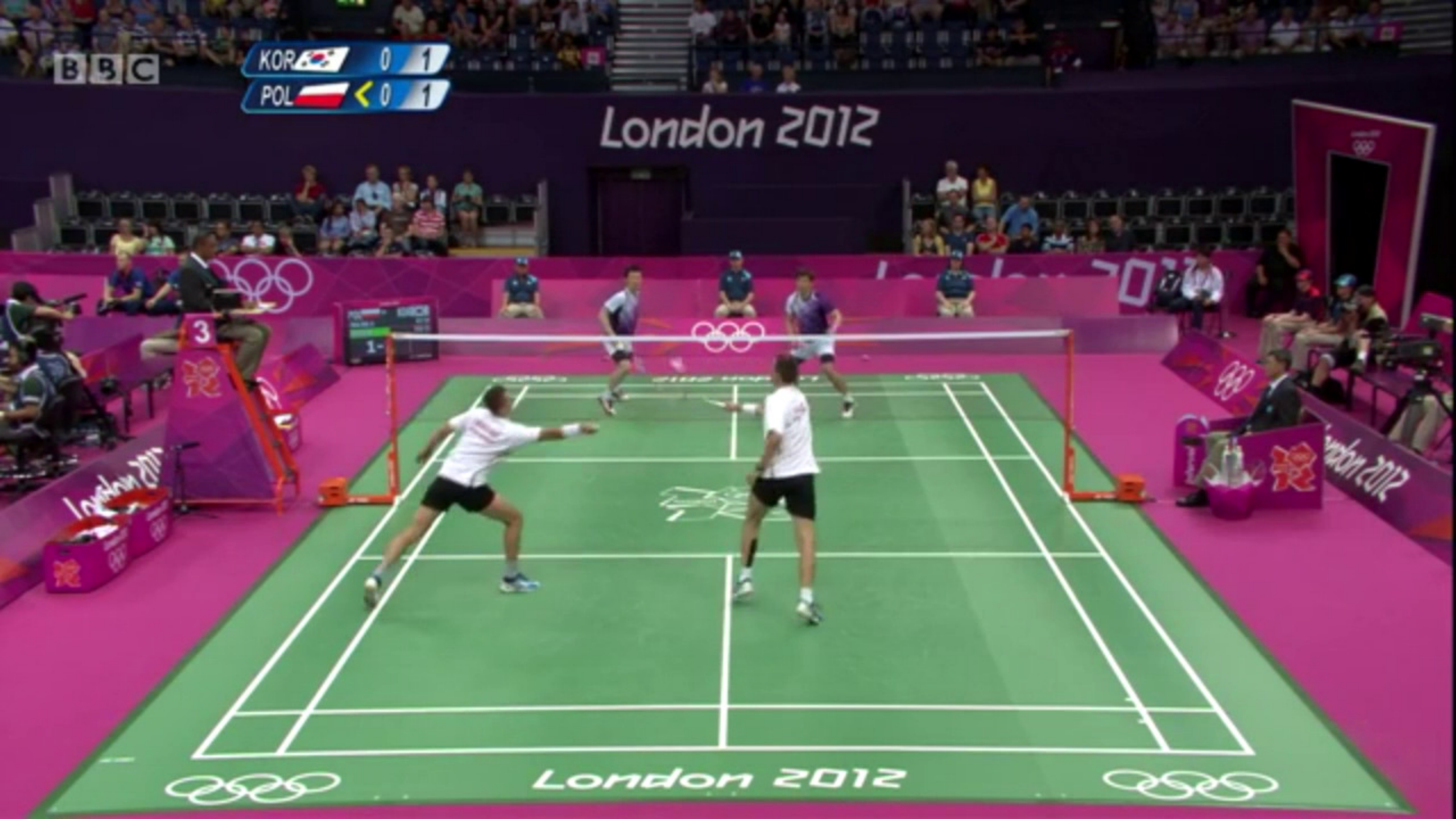}}
   \hfil
   \subfloat{\includegraphics[width=0.3\linewidth]{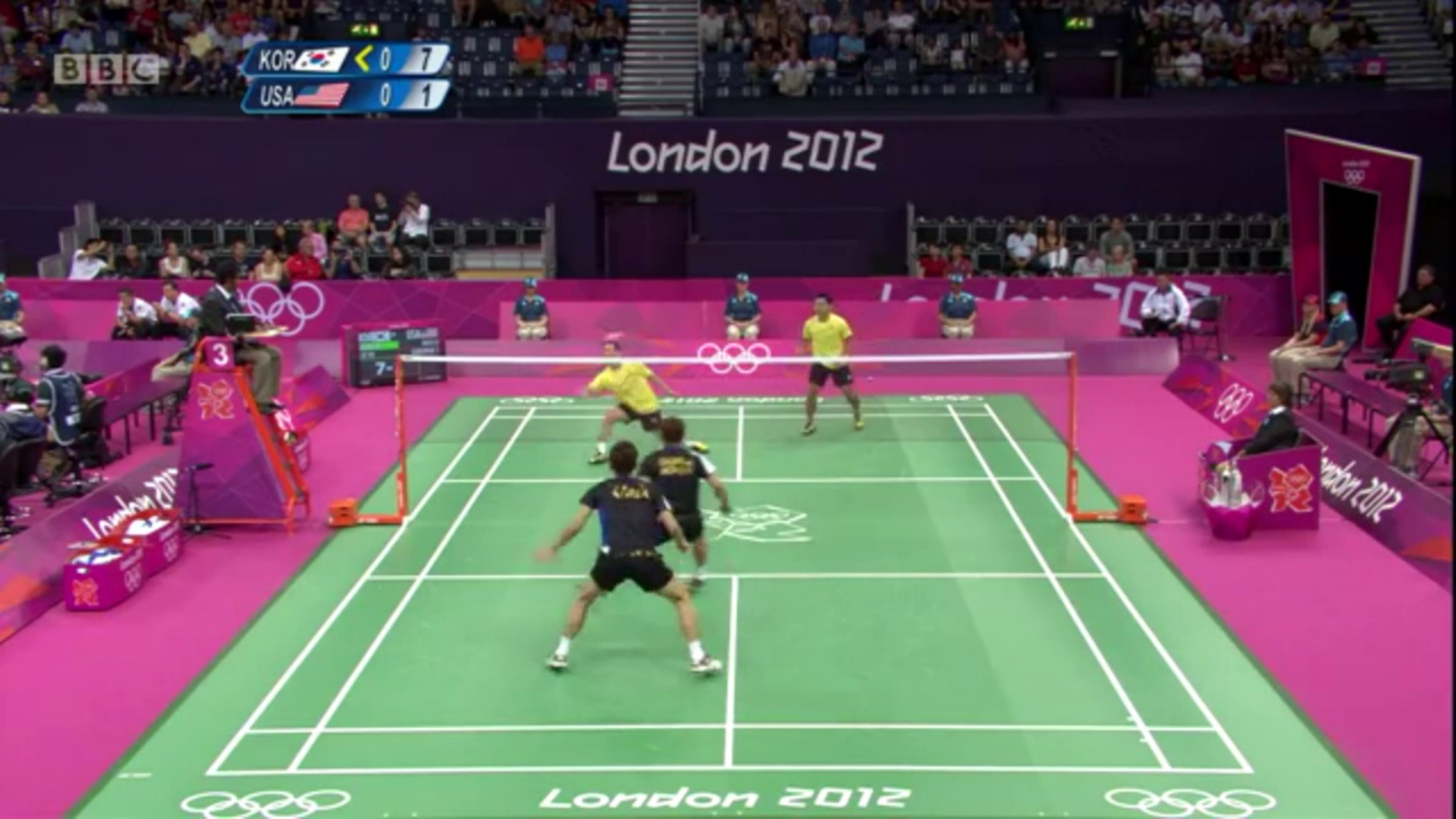}}
   \\
   \subfloat{\includegraphics[width=0.3\linewidth]{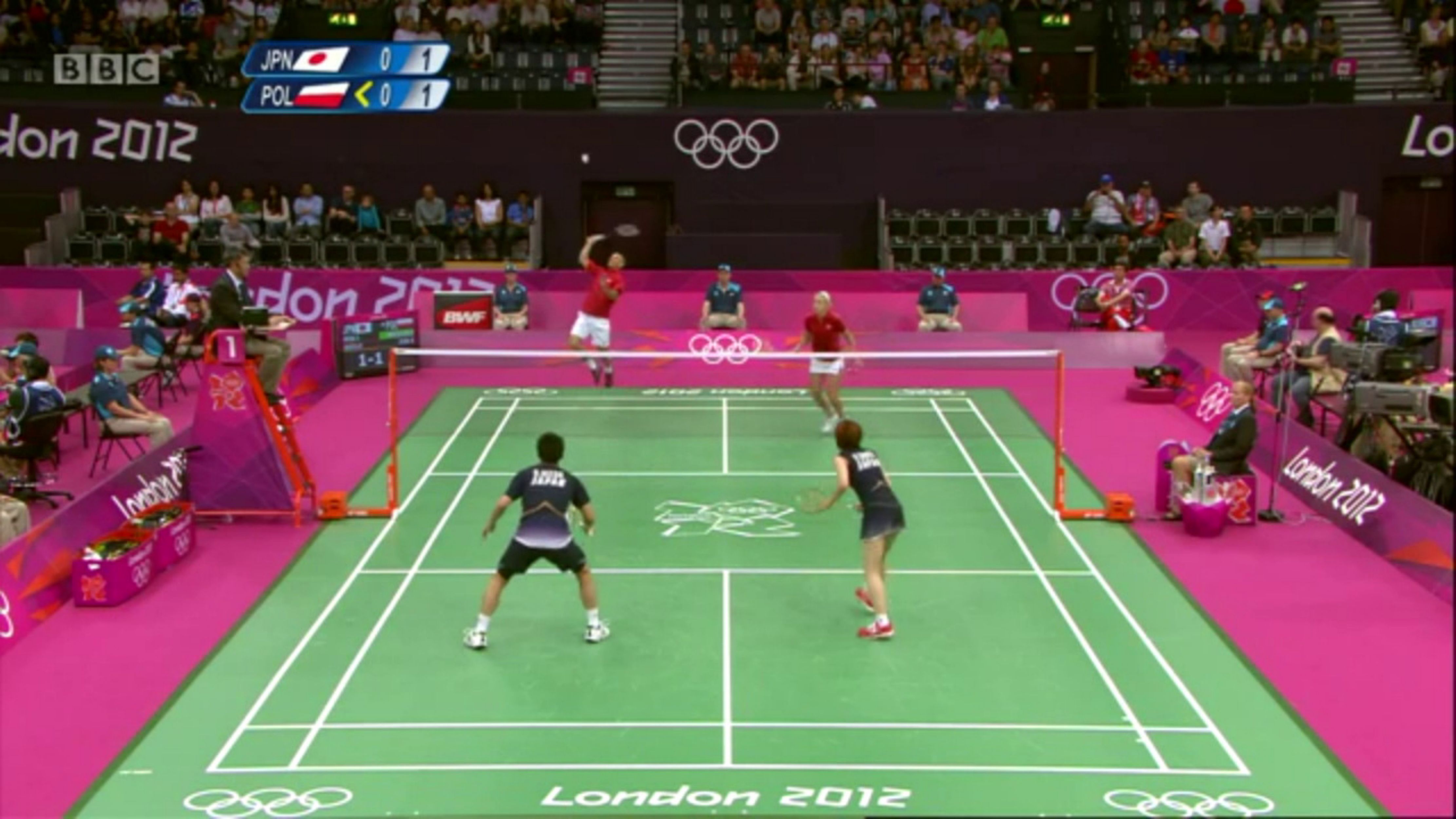}}
   \hfil
   \subfloat{\includegraphics[width=0.3\linewidth]{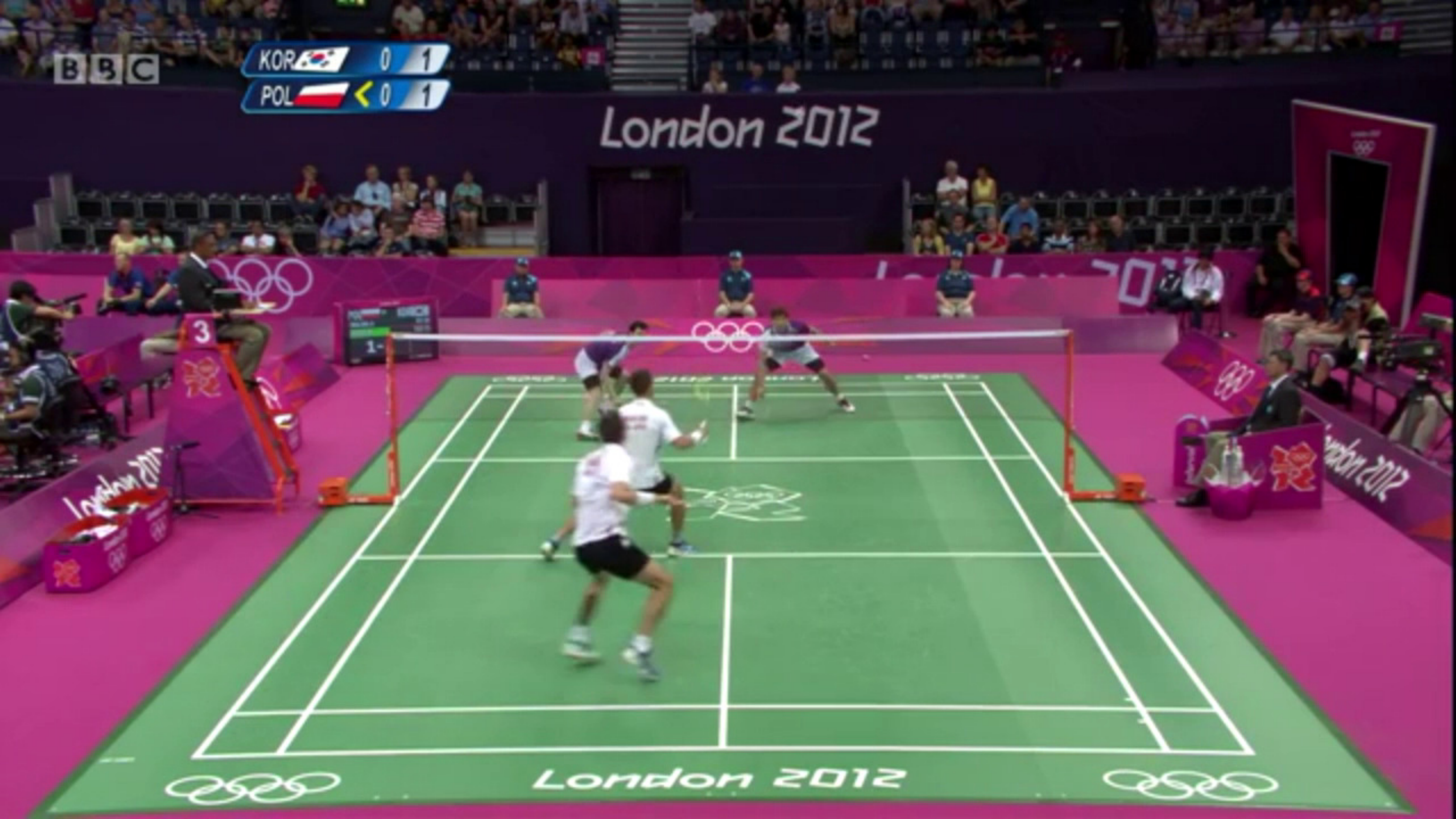}}
   \hfil
   \subfloat{\includegraphics[width=0.3\linewidth]{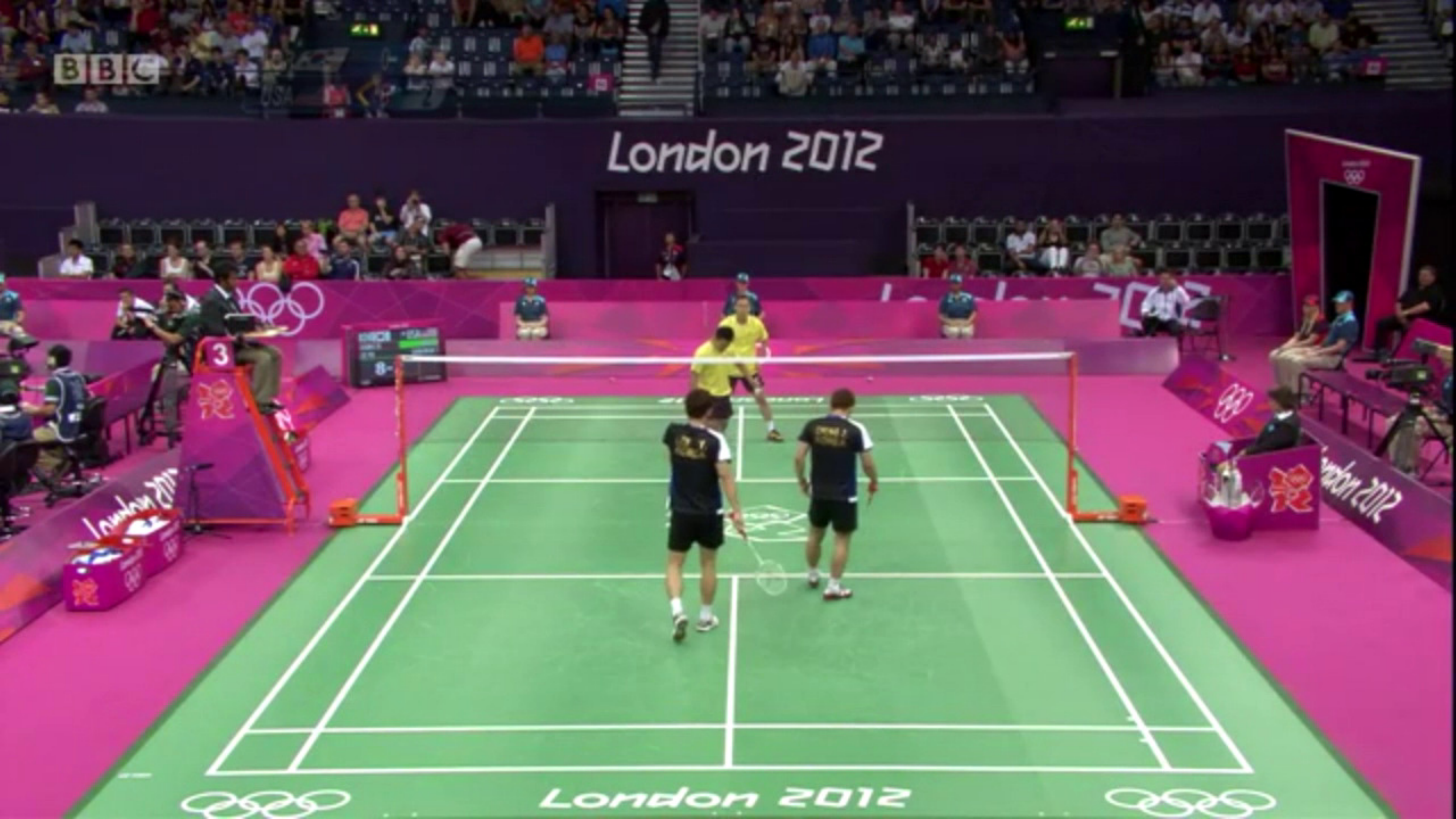}}
\end{center}
   \caption{Sample frames from the ACASVA badminton dataset.}
\label{fig:bm_dataset}
\end{figure}

We selected three videos from the ACASVA~\cite{decampos2011evaluation} dataset\footnote{Access to the dataset
was provided as a courtesy by the original authors~\cite{decampos2011evaluation}} of badminton doubles matches from
the Olympic games in London 2012. As in the table tennis dataset, the videos were edited to remove parts that do
not show the game itself and annotations were created manually to be used as ground truth. The resulting videos
were encoded at $1280 \times 720$ pixels per frame at 30FPS and they contained 5766 frames. Figure~\ref{fig:bm_dataset} displays some
sample frames from this dataset.

\subsubsection{Youtube volleyball}

\begin{figure}[!ht]
\begin{center}
   \subfloat{\includegraphics[width=0.3\linewidth]{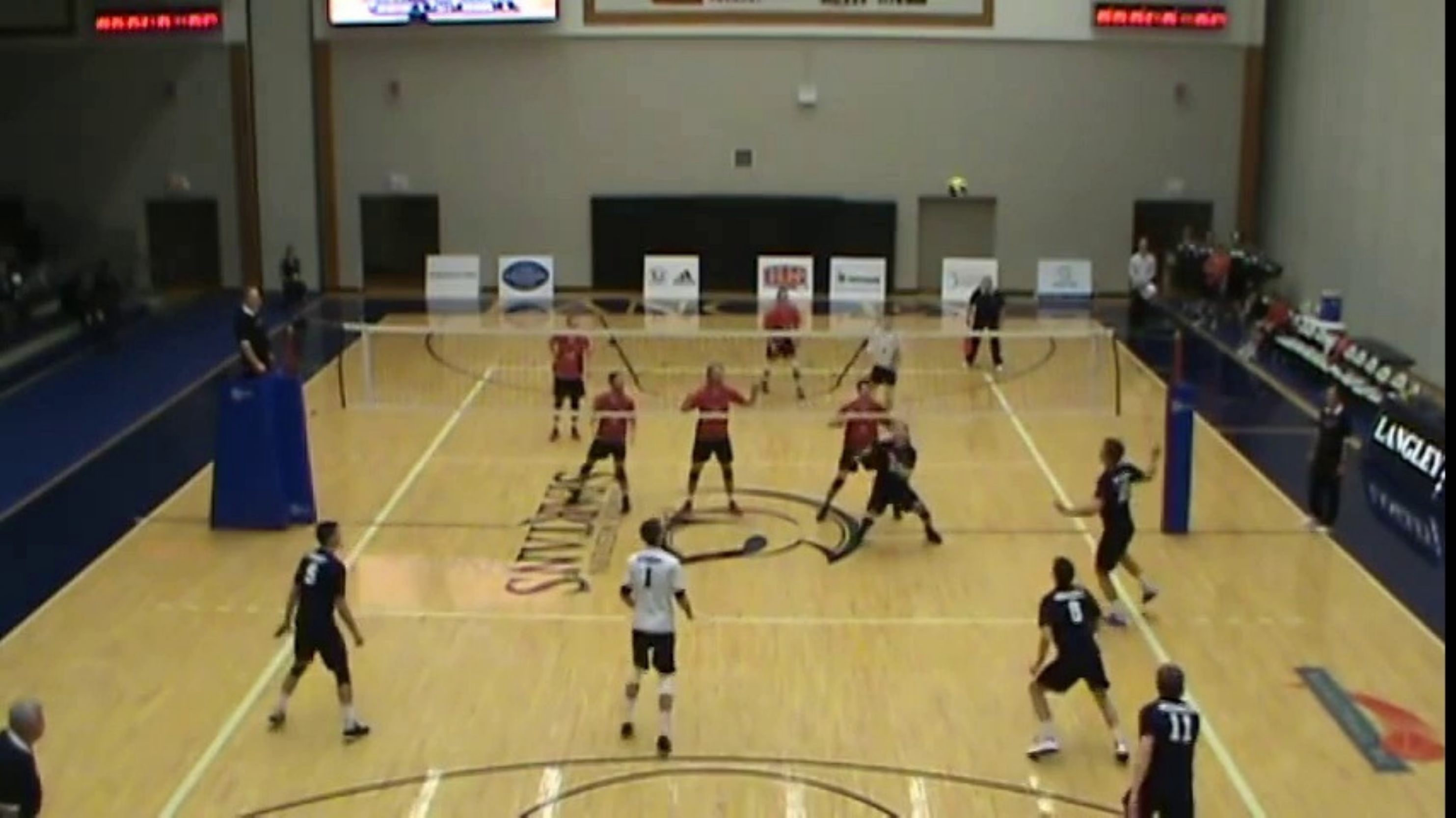}}
   \hfil
   \subfloat{\includegraphics[width=0.3\linewidth]{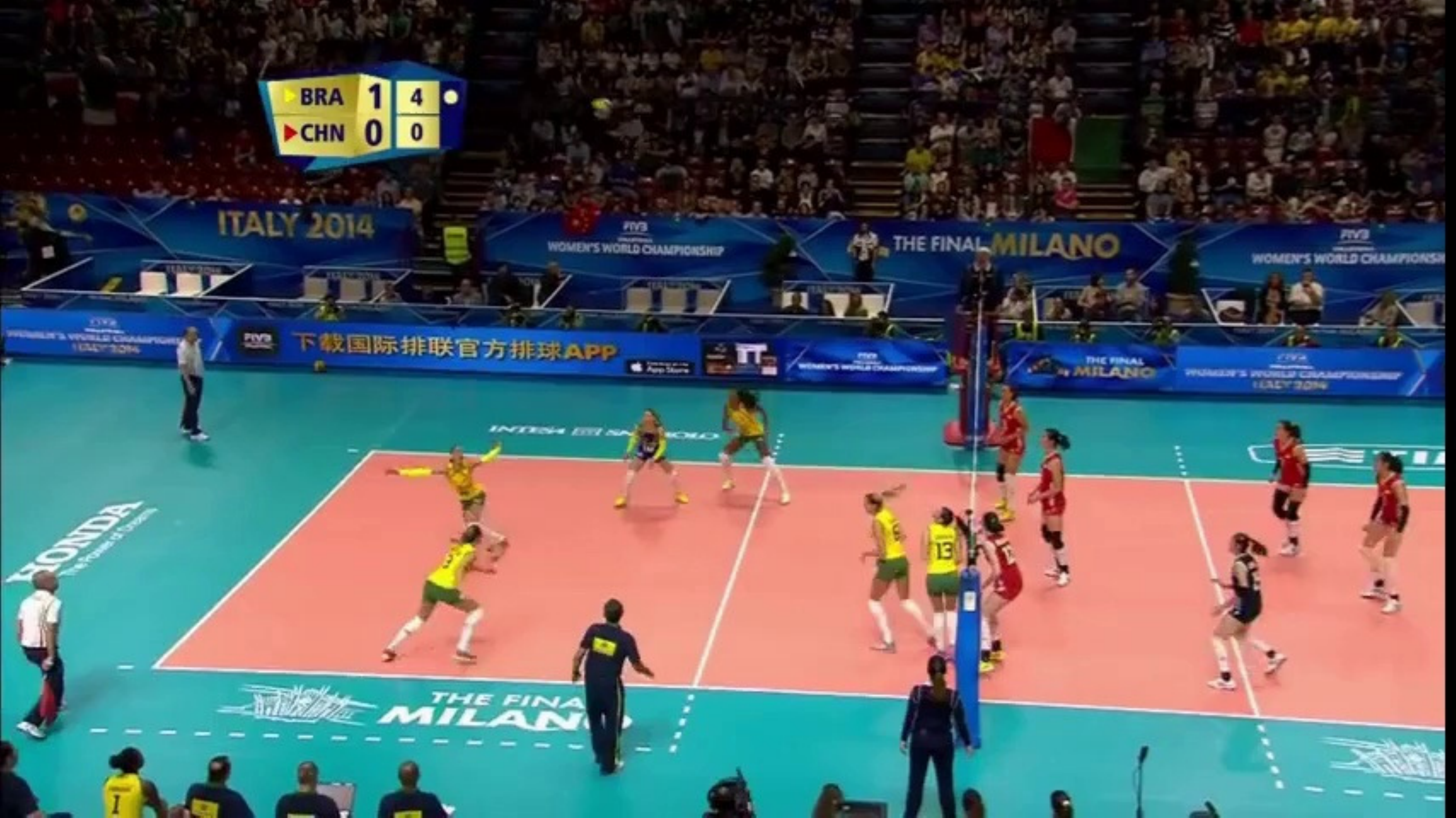}}
   \hfil
   \subfloat{\includegraphics[width=0.3\linewidth]{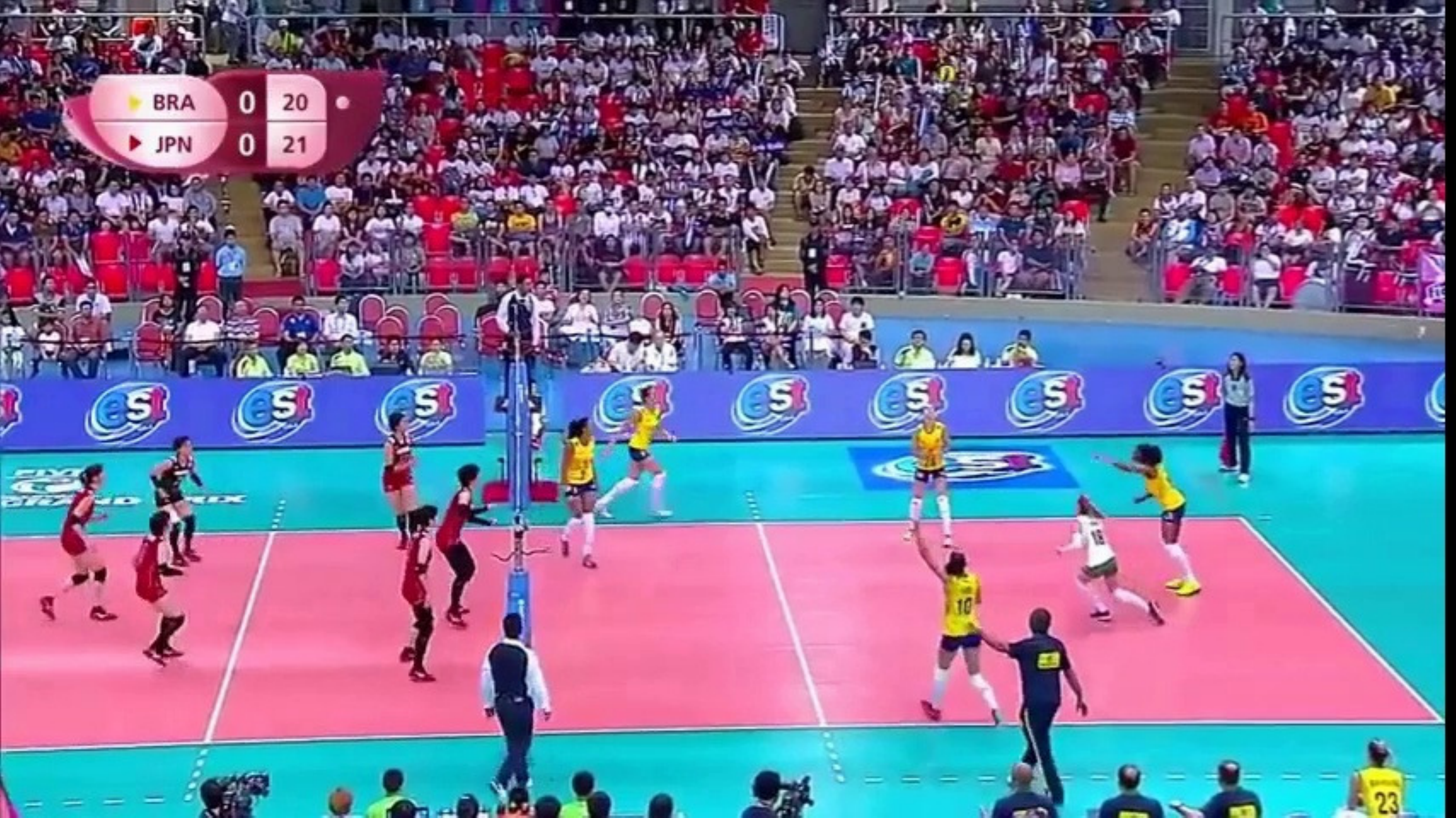}}
   \\
   \subfloat{\includegraphics[width=0.3\linewidth]{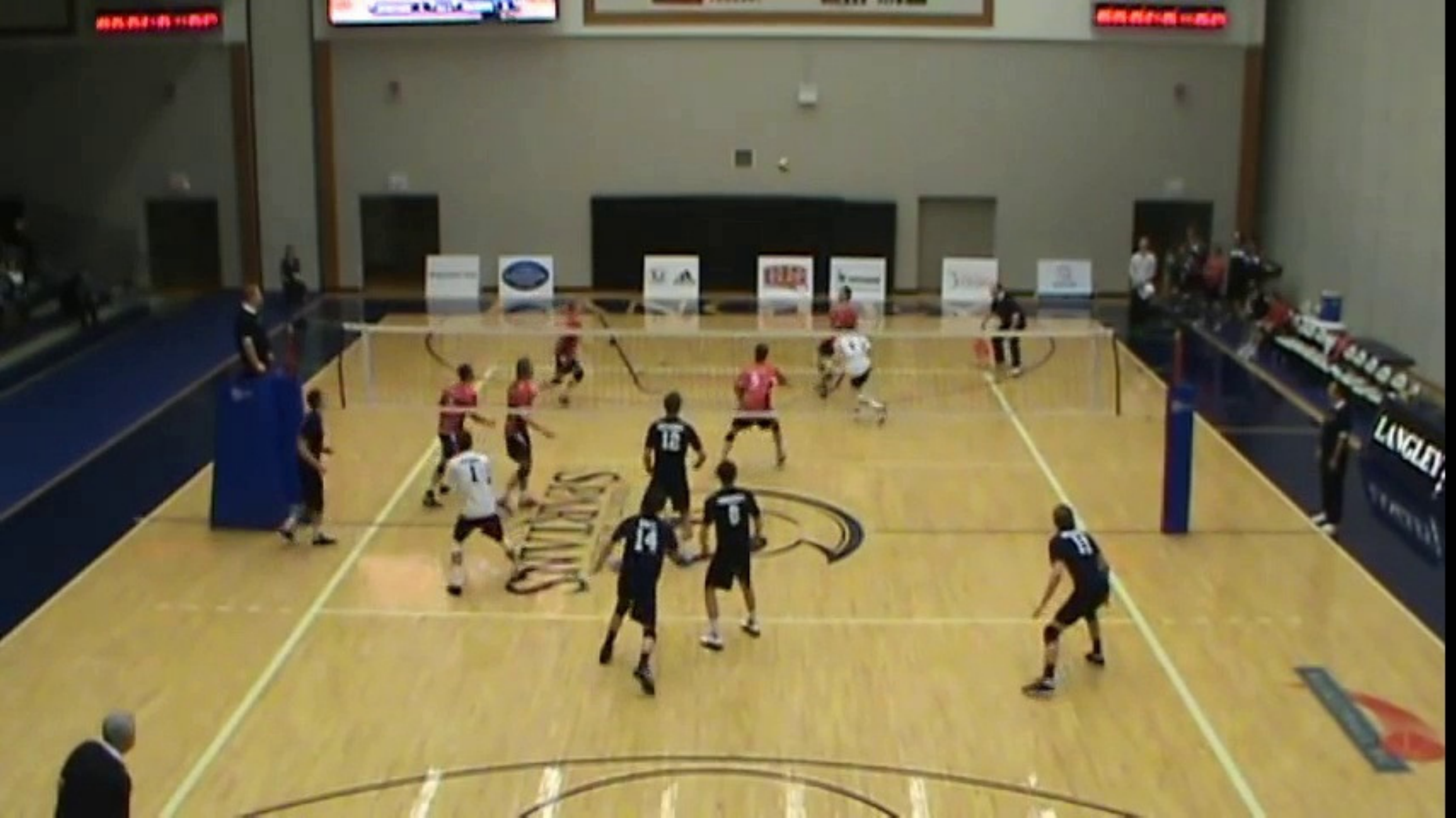}}
   \hfil
   \subfloat{\includegraphics[width=0.3\linewidth]{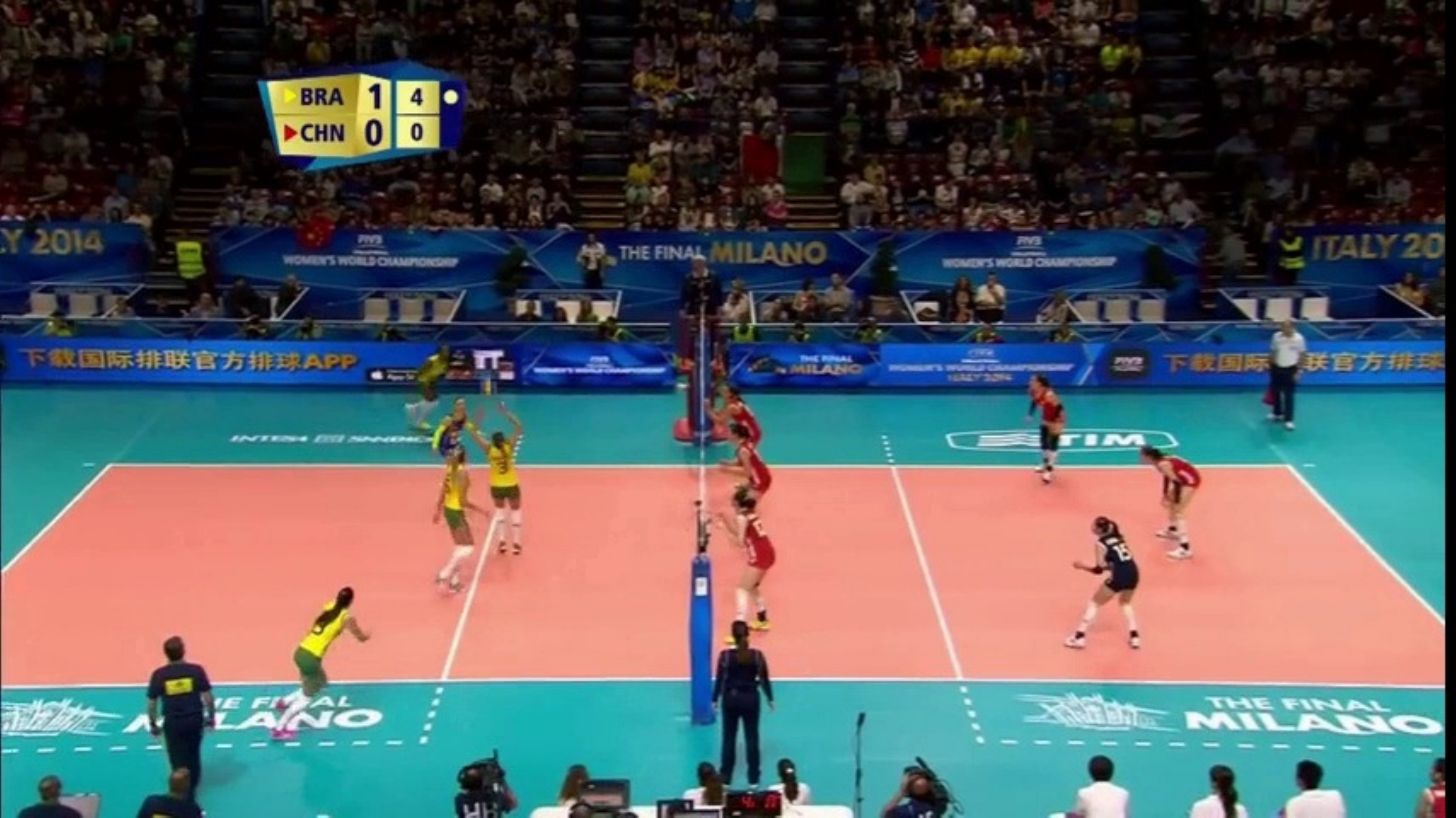}}
   \hfil
   \subfloat{\includegraphics[width=0.3\linewidth]{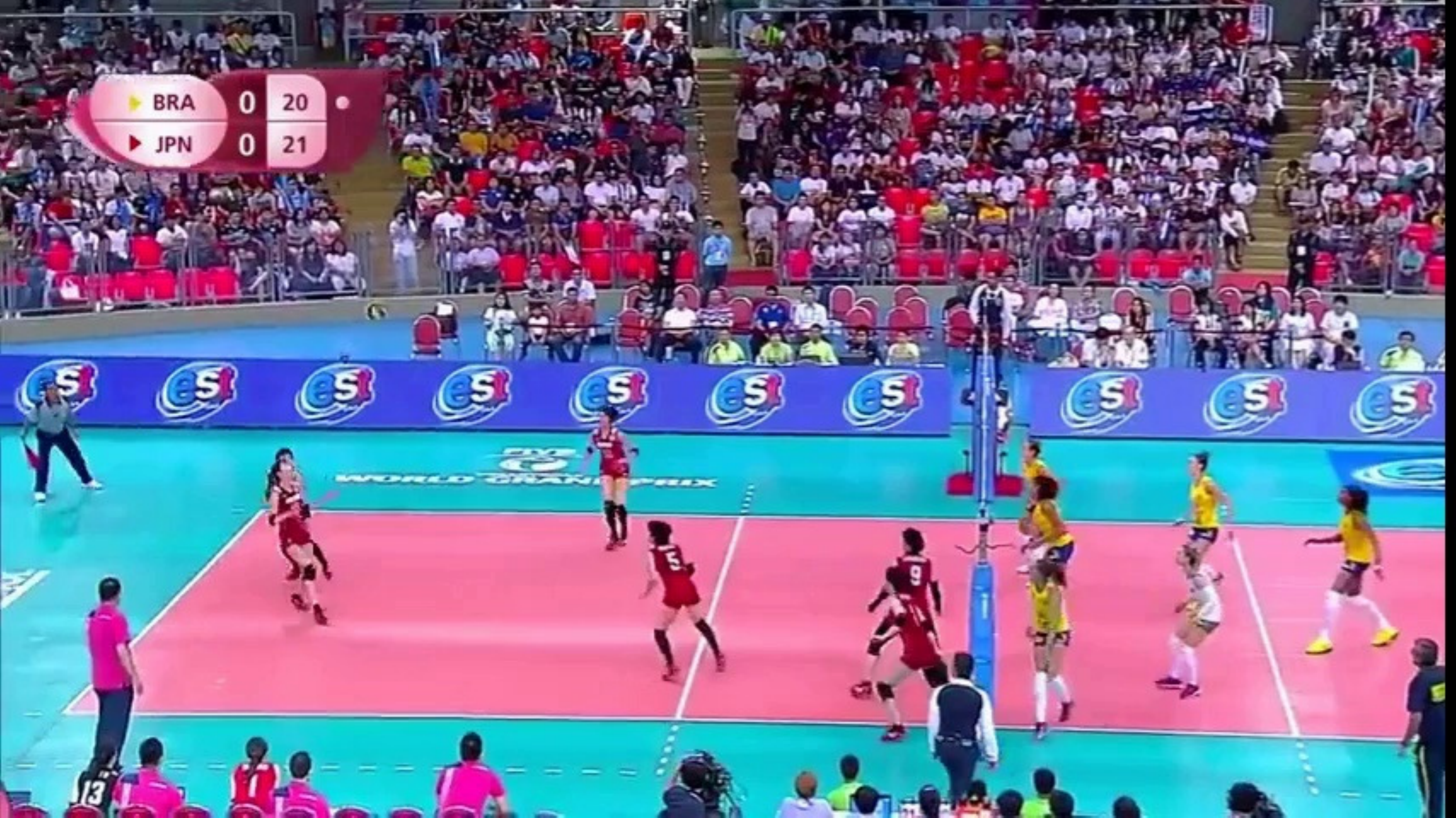}}
\end{center}
   \caption{Sample frames from the Youtube volleyball dataset.}
\label{fig:volley_dataset}
\end{figure}

Following the same guidelines as before, videos of volleyball matches
were chosen and edited. This dataset is composed of 3 videos recorded at 30 FPS at a resolution of
$854 \times 480$, containing 5080 frames. The videos in this dataset do not contain cuts, which
allow us to analyze the behavior of the tested trackers in this configuration. Videos using two
different camera angles were collected, and those captured from a side view present some fast
motion when the camera follows the ball from one side of the court to the other.
This dataset also contains a significantly greater number of players to be tracked (6 on each team).
Figure~\ref{fig:volley_dataset} displays some
sample frames from this dataset.

\subsection{Choosing the parameters}
\label{subsec:tracking_parameters}

We separated the Youtube table tennis videos into two sets, one for parameter estimation and another for evaluation. The set for parameter estimation was composed of
one video containing 2461 frames. This video contained a longer table tennis match with all the expected challenging situation including overlapping between players
of the same team and camera cuts.

We performed a test to find the best weights $\rho_A$, $\rho_S$ and $\rho_O$ for the
scoring function (Equation~\ref{eq:score_function}). Each configuration was evaluated five
times and the presented results correspond to the average between all the observations.

\begin{figure}[!ht]
\begin{center}
   \includegraphics[width=0.9\linewidth]{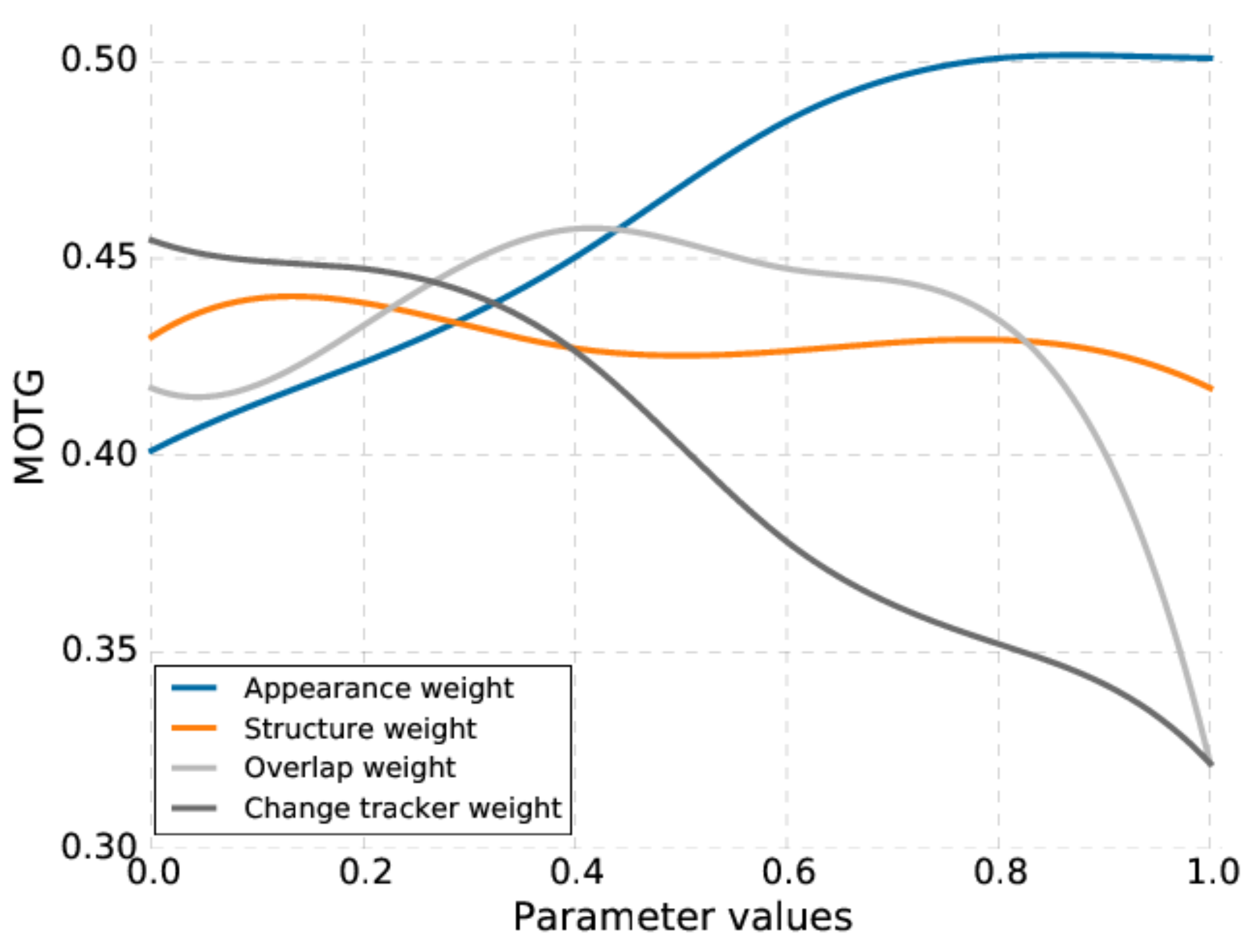}
\end{center}
   \caption{$MOTG$ according to graph parameter selection. Change tracker weight represents $(1 - \rho_A - \rho_S - \rho_O)$.}
\label{fig:graph_params}
\end{figure}

For the test, we tried all combinations of values in the interval $[0, 1]$ with an increment of $0.2$, \ie
each parameter could be 1 out of 6 values. Ignoring the invalid configurations (where $\rho_A + \rho_S + \rho_O > 1$),
this produces $56$ possibilities. We evaluated all of them with the $MOTG$ metric and found out that the best configuration
consisted of $\rho_A = 0.4$, $\rho_S = 0$ and $\rho_O = 0.6$ with $MOTG = 0.61$. Besides finding the best parameters, we were
also interested in verifying how each parameter affected the results. In order to do so, for each parameter value, we computed
the average $MOTG$ of all configurations that used that parameter and evaluated how the scored changed according to each
one of them. Note that lower values have more samples, since more valid combinations of the remaining parameters exist.
Figure~\ref{fig:graph_params} shows the MOTG curves for each parameter value. It is worth noting that the method performance
varies smoothly with the parameters, thus showing robustness to small parametric changes.

It is clear that the appearance is an important feature, showing a direct relation to the score. The change tracker,
on the other hand, proved to hinder the performance and thus, should not be used. The overlap term presents the best
performance at near the middle of the interval, while the structural part does not seem to affect much the results.
Although the structural score is not used in the best configuration found for this test, some other configurations
which did employ structure also appeared near the top results. Besides, the chart in Figure~\ref{fig:graph_params}
shows that the use of structure does not affect negatively the overall scores, thus indicating that the structure could be
useful in some other applications. Although not used for evaluation in this test, the inclusion of structure
to generate candidates for tracking significantly improves the results, as it will be shown in the
next section.

We also experimented with varying the old temporal weight factor $\rho_T$, the score threshold for removing candidates
$\tau_S$ and the score threshold for removing old trackers $\tau_R$. However, the results did not show any clear
behavior, as demonstrated by the previous parameters. Therefore, we just chose one of the top performing configurations.

\begin{figure}[!ht]
\begin{center}
   \includegraphics[width=0.9\linewidth]{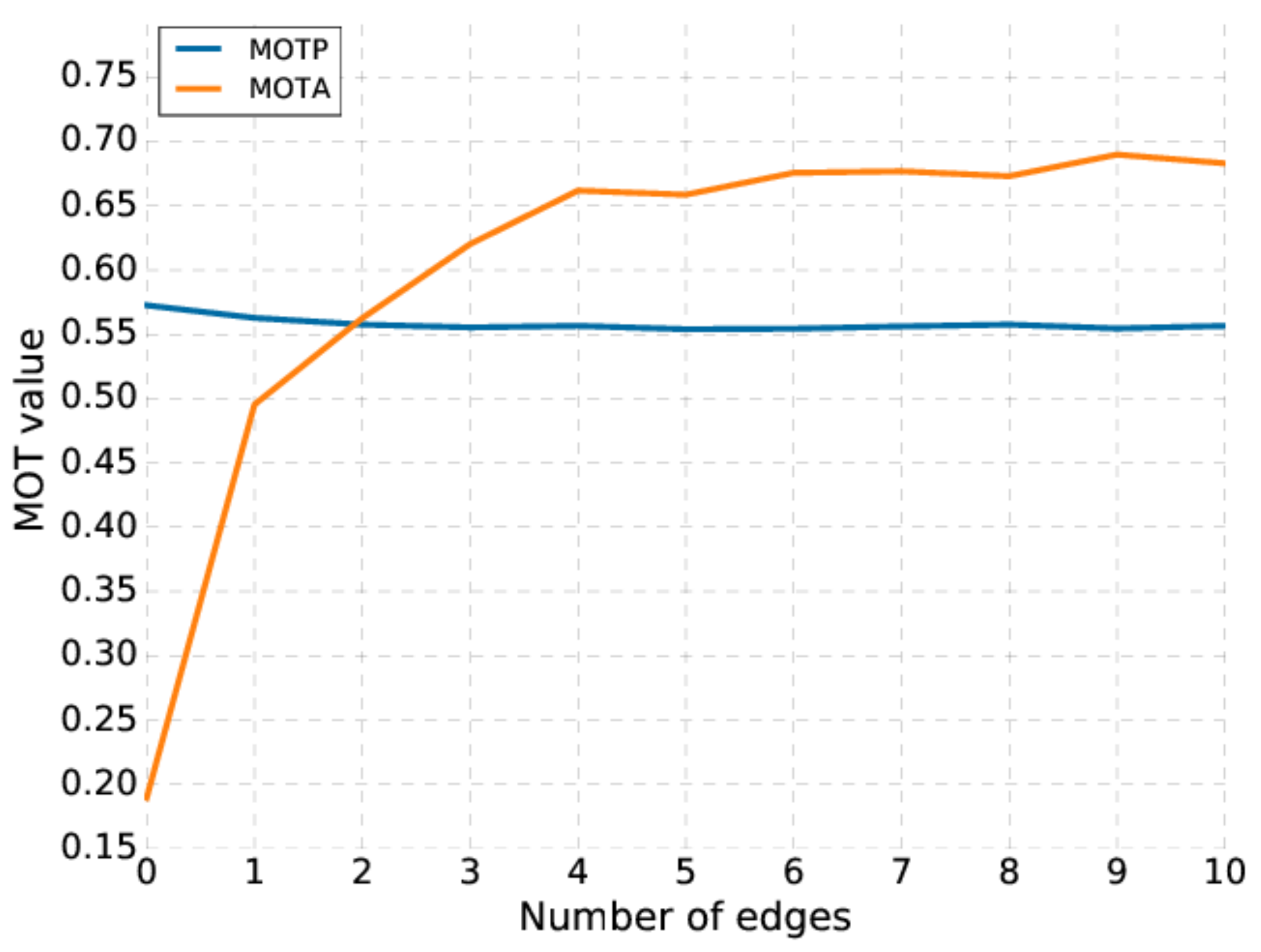}
\end{center}
   \caption{$MOTA$ and $MOTP$ according to the number of edges in $M_A$.}
\label{fig:chart_edges_mot}
\end{figure}

Another parameter that is worth investigating is the graph topology represented by the adjacency matrix $M_A$. We conducted some
tests by varying the number of edges of the graph to see how it affected the performance. As the graph in the table tennis video is composed
of 5 vertices, we tested all configurations until reaching a complete graph (10 edges). For each configuration
with $k$ edges, we randomly generated 5 adjacency matrices to be tested. Notice that the suppression
of edges in $M_A$ affects the information available for generating the candidates using the matrix $M_C$.
In order to decrease the impact of the candidates generation on the evaluation of the topology, we defined
$M_C = (m_{ij} = 5, \text{if } i \neq j, \text{otherwise } 0)$, \ie each vertex generates 5 candidates for all the others.
All the other parameters were fixed according to the tests reported above.
The result is presented in Figure~\ref{fig:chart_edges_mot}

The results evidence that using more edges has a clear impact on the accuracy,
which significantly increases until around 4 edges.
One point worth of note is that the use of additional edges does not negatively impact the results.
Based on these results, and on knowledge about the configuration of the game, we chose to use the same adjacency matrix
defined in Equation~\ref{eq:adjacency_matrix}, where the first four lines and columns represent the players, 
while the last one represents the net or the table (from now on referred to as middle line),
on the badminton and table tennis videos. The adjacency matrix for the
volleyball videos followed a similar design, where the net was connected to every player and the players
of the same team were fully connected.
This matrix considers the relations between players of the same team and all the players
and the middle line. The relationship with the middle line is important because the players should be
close and on opposite sides of it during the game. On the other hand, exploiting the relationships
between players of the same team helps us to handle temporary occlusions. For the remaining
experiments, the candidates matrix $M_C$
was chosen to use the middle line as a reference to generate $10$ candidates for each player.

Table~\ref{tab:parameters} summarizes all the parameters chosen for the evaluation.
The same values were used for all the experiments, independently of the dataset.
\begin{table}[!ht]
\caption{Parameters for the tracking framework.}
\label{tab:parameters}
\centering
\begin{tabular}{l l}
\hline
distance histogram range                        & $H_d[0, 1]$\\
angle histogram range                           & $H_{\theta}[0, 2 \pi]$\\
number of distance histogram bins               & $bins(H_d) = 25$\\
number of angle histogram bins                  & $bins(H_{\theta}) = 18$\\
number of particles per object                  & $N_p = 50$\\
number of random optimization runs              & $N_{RI} = 10$\\
initial particle spread deviation               & $\sigma_c = 10$\\
particle spreading factor                       & $\alpha = 5$\\
maximum particle sum of weights                 & $\beta = 25$\\
appearance weight                               & $\rho_A = 0.4$\\
structure weight                                & $\rho_S = 0$\\
overlapping weight                              & $\rho_O = 0.6$\\
old temporal weight factor                      & $\rho_T = 0.8$\\
score threshold for removing candidates         & $\tau_S = 0.4$\\
score threshold for removing old trackers       & $\tau_R = 0.2$\\
overlap threshold for removing candidates       & $\tau_O = 0.25$\\
graph optimization iteration threshold          & $\tau_I = 10$\\
conv. kernel for confidence in $(0.7, 1.0]$     & $k_C = [0.3, 0.4, 0.3]$\\
conv. kernel for confidence in $(0.3, 0.7]$     & $k_C = [0.15, 0.2, 0.3, 0.2, 0.15]$\\
conv. kernel for confidence in $[0.0, 0.3]$     & $k_C = [0.1, 0.13, 0.17, 0.2, 0.17, 0.13, 0.1]$\\
\hline
\end{tabular}
\end{table}

\subsection{Experimental setup}
\label{subsec:experimental_setup}

We tested our approach on all the datasets previously presented. For the table tennis dataset, the video used
for estimating the parameters was not included in this evaluation step.
The task in these videos was to track all the players and the middle line.
We purposely track using only the torso of the players in order to create more appearance ambiguity
and check whether the graph model can deal with this situation. All the videos are captured from a
single camera, which moves in 2 of the volleyball videos, but it is fixed in all the others. As before, all the tests were performed five times and
the average of all of the results was taken.

We compare our approach (Online Graph) with other methods from the literature. The first one was the same
particle filter color tracker we used (PF), but without the graph information. In this way, we could verify
whether the addition of graphs brings any significant improvement to the classical approach.
The second is a previous version of this approach (Offline Graph) as proposed in~\cite{morimitsu2015attributed}.
This method requires an annotated dataset for training the model offline, as well as uses
a somewhat simpler score function.
This test allows us to demonstrate the contributions of this method over the previous one.
The third one was SPOT~\cite{zhang2014preserving}. This tracker also uses a structural graph as ours,
but only uses distance information for structure. The tracking procedure
consists in classifying the multiple graphs generated from HOG detectors using a structured SVM.
The last tracker is STRUCK~\cite{hare2011struck}, a single object tracker
that, according to a recent benchmark~\cite{wu2015object}, was the best performing method in several datasets.
This method employs kernelized structured SVM to handle tracking by classifying the input
directly into the spatial position domain. In this way, the intermediate step of dividing
the input templates into positive and negative training samples is skipped. Both PF and STRUCK
are single object trackers and, thus, they track each object independently. STRUCK was included in the comparison
to verify whether the use of a set of highly discriminative trackers alone would be able to solve
the proposed problem.

\subsection{Evaluation on the datasets}
\label{subsec:evaluation_dataset}

The results are presented in Table~\ref{tab:results_mot}. The values correspond to the
average of the results obtained from all videos. Although the precision of the proposed method
is a bit lower than the ones obtained by other approaches, there is a significant increase in accuracy. This result
is further evidenced by the best true and false positive rates. Even on the volleyball dataset, which
does not contain camera cuts, our graph approach is better than STRUCK, showing that the structural constraints
are a valuable aid in improving tracking in more cluttered scenes. Also, the online method
showed a significant improvement over the offline one in the number of ID switches.
This shows that the proposed approach is much more stable and do not cause many tracking failures.
One point to note is that STRUCK performed similarly or worse than
the particle filter approach in the badminton and table tennis sequences. This is explained because the videos in these datasets often
contain many situations of camera cut. When this happens, both PF and STRUCK can only recover tracking
when the target gets close to the point where it was lost. In that sense, the particle filters usually
are able to recover the target more often because the particles are spread in a broader area
than the STRUCK search radius. Since STRUCK conducts a dense neighbor search, as opposed to the sampled
spread of PF, its search area must be kept smaller, and thus it is unable to detect the target in many situations.
Another reason is that STRUCK updates the model along the video. In this case, if the target is tracked
incorrectly, the model tends to deteriorate if the target is lost. It can also be observed that SPOT did not show good results on these datasets. According to the observed
results, the main reason seems to be that the structural model used by SPOT is sometimes too rigid
and not well suited for a situation where the structural properties between the objects are subject
to large changes in a short amount of time.
\begin{table}[!ht]
\caption{Observed results on the datasets. The arrows indicate whether lower or higher values
are better.}
\label{tab:results_mot}
\centering
\begin{footnotesize}
\begin{tabular}{l | l | c | c | c | c | c}
\hline
Dataset      & Method        & $MOTP$ $\uparrow$ & $MOTA$ $\uparrow$ & $IDSW$ $\downarrow$ & $TPrate$ $\uparrow$ & $FPrate \downarrow$\\
\hline
Youtube      & Online graph  & 0.589             & \textbf{0.796}    & 85                  & \textbf{0.893}      & \textbf{0.096}     \\
table tennis & Offline graph & \textbf{0.622}    & 0.515             & 207                 & 0.743               & 0.229              \\
\&           & PF            & 0.608             & 0.46              & 89                  & 0.724               & 0.264              \\
ACASVA       & SPOT          & 0.539             & -0.008            & \textbf{51}         & 0.492               & 0.5                \\
badminton    & STRUCK        & 0.619             & 0.486             & 126                 & 0.734               & 0.229              \\
\hline
Youtube      & Online graph  & 0.495             & \textbf{0.367}    & \textbf{624}        & \textbf{0.667}      & \textbf{0.302}     \\
volleyball   & PF            & 0.503             & 0.189             & 850                 & 0.575               & 0.386              \\
             & SPOT          & 0.447             & -0.608            & 836                 & 0.179               & 0.786              \\
             & STRUCK        & \textbf{0.552}    & 0.294             & 675                 & 0.631               & 0.338              \\
\hline
\end{tabular}
\end{footnotesize}
\end{table}

\begin{figure}[!ht]
\begin{center}
   \subfloat{\includegraphics[width=0.45\linewidth]{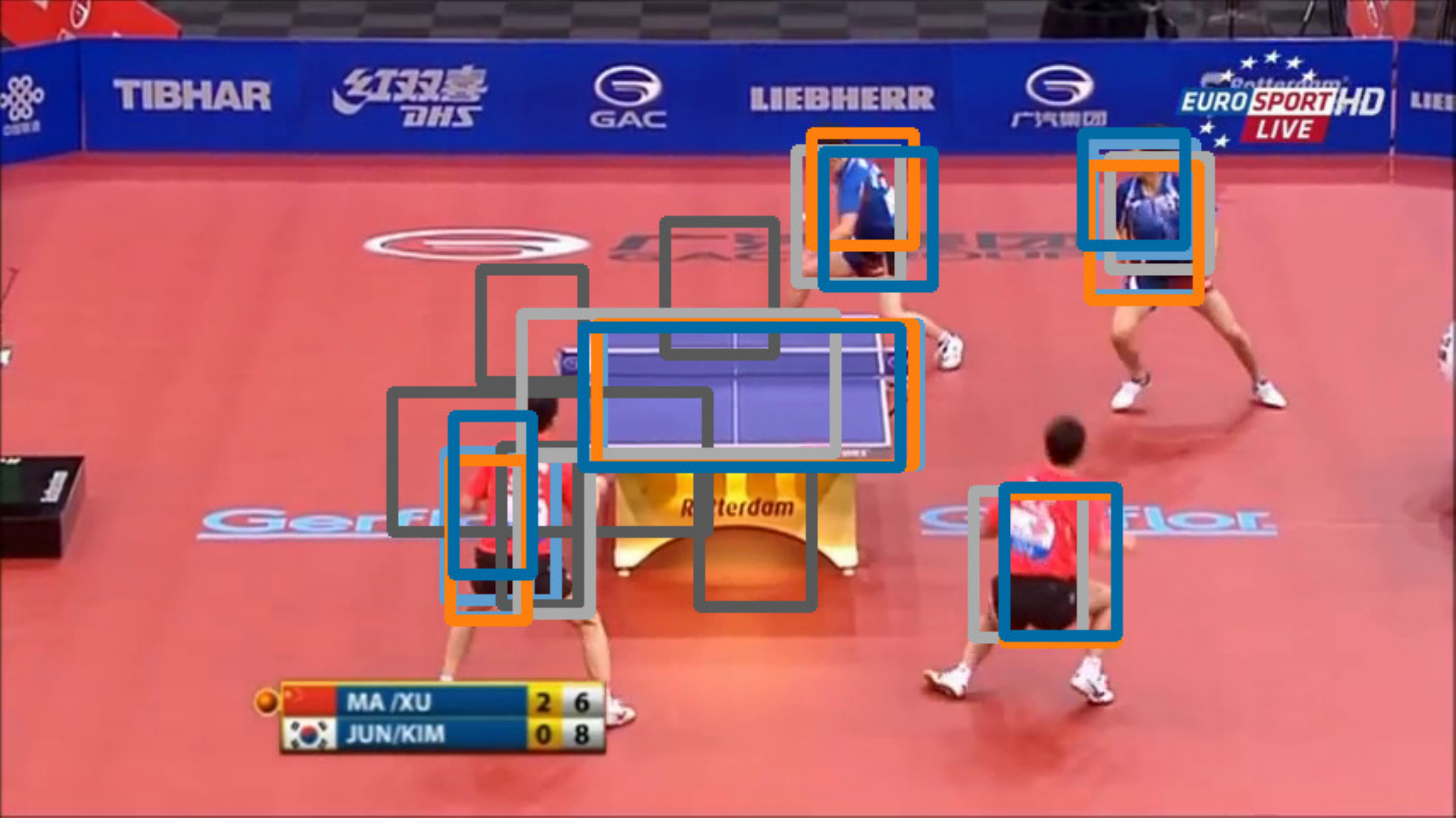}}
   \hfil
   \subfloat{\includegraphics[width=0.45\linewidth]{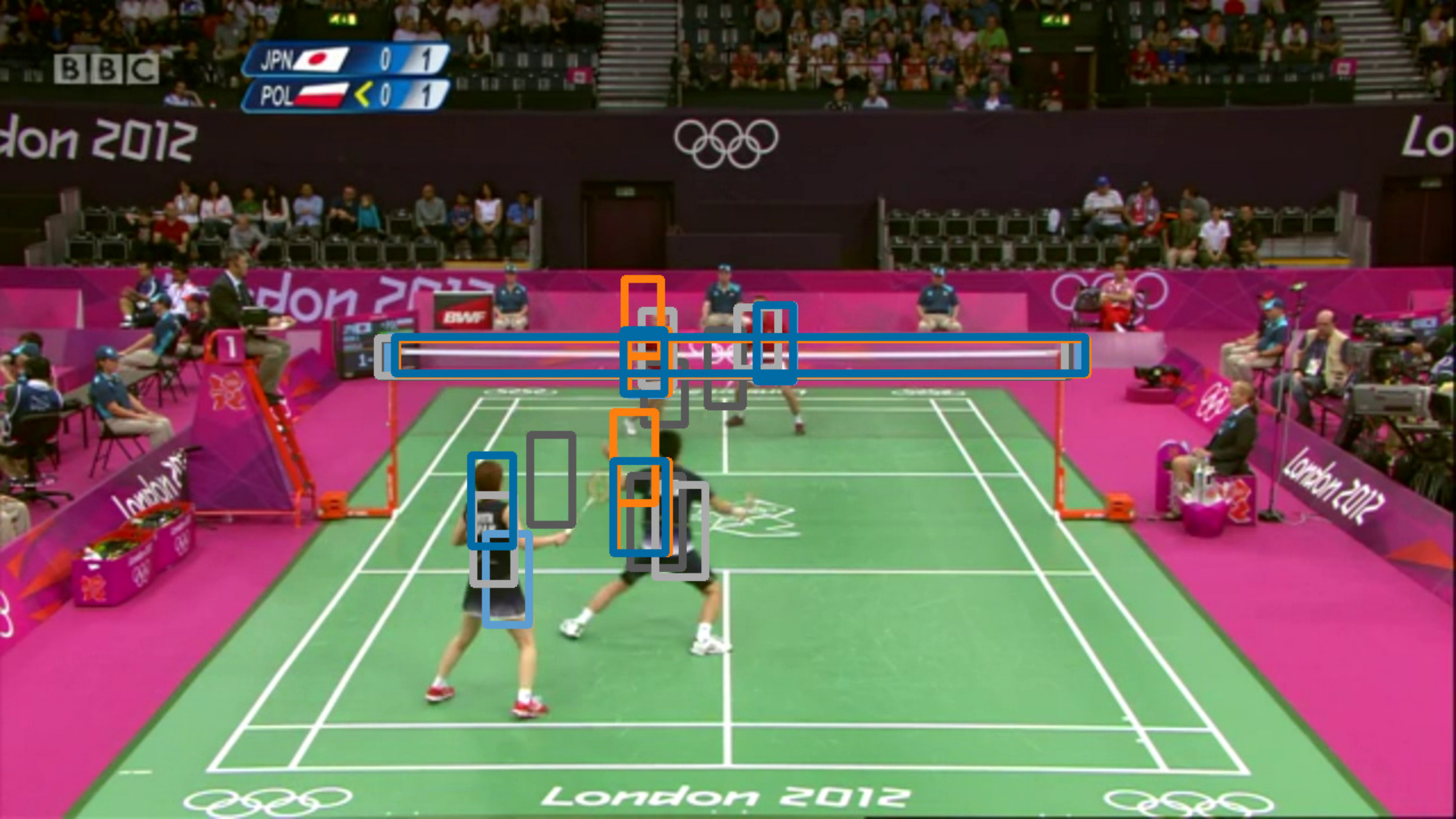}}
   \\
   \subfloat{\includegraphics[width=0.45\linewidth]{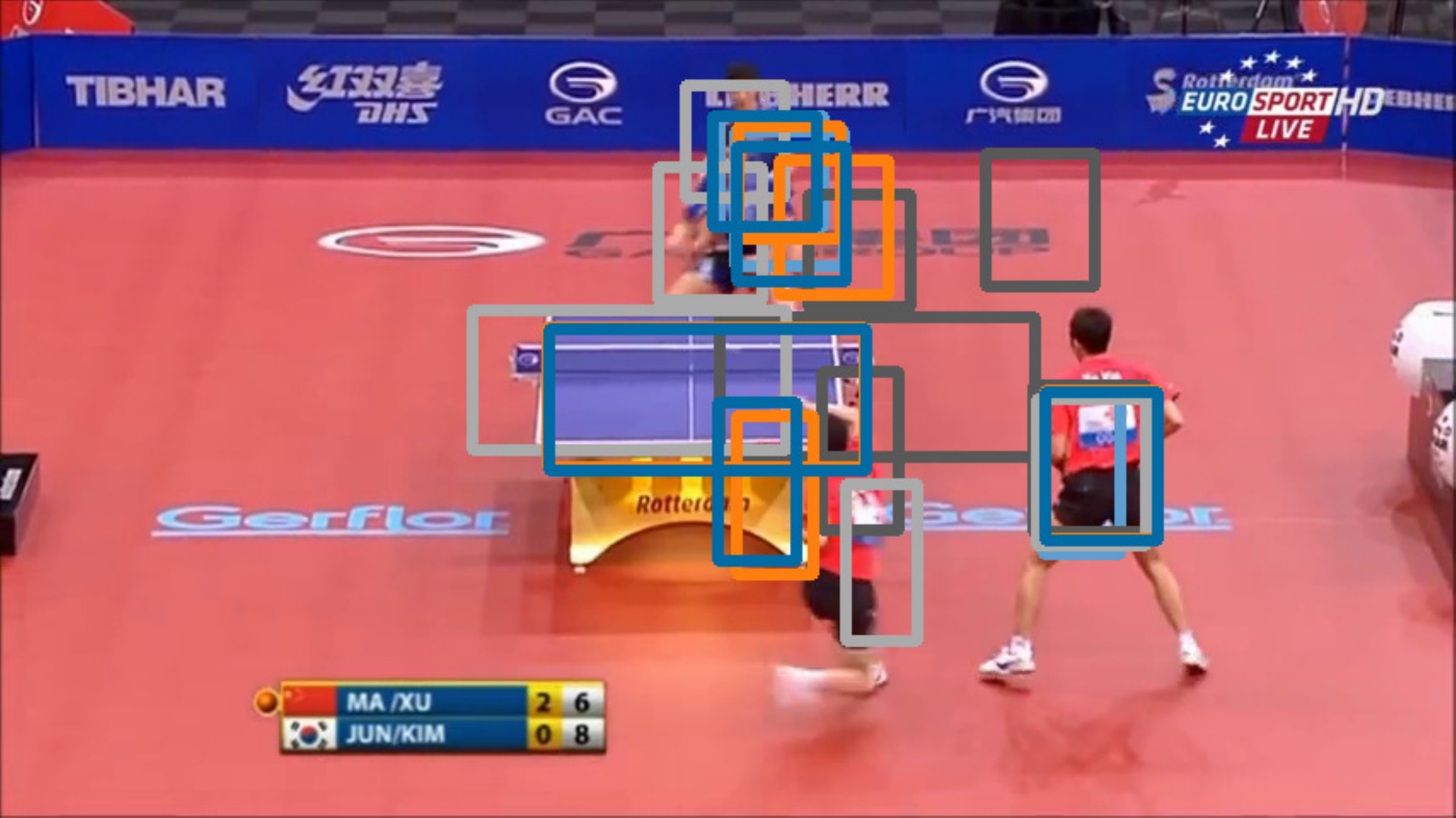}}
   \hfil
   \subfloat{\includegraphics[width=0.45\linewidth]{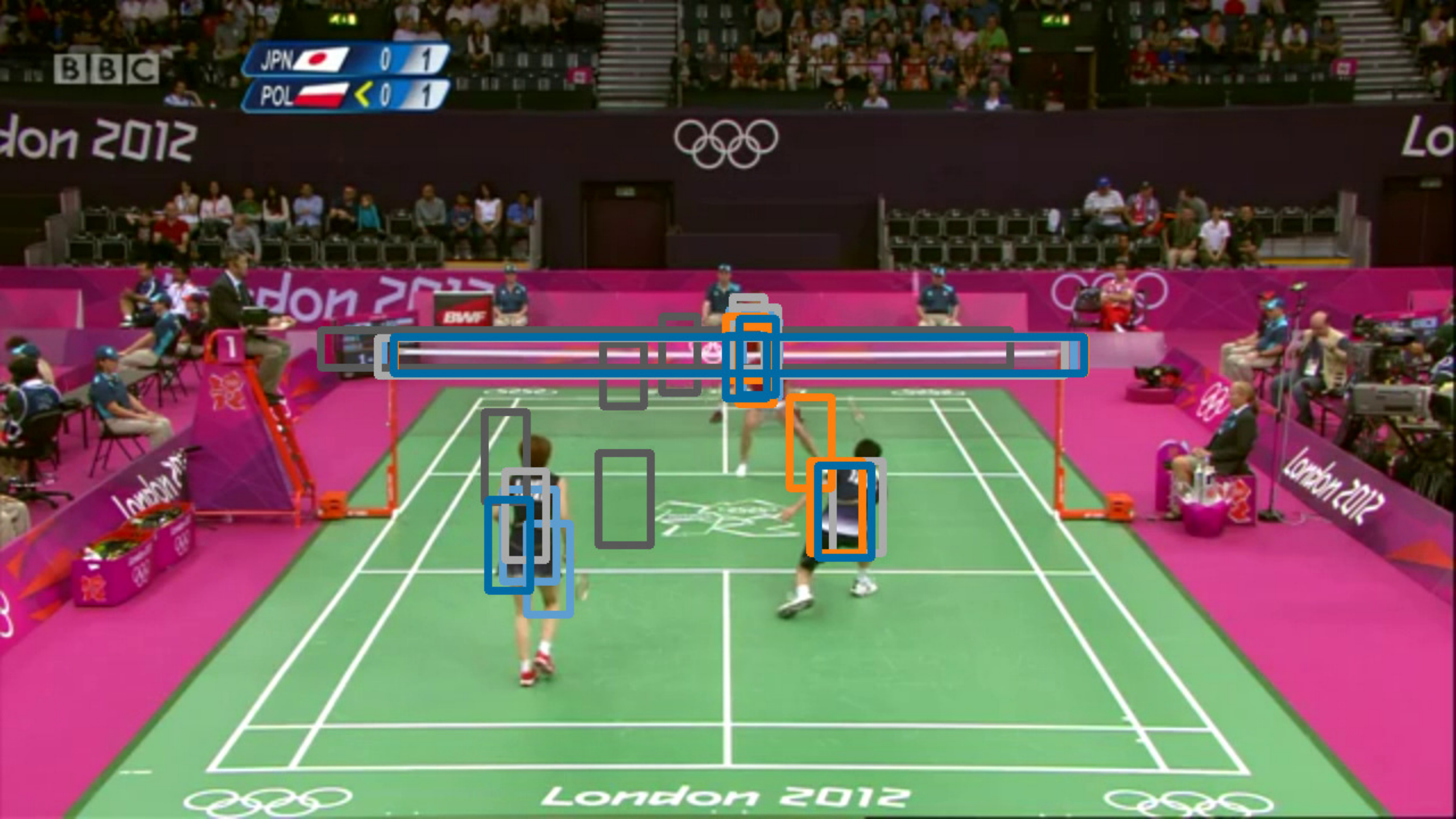}}
   \\
   \subfloat{\includegraphics[width=0.45\linewidth]{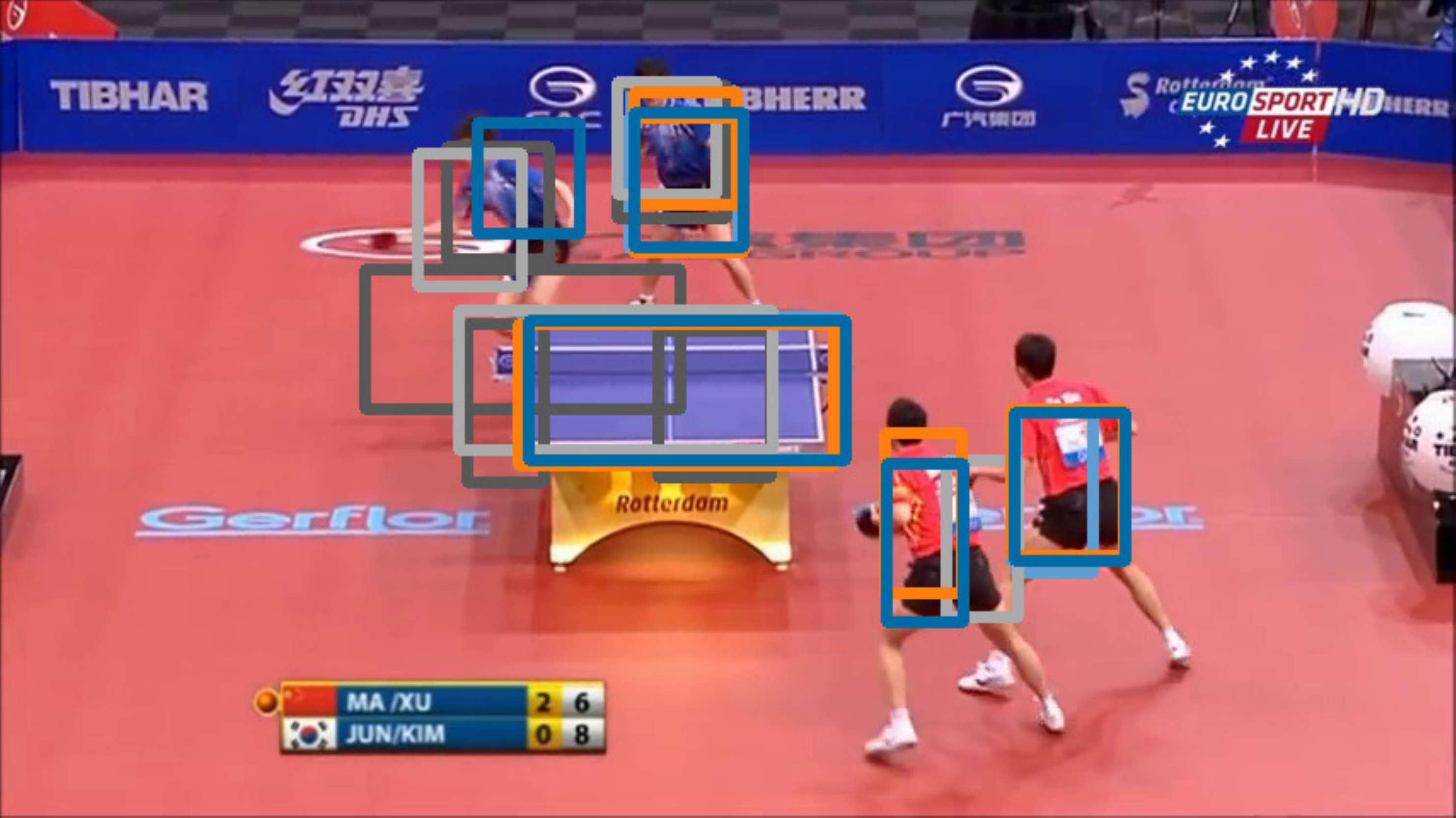}}
   \hfil
   \subfloat{\includegraphics[width=0.45\linewidth]{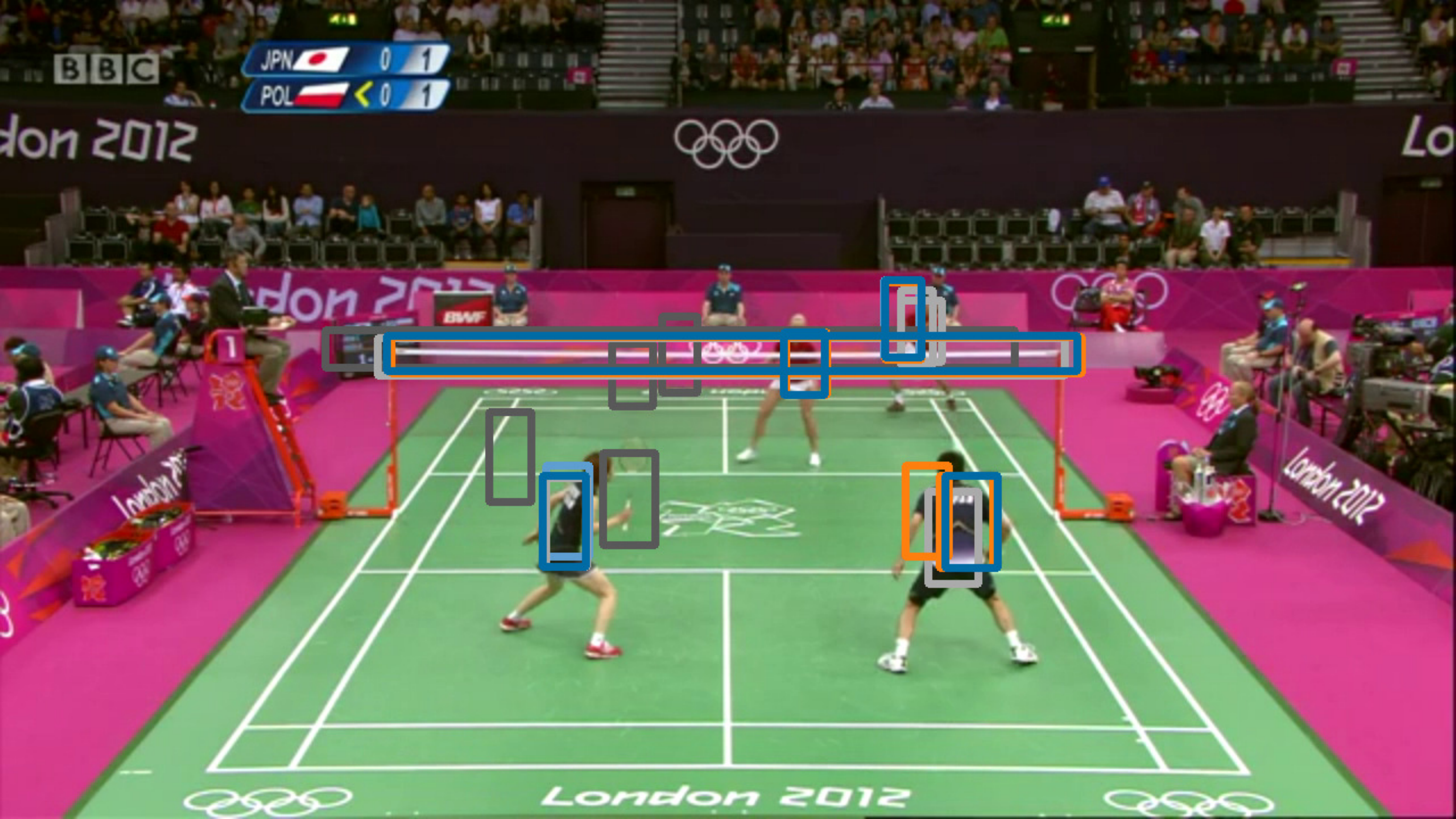}}
\end{center}
   \caption{Tracking results after occlusion. Each method is represented by a different color. Dark blue: Online Graph (ours), orange: PF,
   light gray: STRUCK, dark gray: SPOT, light blue: Offline Graph.}
\label{fig:results_occlusion}
\end{figure}

\begin{figure}[!ht]
\begin{center}
   \subfloat{\includegraphics[width=0.45\linewidth]{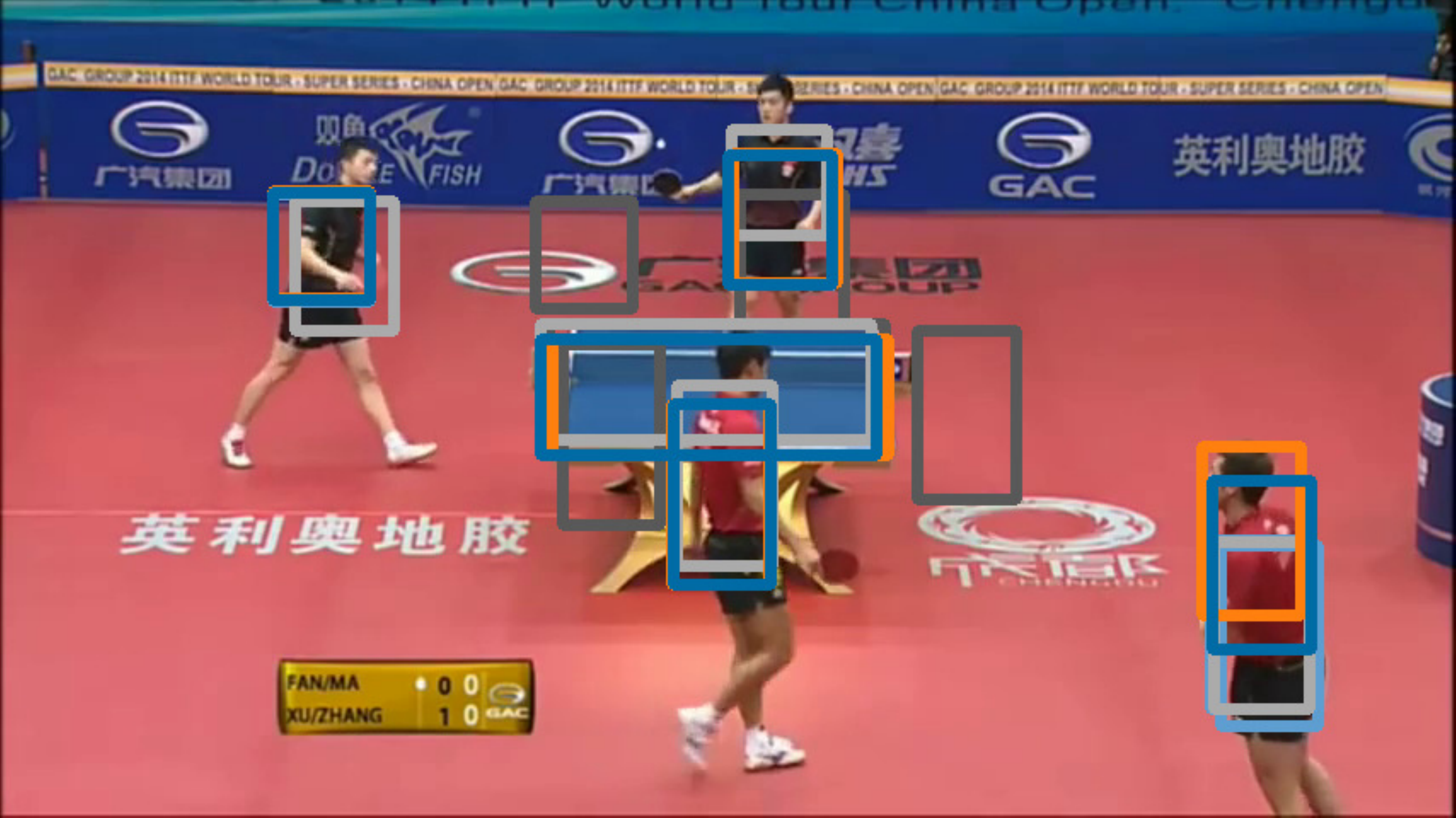}}
   \hfil
   \subfloat{\includegraphics[width=0.45\linewidth]{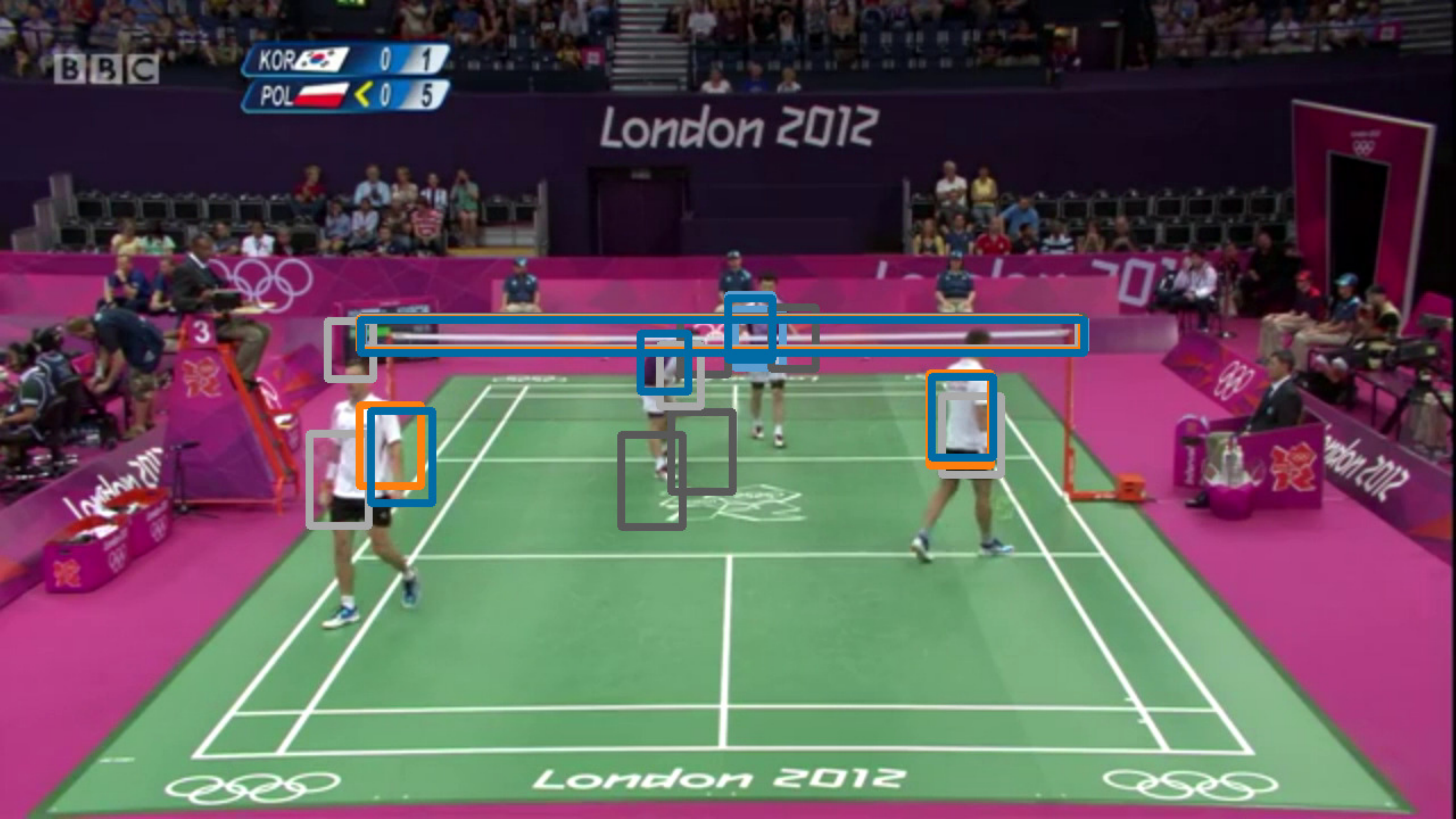}}
   \\
   \subfloat{\includegraphics[width=0.45\linewidth]{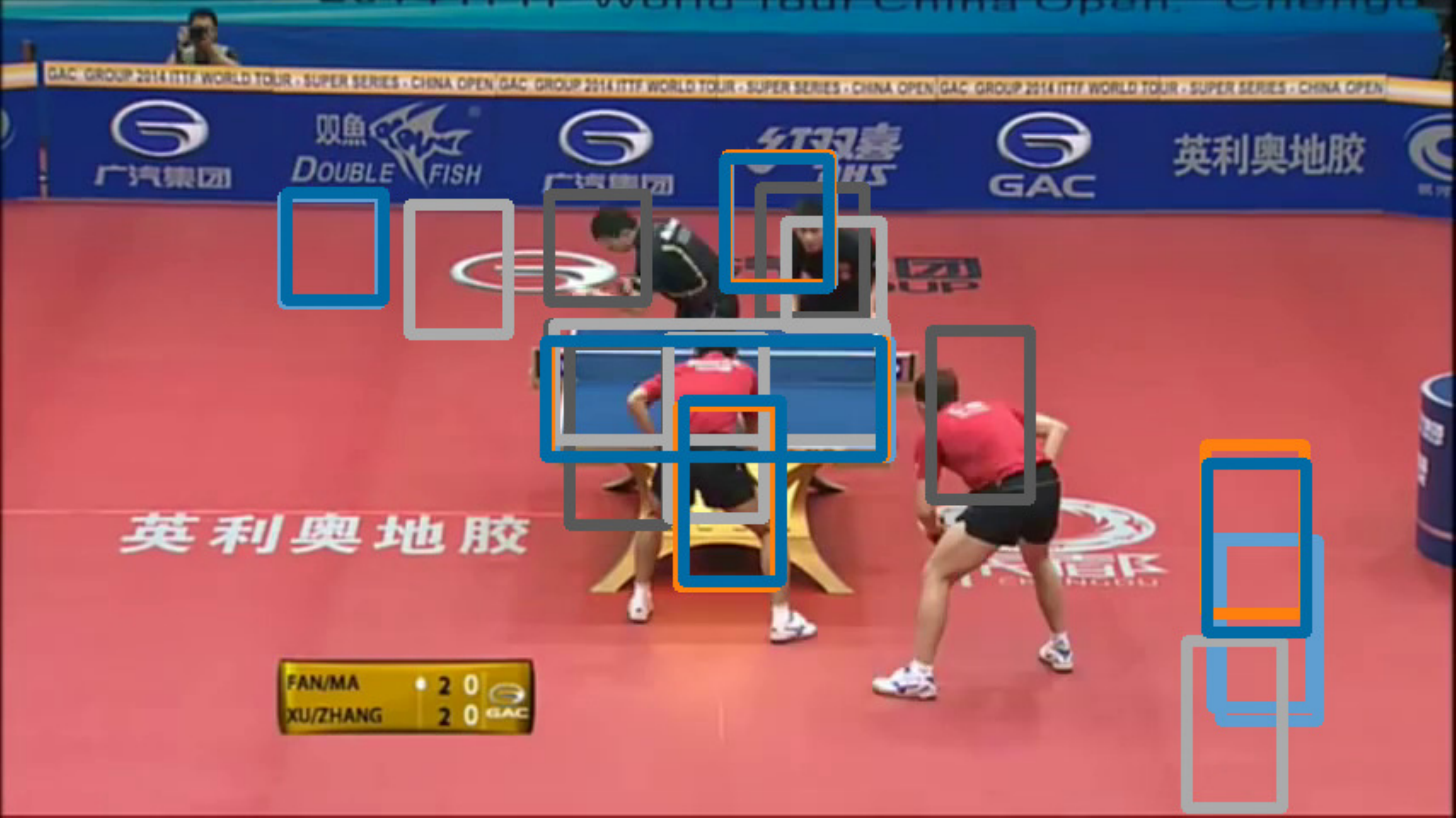}}
   \hfil
   \subfloat{\includegraphics[width=0.45\linewidth]{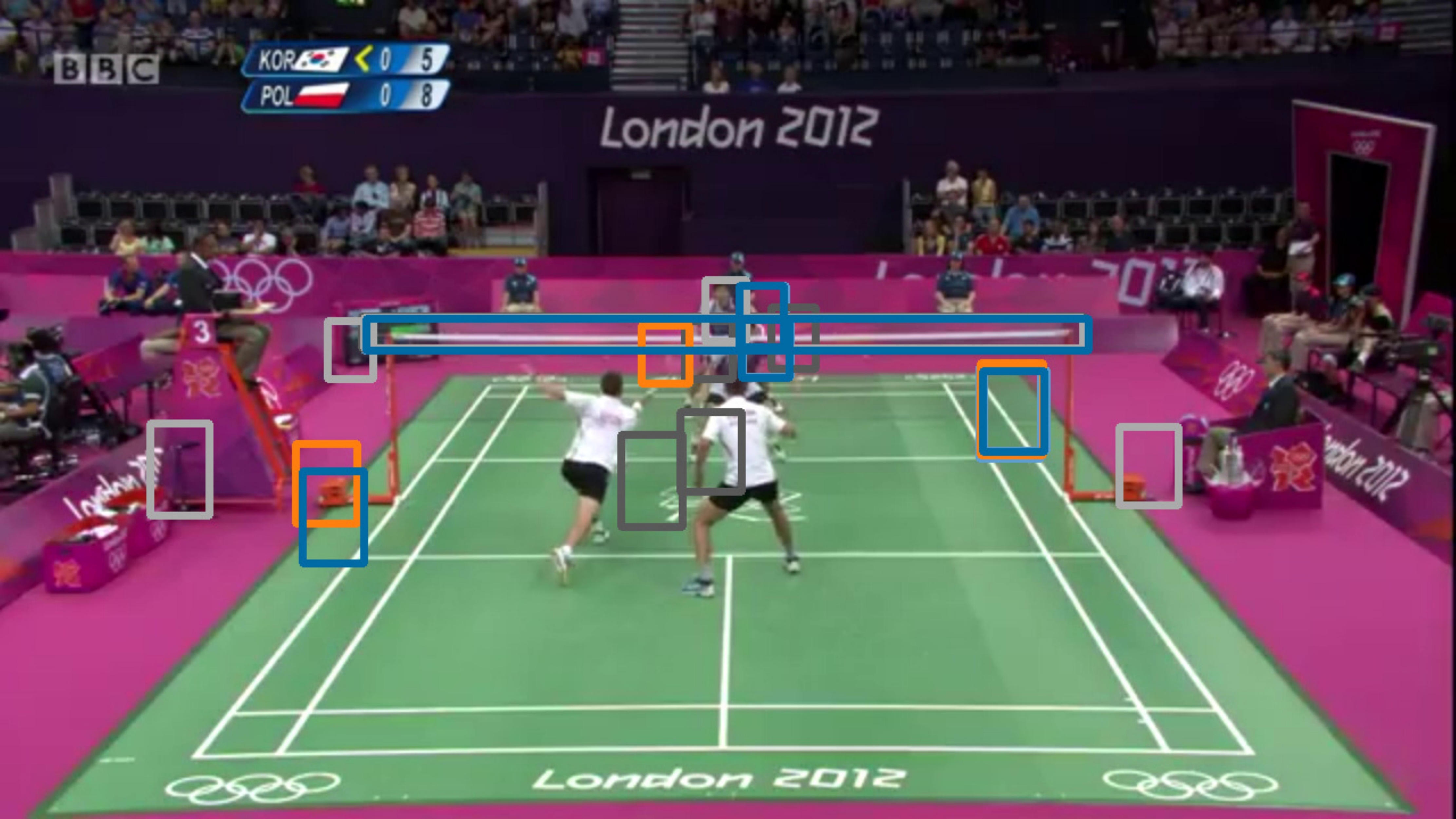}}
   \\
   \subfloat{\includegraphics[width=0.45\linewidth]{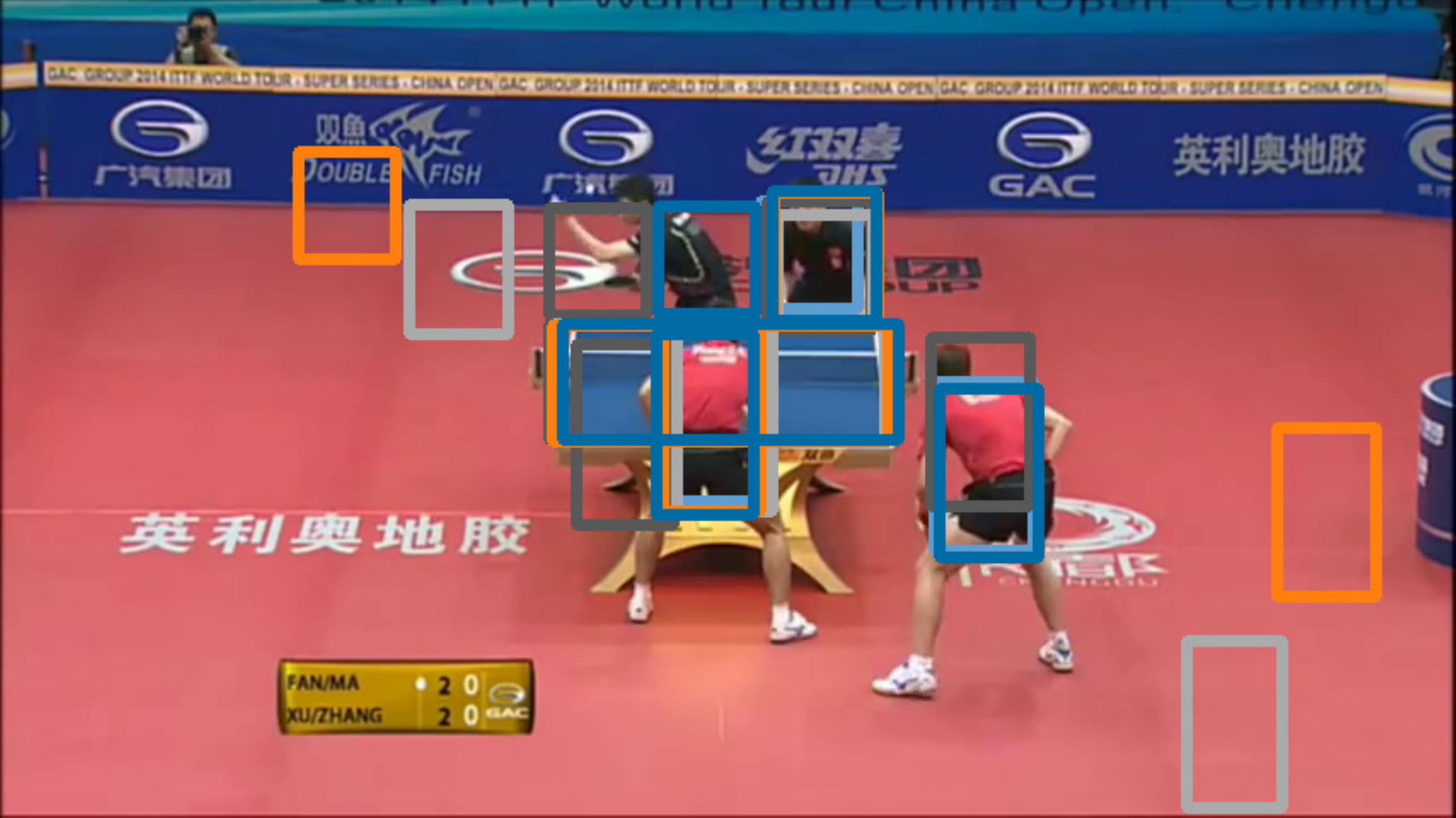}}
   \hfil
   \subfloat{\includegraphics[width=0.45\linewidth]{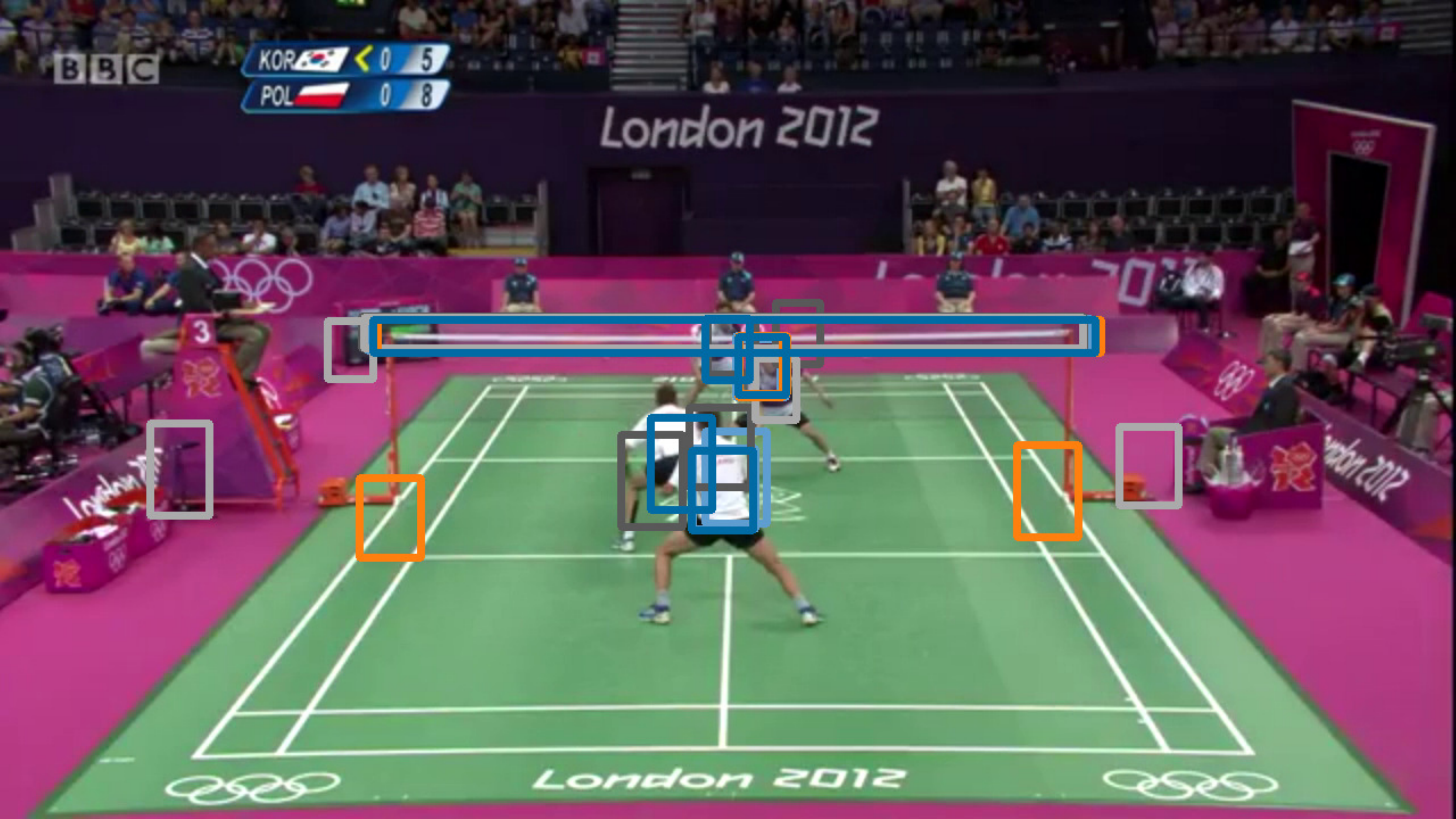}}
\end{center}
   \caption{Tracking results after a camera cut. Each method is represented by a different color. Dark blue: Online Graph (ours), orange: PF,
   light gray: STRUCK, dark gray: SPOT, light blue: Offline Graph.}
\label{fig:results_camera_cut}
\end{figure}

\begin{figure}[!ht]
\begin{center}
   \subfloat{\includegraphics[width=0.45\linewidth]{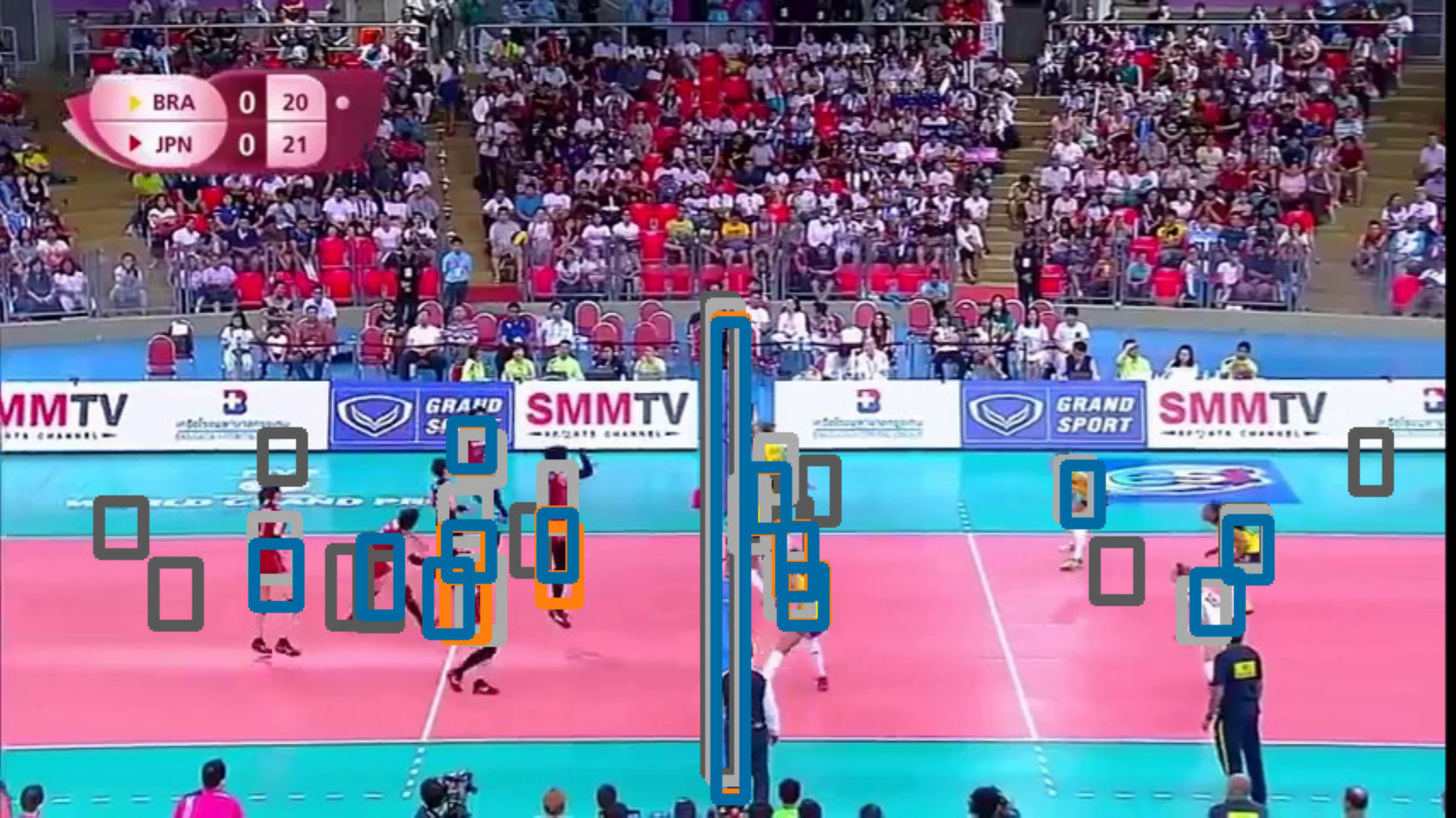}}
   \hfil
   \subfloat{\includegraphics[width=0.45\linewidth]{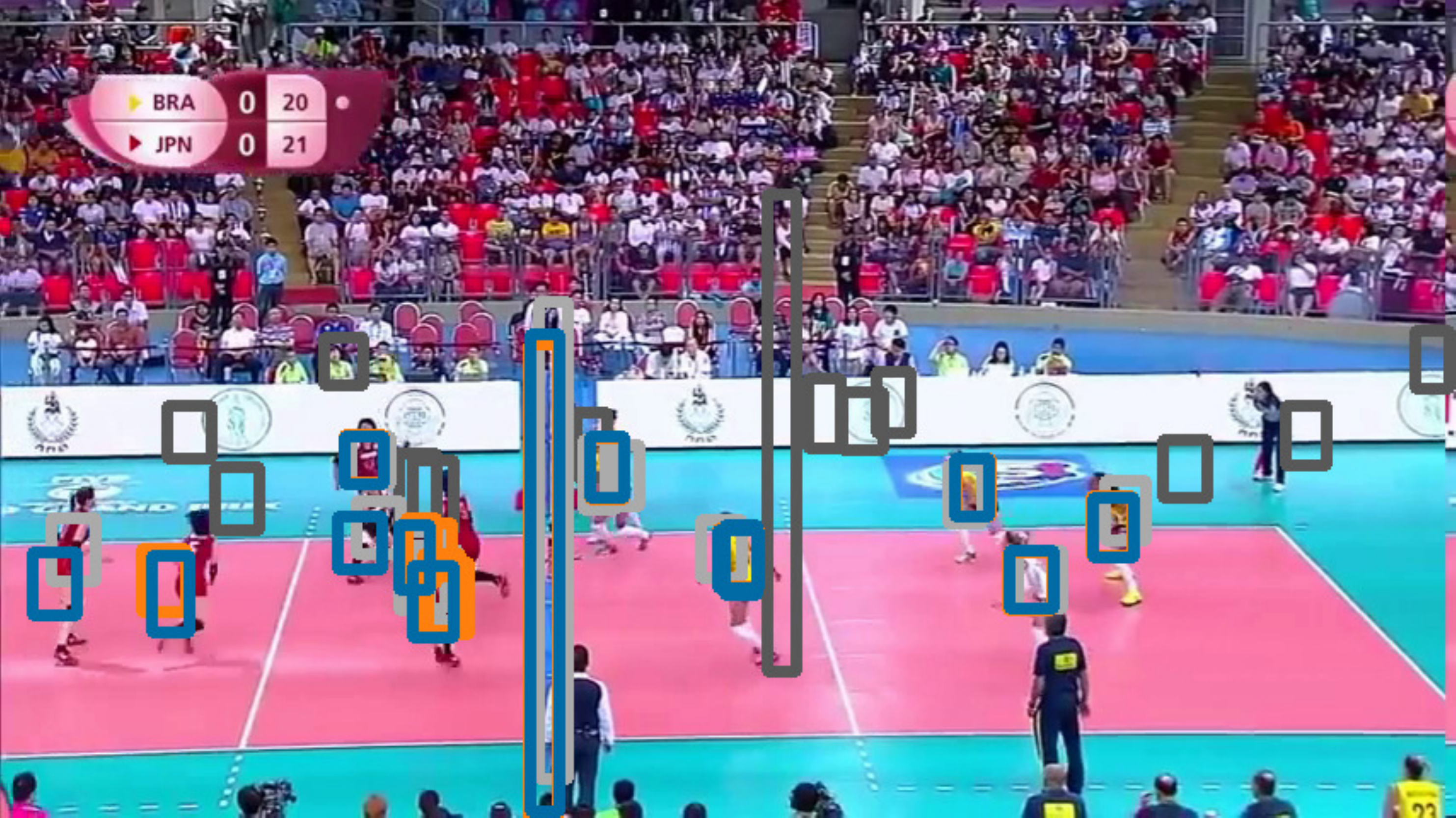}}
   \\
   \subfloat{\includegraphics[width=0.45\linewidth]{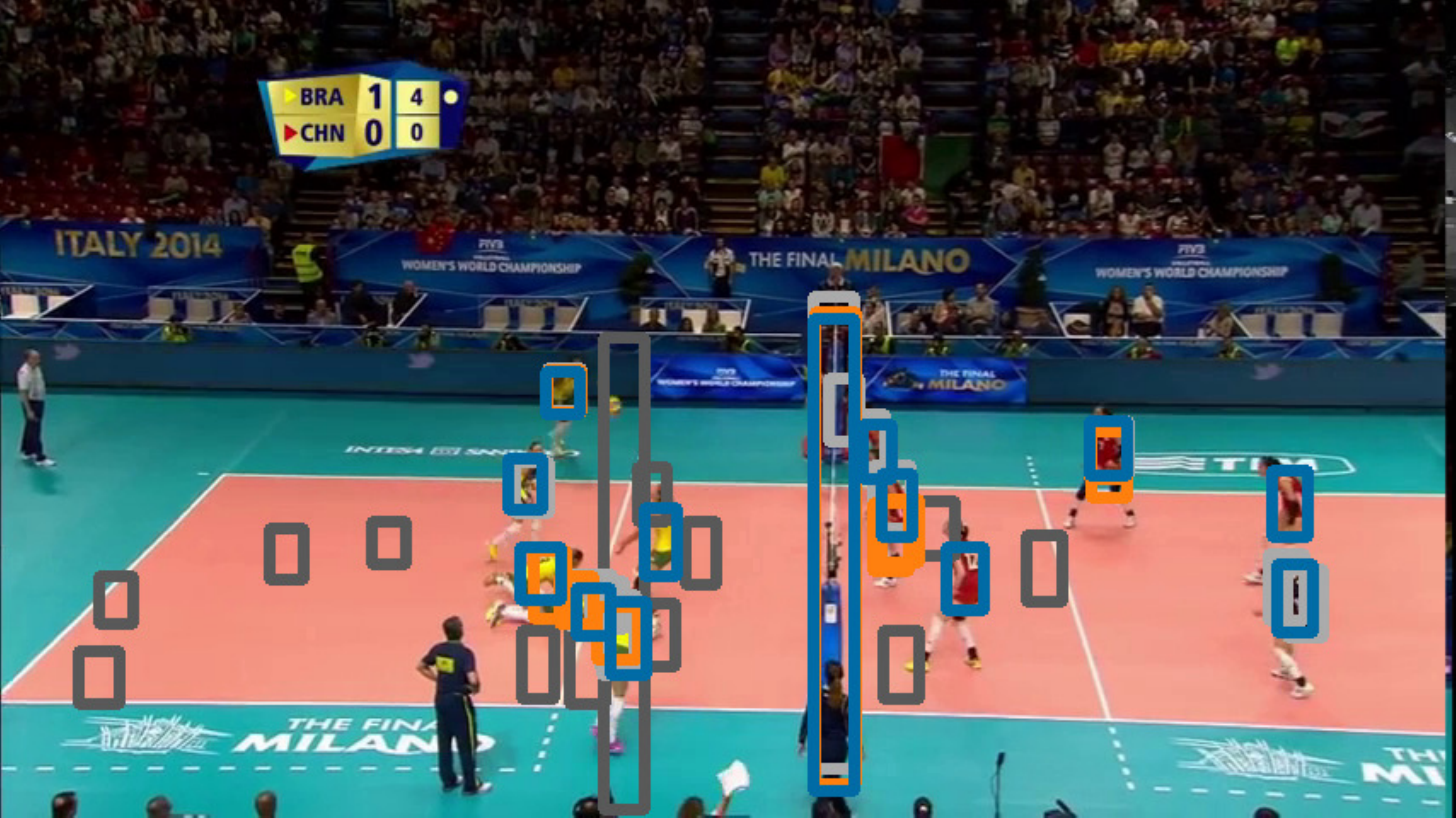}}
   \hfil
   \subfloat{\includegraphics[width=0.45\linewidth]{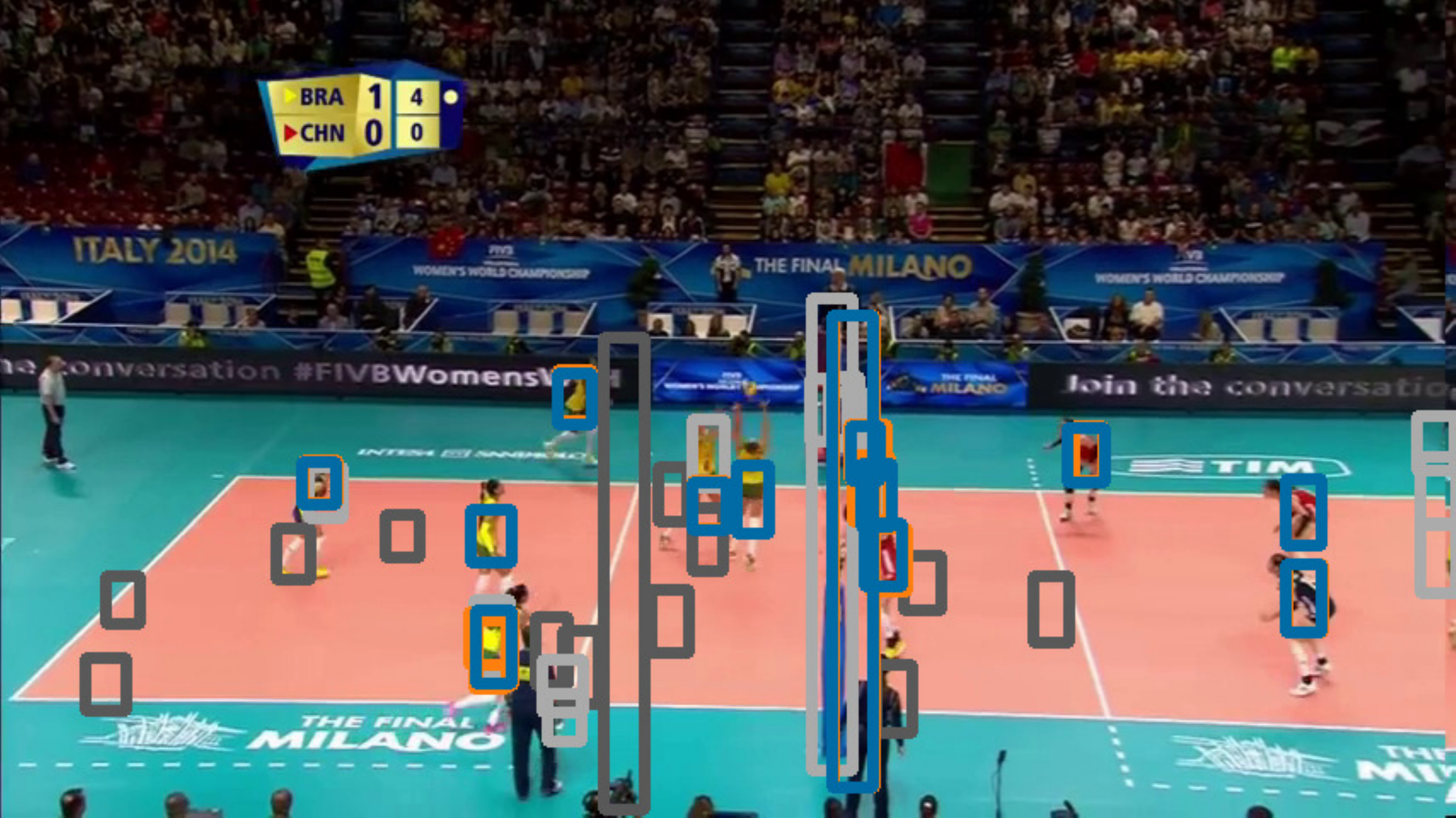}}
   \\
   \subfloat{\includegraphics[width=0.45\linewidth]{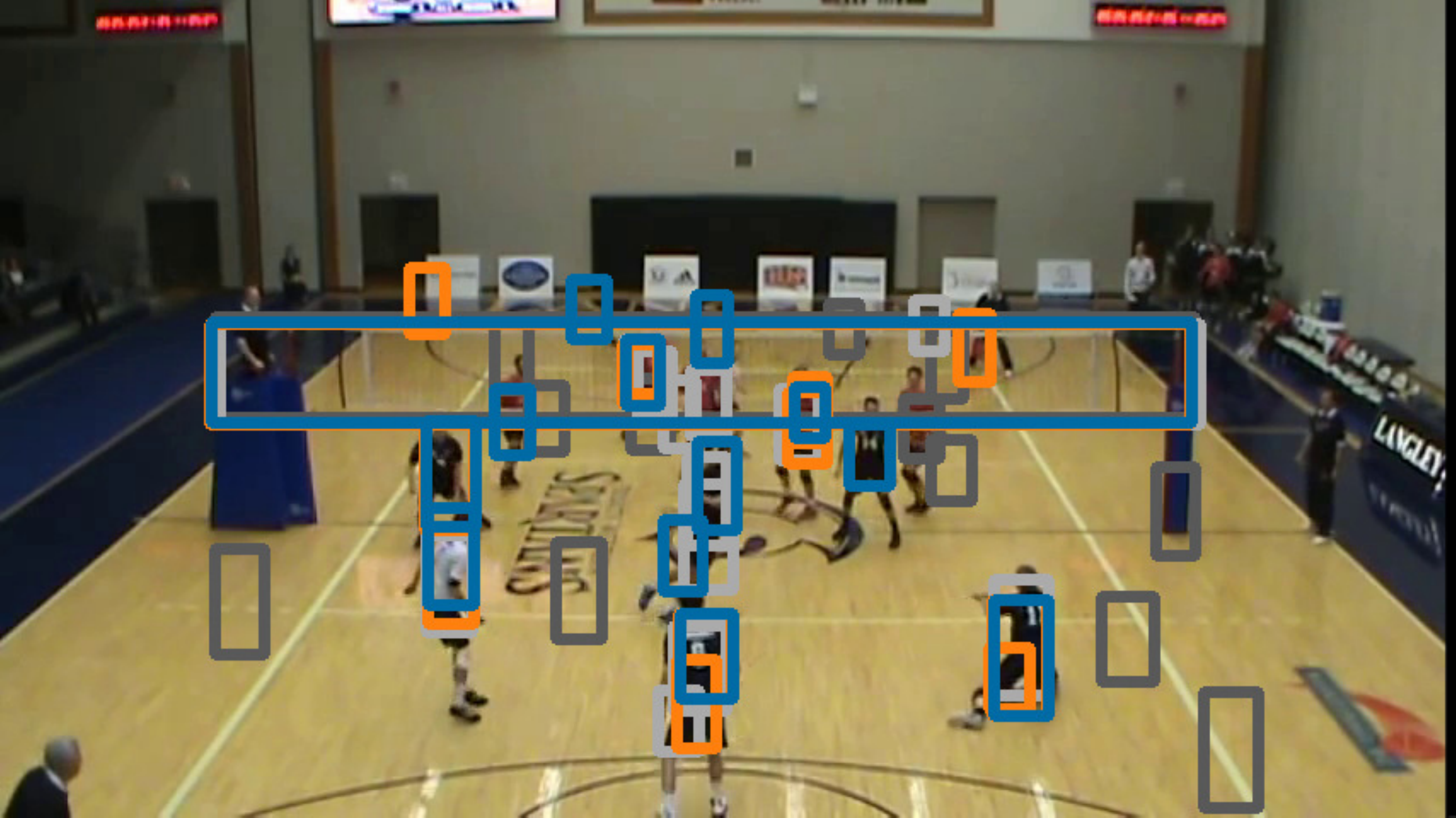}}
   \hfil
   \subfloat{\includegraphics[width=0.45\linewidth]{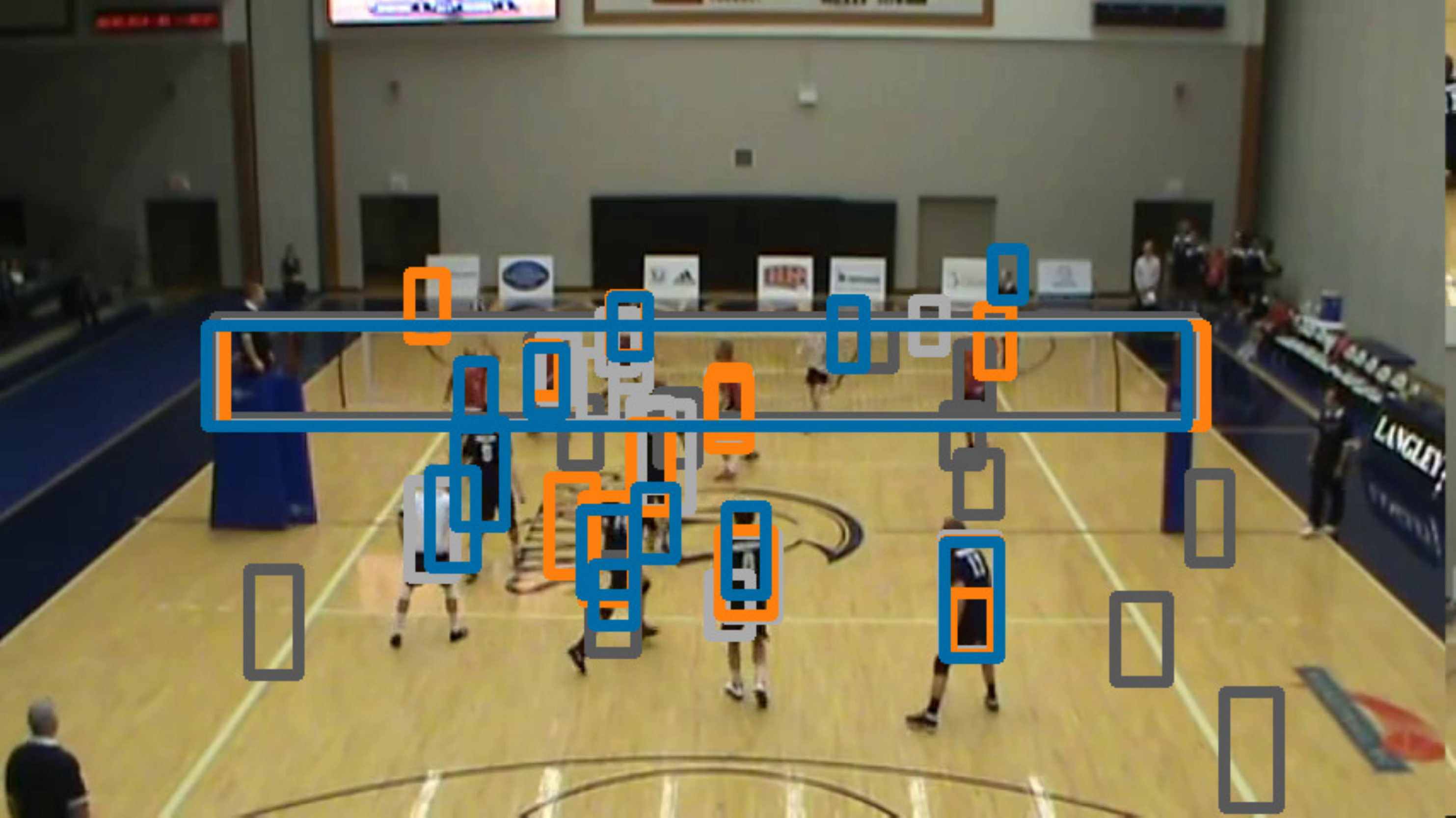}}
\end{center}
   \caption{Tracking results with many objects and a more cluttered scene. Each method is represented by a different color. Dark blue: Online Graph (ours), orange: PF,
   light gray: STRUCK, dark gray: SPOT.}
\label{fig:results_volley}
\end{figure}

Figures~\ref{fig:results_occlusion}, \ref{fig:results_camera_cut} and \ref{fig:results_volley} shows some results observed on the videos. As it can be seen,
our approach successfully recovers tracking after occlusion (Figure~\ref{fig:results_occlusion}) or camera cuts (Figure~\ref{fig:results_camera_cut}), while PF and STRUCK are not
able to re-detect the target after such situations. The videos from the volleyball dataset also present
scenes captured from two different angles and with some camera motion. Even in these more challenging scenes,
with many more objects, the graphs help to keep more correct tracks (Figure~\ref{fig:results_volley}).
It is interesting to note that sometimes even the more robust
STRUCK tracker is not able to deal with temporary occlusion, losing one of the targets, as shown in the last picture
of Figure~\ref{fig:results_occlusion}. SPOT, on the other hand, does not suffer significantly from
abrupt motion. However, as it is evident from the pictures, sometimes the more rigid model ends up causing
many tracking misses at the same time. These results further evidence the flexibility of the proposed method,
that is able to accept a wide range of spatial configurations.

\begin{figure}
\begin{center}
   Youtube table tennis
   
   \subfloat{\includegraphics[width=0.45\linewidth]{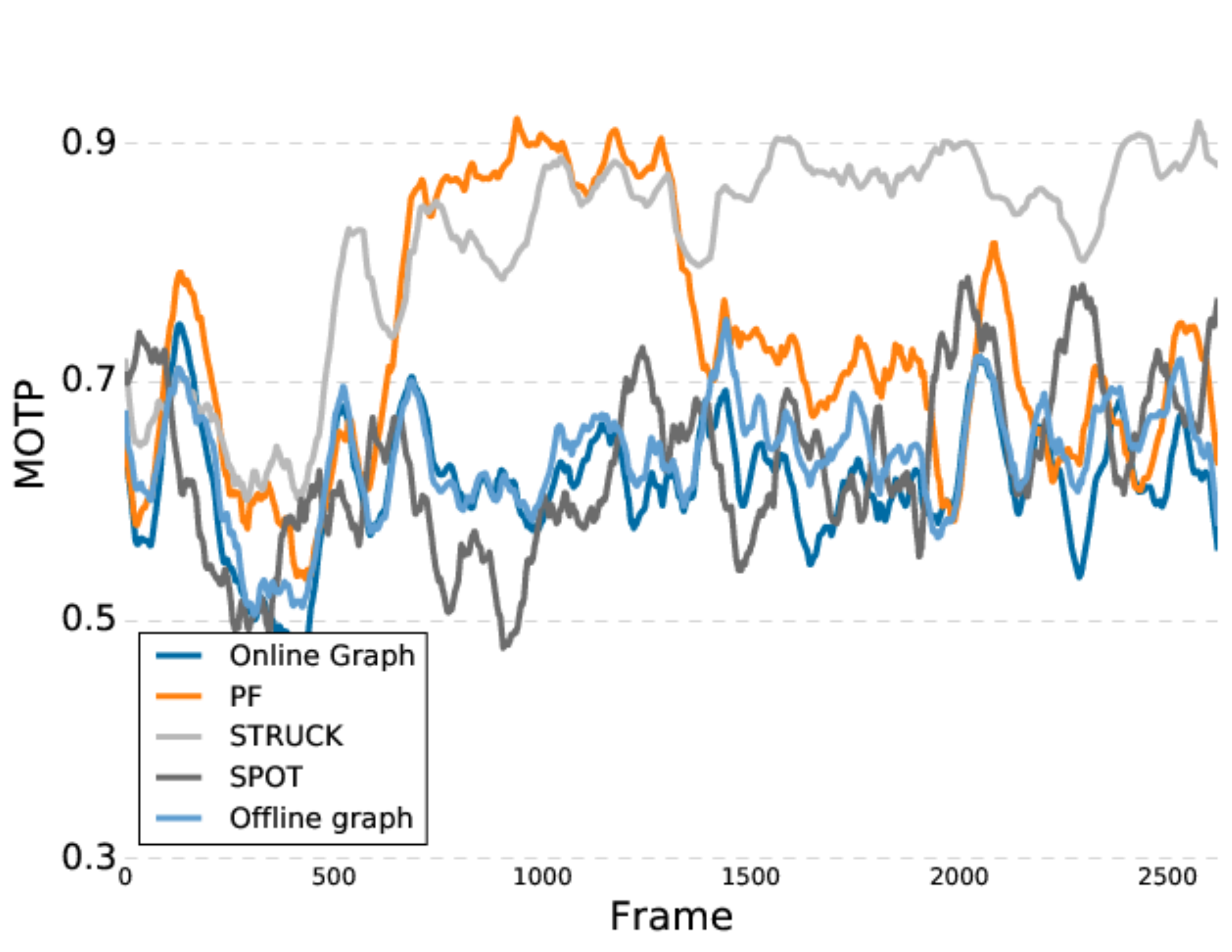}}
   \hfil
   \subfloat{\includegraphics[width=0.45\linewidth]{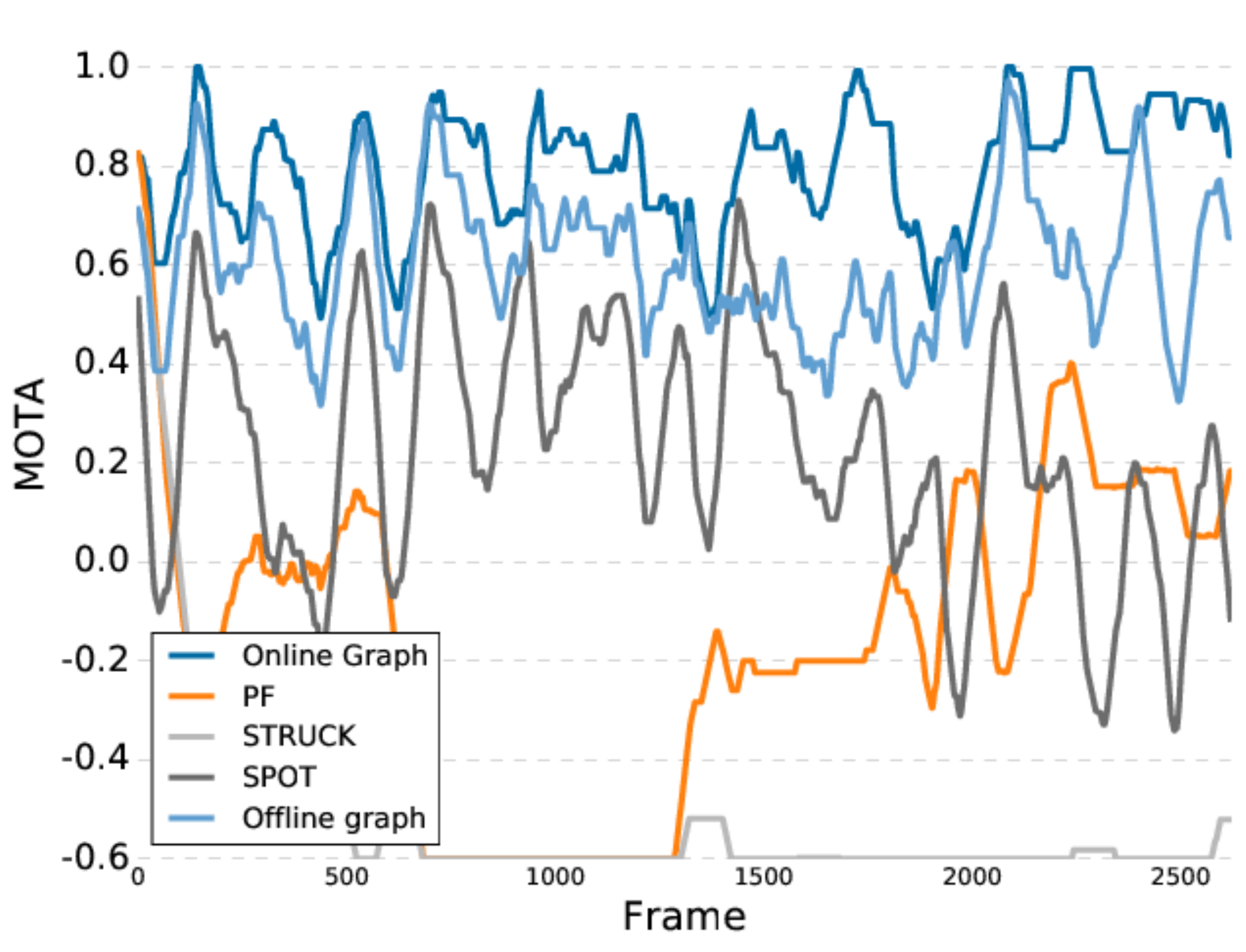}}
   \\
   ACASVA badminton
   
   \subfloat{\includegraphics[width=0.45\linewidth]{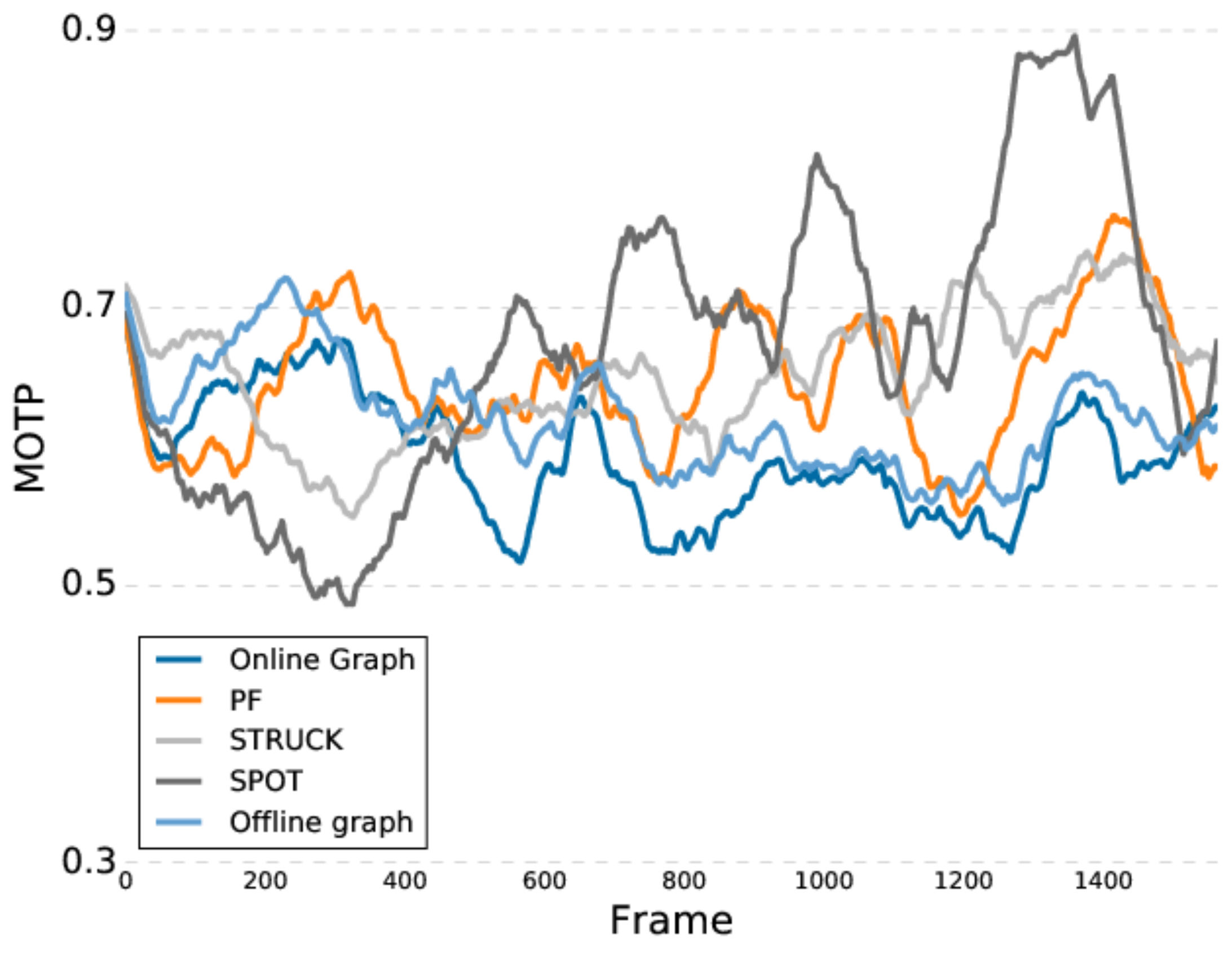}}
   \hfil
   \subfloat{\includegraphics[width=0.45\linewidth]{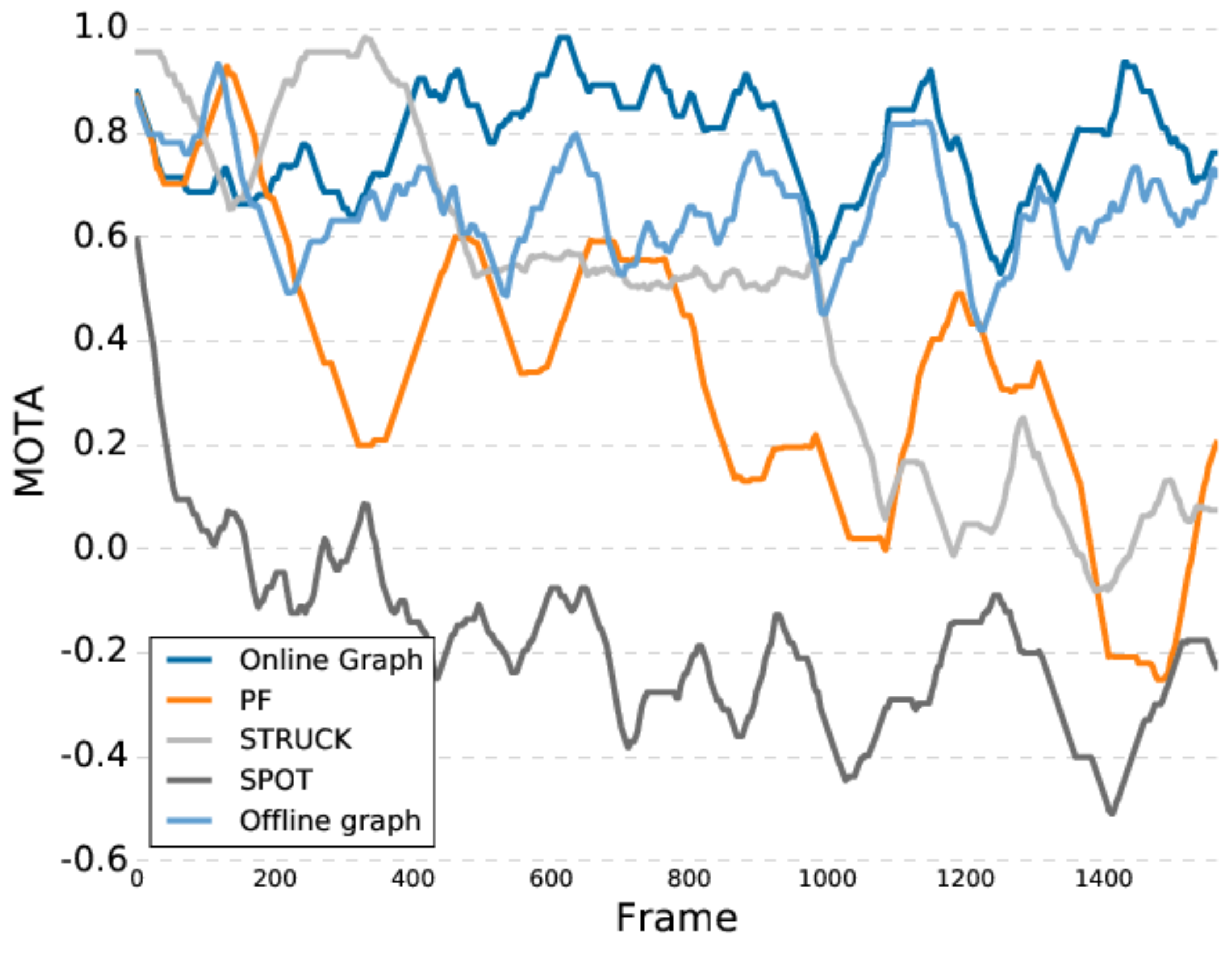}}
   \\
   Youtube volleyball
   
   \subfloat{\includegraphics[width=0.45\linewidth]{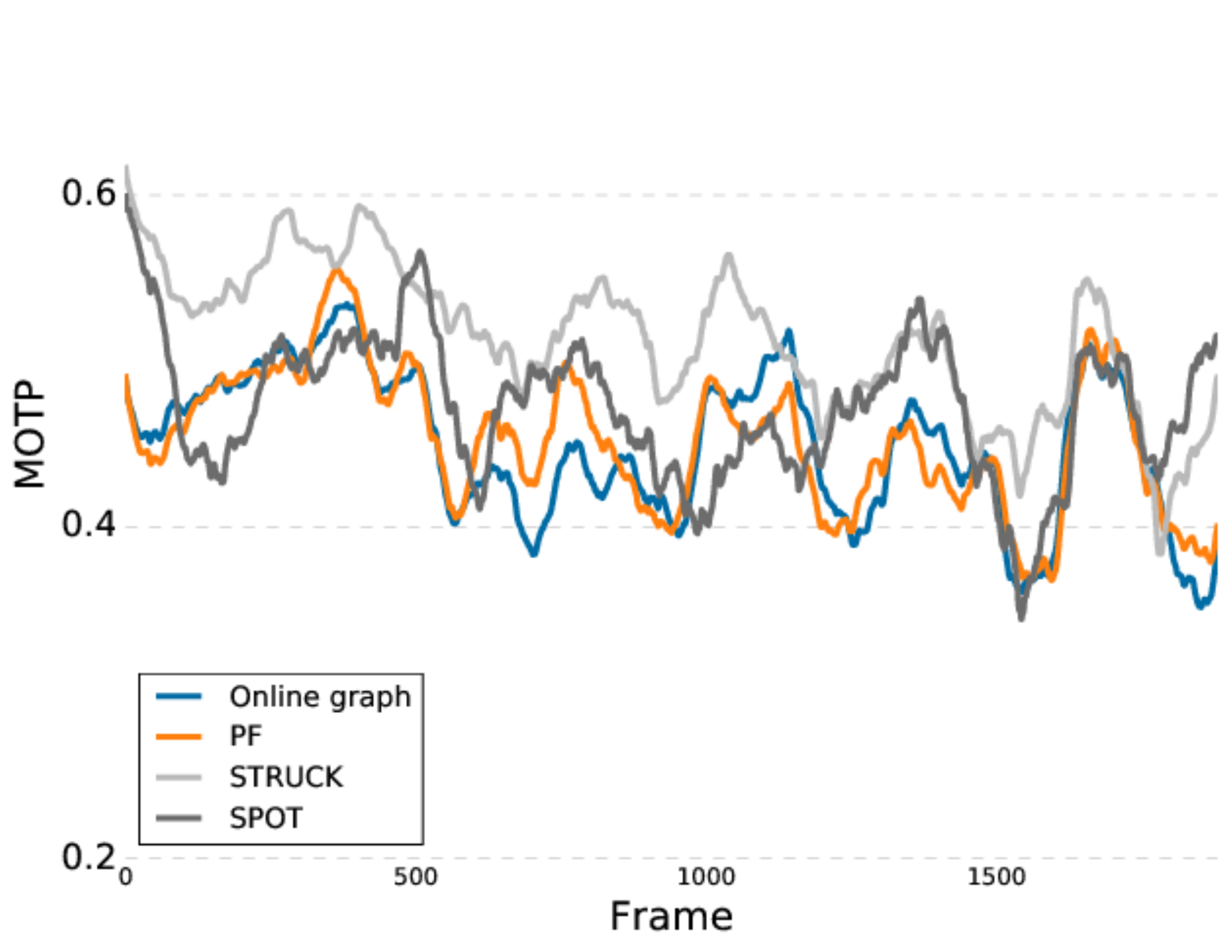}}
   \hfil
   \subfloat{\includegraphics[width=0.45\linewidth]{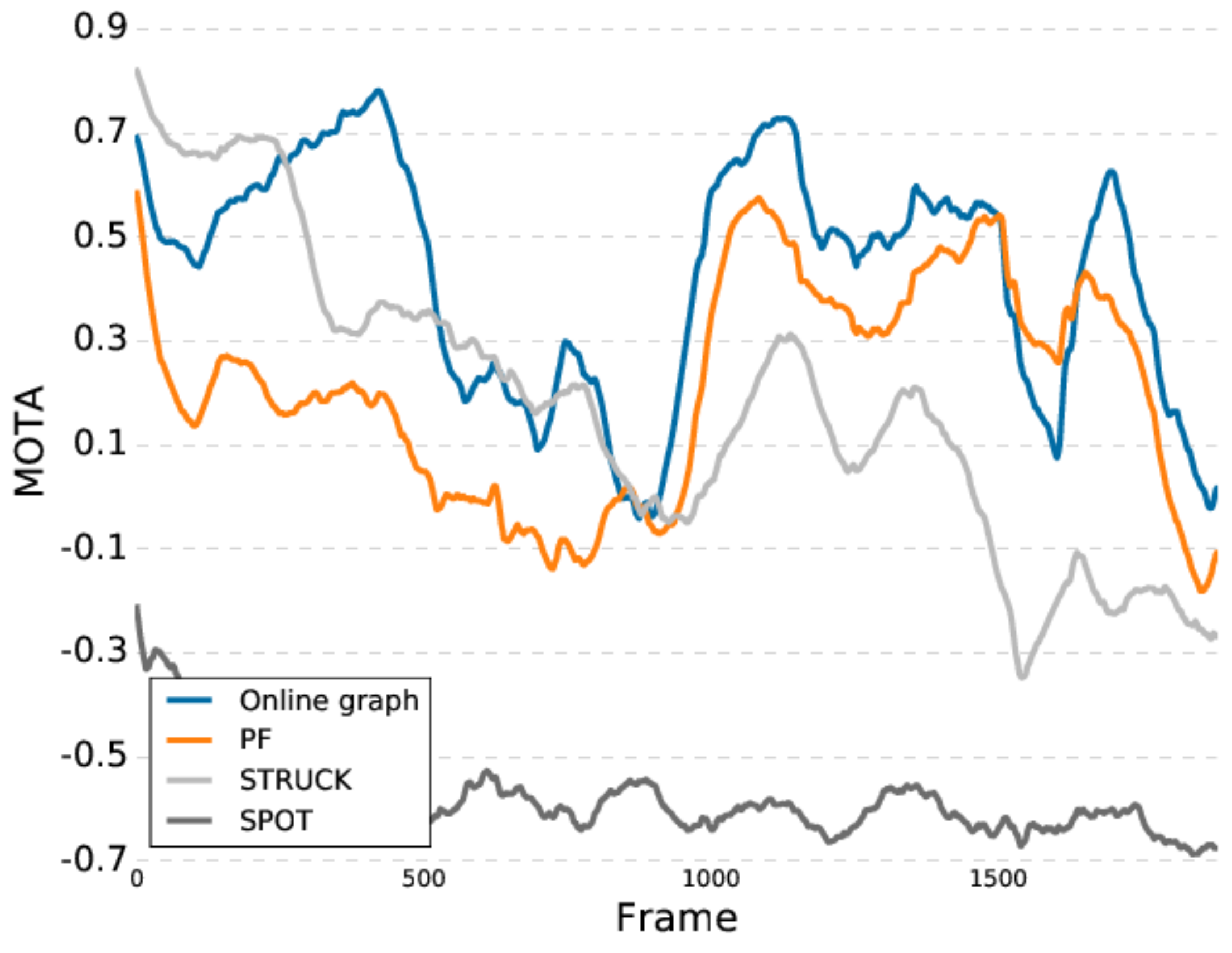}}
\end{center}
   \caption{Instantaneous MOTP and MOTA for one video from each dataset including object overlapping and camera cuts. 
   The charts were smoothed using a moving average window of 100 frames.}
\label{fig:results_charts}
\end{figure}

We also evaluated the behavior of each tracker during the video. For that, we computed the instantaneous $MOTA$ and $MOTP$
values in a single frame. Figure~\ref{fig:results_charts} shows how they vary along time in a video.

One of the first things we can notice is that STRUCK, and sometimes PF present very high MOTP in the Youtube table tennis video.
However, analyzing only this metric may be misleading. This happens because, after a camera cut, most PF and STRUCK trackers
are lost. This causes the MOTP to be evaluated only on the few remaining trackers, which usually shows a higher precision than
when evaluating many matchings.
As can be observed from the MOTA, the accuracy is very low when such events occur, demonstrating that the actual results
are not very good. From the charts, the better accuracy of our
method is evidenced, as it presents the best overall performance in all the videos. By evaluating the accuracy curves,
we can also observe that the other trackers, especially PF and STRUCK, suffer from the drifting problem, \ie performance deteriorates
as the video goes on. The use of graphs fixes this problem, creating much more stable results even
on longer sequences.

\section{Conclusion}
\label{sec:conclusion}

We proposed a graph-based approach to exploit the structural information of a filmed scene and to use it
to improve tracking of multiple objects in structured videos. Each object in the scene represents
one vertex of the graph, and edges are included to consider their spatial relations. The graph is then used to
generate new likely target locations to try to improve tracking during longer periods of occlusion and camera cut.
During the tracking, each object is individually tracked using particle filters. By merging the
current tracking with the candidates generated by the model, multiple graphs are built. They are then evaluated
according to the model, and the best one is chosen as the new global tracking state.
The source code of the developed framework is publicly available for testing.

One of the advantages of the proposed framework is that it does not really rely on any information specific
to a particular tracker. Therefore, the single object tracker could be potentially replaced by any other more suitable choice
for other types of objects. This makes this approach very flexible and able to deal with a wider range of
applications.

The results show that the proposed method successfully increases the tracking precision over other
state-of-the-art methods for sports structured videos. As shown by the results, the candidates
generated by the structural properties are successfully able to recover tracking in case of loss,
while keeping their identities correct. This, in turn, greatly contributes to decrease drifting,
which is also a challenging condition to deal with in longer videos. These experiments
showed the robustness of the method to handle inter-object occlusion and video cuts
from a single camera. Situations depicting the use of moving cameras are also
supported, as long as the camera movement is limited (as evidenced by the volleyball tests), or smooth enough to provide
sufficient time for the graphs to adapt to the new distributions. Video cuts between different cameras can
also be accepted to some extent. If the cameras maintain roughly the same scene structure, \eg slightly
different angles or a change from a rear/side view to a top one without changing orientation,
the graph configuration would not change so abruptly and the graph would be able to incorporate
these changes. On the other hand, if the change is large, such as changing from one side of the court
to the other, the method most likely would not be able to adapt well to the new situation.

One limitation of the proposed framework is that the color-based tracker used for each object is not
very robust against appearance or illumination changes. It is also sensitive to initialization parameters,
\ie tracking may present poor results if the provided bounding box does not cover the object properly.
As the graphs also use the tracker score as a vertex attribute for evaluating, if the color model is not representative enough,
the whole tracking may be affected. Therefore, one future extension of this work consists in
replacing the color-based particle filter single tracker by a more robust one, such as STRUCK~\cite{hare2011struck}.
The graphs could also be enriched by including other types of edge attributes. For example,
we could encode more complex appearance and structural relations~\cite{bloch2006ternary}.

As another future development, we want to improve the method by making it more self-adaptive.
One way to do so is to automatically adjust the number of candidates generated
from each reference object, or to use the global structure as a whole to choose the best locations.
This could be done by computing a reliability score for each object, combining the hypothesis of all of
references into a single set and choosing only the best options.

\section*{Acknowledgements}

The authors are grateful to S\~ ao Paulo Research Foundation (FAPESP) (grants \#2011/50761-2, \#2012/09741-0, \#2013/08258-7
and \#2014/50135-2), CNPq, CAPES, NAP eScience - PRP - USP, and PICS CNRS/FAPESP 227332 (2014-2015).



\section*{References}
\bibliographystyle{elsarticle-num} 
\bibliography{cviu-struct}

\end{document}